\documentclass[11pt,a4paper,preprint, authoryear]{elsarticle}

\usepackage{makecell}

\newlength{\cslhangindent}
\newenvironment{CSLReferences}%

\usepackage{lmodern}
\usepackage{setspace}
\DeclareMathSizes{12}{14}{10}{10}

\usepackage{float}
\let\origfigure\figure
\let\endorigfigure\endfigure
\renewenvironment{figure}[1][2] {
    \expandafter\origfigure\expandafter[H]
} {
    \endorigfigure
}

\let\origtable\table
\let\endorigtable\endtable
\renewenvironment{table}[1][2] {
    \expandafter\origtable\expandafter[H]
} {
    \endorigtable
}

\usepackage{ifxetex,ifluatex}
\usepackage{fixltx2e} 
\ifnum 0\ifxetex 1\fi\ifluatex 1\fi=0 
  \usepackage[T1]{fontenc}
  \usepackage[utf8]{inputenc}
\else 
  \ifxetex
    \usepackage{mathspec}
    \usepackage{xltxtra,xunicode}
  \else
    \usepackage{fontspec}
  \fi
  \defaultfontfeatures{Mapping=tex-text,Scale=MatchLowercase}
  
\fi

\usepackage{amssymb, amsmath, amsthm, amsfonts}
\theoremstyle{plain}

\newtheorem{proposition}{Proposition}

\usepackage[round]{natbib}

\usepackage{longtable}
\usepackage[margin=2.3cm,bottom=2cm,top=2.5cm, includefoot]{geometry}
\usepackage{fancyhdr}
\usepackage[bottom, hang, flushmargin]{footmisc}
\usepackage{graphicx}
\numberwithin{equation}{section}
\numberwithin{figure}{section}
\numberwithin{table}{section}
\usepackage{textcomp}

\usepackage{array}
\newcolumntype{x}[1]{>{\centering\arraybackslash\hspace{0pt}}p{#1}}

\makeatletter
\def\ps@pprintTitle{%
  \let\@oddhead\@empty
  \let\@evenhead\@empty
  \let\@oddfoot\@empty
  \let\@evenfoot\@oddfoot
}
\makeatother

\usepackage{hyperref}
\hypersetup{breaklinks=true,
            bookmarks=true,
            colorlinks=true,
            citecolor=blue,
            urlcolor=blue,
            linkcolor=blue,
            pdfborder={0 0 0}}

\usepackage{siunitx}
\usepackage{multirow}
\usepackage{hhline}
\usepackage{calc}
\usepackage{tabularx}
\usepackage{booktabs}
\usepackage{caption}

\def\useignorespacesandallpars#1\ignorespaces\fi{%
#1\fi\ignorespacesandallpars}

\makeatletter
\def\ignorespacesandallpars{%
  \@ifnextchar\par
    {\expandafter\ignorespacesandallpars\@gobble}%
    {}%
}
\makeatother

\let\rmarkdownfootnote\footnote%
\def\footnote{\protect\rmarkdownfootnote}
\IfFileExists{upquote.sty}{\usepackage{upquote}}{}

\begin{document}

\begin{frontmatter}  %

\title{An Interpretable Neural Network for Parameter Inference}

\author[Add1]{Johann Pfitzinger\footnote{This paper represents a chapter
  of my PhD thesis submitted at Goethe University Frankfurt. I thank my
  supervisor Uwe Hassler for his advice.}}
\ead{johann.pfitzinger@gmail.com}

\address[Add1]{Goethe University, Frankfurt am Main}

\begin{abstract}
\small{
Adoption of deep neural networks in fields such as economics or finance
has been constrained by the lack of interpretability of model outcomes.
This paper proposes a generative neural network architecture --- the
parameter encoder neural network (PENN) --- capable of estimating local
posterior distributions for the parameters of a regression model. The
parameters fully explain predictions in terms of the inputs and permit
visualization, interpretation and inference in the presence of complex
heterogeneous effects and feature dependencies. The use of Bayesian
inference techniques offers an intuitive mechanism to regularize local
parameter estimates towards a stable solution, and to reduce
noise-fitting in settings of limited data availability. The proposed
neural network is particularly well-suited to applications in economics
and finance, where parameter inference plays an important role. An
application to an asset pricing problem demonstrates how the PENN can be
used to explore nonlinear risk dynamics in financial markets, and to
compare empirical nonlinear effects to behavior posited by financial
theory.
}
\end{abstract}

\vspace{1cm}

\begin{keyword}
\footnotesize{
Self-explaining neural networks \sep explainable machine learning
\sep deep learning \sep Bayesian machine learning \sep variational
inference \sep parameter inference \sep asset pricing \\ \vspace{0.3cm}
\textit{JEL classification} C11, C14, C45, C52, G12
}
\end{keyword}
\vspace{0.5cm}
\end{frontmatter}

\newpage

\pagestyle{fancy}
\chead{}
\rhead{}
\lfoot{}
\rfoot{\footnotesize Page \thepage}
\lhead{}
\cfoot{}


\headsep 35pt 

\hypertarget{introduction}{%
\section{Introduction}\label{introduction}}

Deep learning is rapidly emerging as the most influential sub-field of
machine learning, due in large part to substantial gains in predictive
performance achieved by deep neural networks (DNN) in the fields of
computer vision and natural language processing
(\protect\hyperlink{ref-goodfellowDeepLearning2016}{Goodfellow \emph{et
al.}, 2016}). Fueled by successes in these and other domains, a large
literature has developed, applying DNN techniques to economic and
financial problems. However, despite some appealing properties of DNN
(e.g.~the oft-cited universal approximation
(\protect\hyperlink{ref-hornikMultilayerFeedforwardNetworks1989}{Hornik
\emph{et al.}, 1989}) or predictive accuracy
(\protect\hyperlink{ref-fernandez-delgadoWeNeedHundreds2014}{Fernandez-Delgado
\emph{et al.}, 2014})), the scope for their use in econometrics remains
limited for several reasons.

The most important barrier to the wider adoption of DNN in the field of
econometrics is their lack of interpretability. Econometric analysis is
typically concerned with inference about the causal dynamics governing
economic processes (e.g.~expected responses to policy innovations). This
requires an identifiable parametric representation of the process,
precluding the use of DNN as well as most other machine learning
methods. A growing literature proposes the use of \emph{post hoc}
algorithms to interpret the results of neural networks, however these
methods often lack robustness, are cumbersome to implement and
computationally demanding
(\protect\hyperlink{ref-alvarez-melisRobustnessInterpretabilityMethods2018}{Alvarez-Melis
\& Jaakkola, 2018}). In addition, --- and perhaps more importantly ---
\emph{post hoc} interpretation is generally facilitated by imposing
simplifying assumptions, such as feature independence, onto complex
algorithms. Since the structure of DNN makes them appropriate precisely
for problems characterized by feature dependencies, removing this
property from the interpretation vastly reduces its power to describe
the underlying data generating process.

Another impediment to the use of DNN in econometrics is the relatively
small data typically available in empirical applications. Skillful
consideration of the network architecture and regularization are
required to avoid overfitting of DNN in small samples, while
simultaneously capturing systematic nonlinearities. The reliance on
pseudo out-of-sample model selection algorithms, such as
cross-validation, further exacerbates the issue, and puts DNN at a
disadvantage compared to simpler methods like linear regression or
nonlinear additive models.

In this paper, I propose a neural network architecture that aims to
solve both of the above obstacles to the application of DNN in
econometrics. The parameter encoder neural network (PENN) represents a
novel contribution to the nascent field of self-explaining neural
networks, where the architecture of the DNN is designed in such a manner
as to produce interpretable outputs natively. The method retains the
flexibility of the neural network to encode complex nonlinear behavior,
but simultaneously generates interpretable posterior densities of local
regression parameters. A Bayesian prior shrinks the local parameter
estimates towards global means, reducing the process to static
parameters when the data do not support nonlinearity. This form of
regularization is extremely intuitive and permits the PENN to be used
even in comparatively data-constrained environments.

The contribution of the PENN model can be viewed from two perspectives.
On the one hand, it represents an explainability method, that compels a
complex neural network to encode effects via a latent channel of local
regression parameters. The regression parameters define a linear
decomposition of each prediction that corresponds to the local
contributions produced by popular explainability algorithms such as SHAP
(\protect\hyperlink{ref-lundbergUnifiedApproachInterpreting2017}{Lundberg
\& Lee, 2017}) or LIME
(\protect\hyperlink{ref-ribeiroWhyShouldTrust2016}{Ribeiro \emph{et
al.}, 2016}), with the important distinction that the PENN requires no
assumption of feature independence.\footnote{The explainability
  algorithms mentioned here are discussed in detail in subsequent
  sections.} On the other hand, the PENN is a nonlinear regression
technique that can be used to conduct parameter inference in econometric
models and to explore marginal effects at the local level. Examples are
presented in this paper to highlight each of these facets.

The role of the PENN as an explainability method is explored using a
series of simulations, which demonstrate that the PENN can generate
consistent local parameter estimates superior, in several respects, to
the feature contributions obtained using popular \emph{post hoc}
explainability algorithms, as well as other interpretable nonlinear
frameworks. The presence of interaction effects among covariates can
lead to large inaccuracies in estimators that assume an additive effects
structure (as is the case for most existing explainability algorithms).
The parameters estimated using the PENN capture non-additive effects
correctly, and are comparatively robust to several characteristics
commonly observed in economic data, such as reduced data availability,
multicollinearity and a low signal-to-noise ratio.

In an applied econometric setting, the PENN method is used to estimate a
nonlinear version of the popular capital asset pricing model (CAPM). The
approach permits the exploration of dynamic dependencies between equity
risk premia and the economic regime, and can be used to test theoretical
assumptions about equity risk and return characteristics. Financial
theory suggests that an asset's sensitivity to systematic risk sources
is not static, but depends on the state that the economy resides in at
any point in time. The PENN is uniquely suited to the estimation of
dynamic risk premia conditional on a nonlinear and highly flexible
macroeconomic regime characterization. In addition, the neural network
structure of the PENN permits real-time predictions of the cost of
equity, addressing one of the most important empirical critiques to the
CAPM: its backward-looking nature.

The remainder of the paper is structured as follows: Section
\ref{literature} examines the existing literature on explainable machine
learning methods. Section \ref{method} introduces the PENN framework,
while Sections \ref{simulation} and \ref{application} provide simulated
and empirical applications, respectively. Finally, Section
\ref{conclusion} concludes the paper.

\hypertarget{literature-review}{%
\section{\texorpdfstring{Literature review
\label{literature}}{Literature review }}\label{literature-review}}

The field of explainable machine learning has recently experienced an
explosion in research interest, driven by the need to develop
interpretable techniques in decision-critical domains --- a requirement
that is increasingly also reflected in legal frameworks
(\protect\hyperlink{ref-euRegulationEU20162016}{EU, 2016}). Several
approaches have been put forward to understand the inner workings of
black box algorithms such as neural networks, and can be grouped into
local and global attribution methods
(\protect\hyperlink{ref-molnarInterpretableMachineLearning2020}{Molnar,
2020};
\protect\hyperlink{ref-montavonMethodsInterpretingUnderstanding2018}{Montavon
\emph{et al.}, 2018}). The algorithms are usually evaluated \emph{post
hoc}, by perturbing the input to a fitted machine learning model and
assessing the effect on predictions. Global attribution methods, such as
\protect\hyperlink{ref-friedmanGreedyFunctionApproximation2001}{Friedman}
(\protect\hyperlink{ref-friedmanGreedyFunctionApproximation2001}{2001}),
\protect\hyperlink{ref-breimanRandomForests2001}{Breiman}
(\protect\hyperlink{ref-breimanRandomForests2001}{2001}) and
\protect\hyperlink{ref-fisherAllModelsAre2019}{Fisher \emph{et al.}}
(\protect\hyperlink{ref-fisherAllModelsAre2019}{2019}), derive an
overall estimate of the sensitivity of predictions to each covariate.
Conversely, local attribution methods calculate the marginal
contribution of each covariate to a single prediction. Notable instances
of the latter have been proposed by
\protect\hyperlink{ref-goldsteinPeekingBlackBox2014}{Goldstein \emph{et
al.}} (\protect\hyperlink{ref-goldsteinPeekingBlackBox2014}{2014}),
\protect\hyperlink{ref-strumbeljExplainingPredictionModels2014}{Štrumbelj
\& Kononenko}
(\protect\hyperlink{ref-strumbeljExplainingPredictionModels2014}{2014}),
\protect\hyperlink{ref-ribeiroWhyShouldTrust2016}{Ribeiro \emph{et al.}}
(\protect\hyperlink{ref-ribeiroWhyShouldTrust2016}{2016}),
\protect\hyperlink{ref-lundbergUnifiedApproachInterpreting2017}{Lundberg
\& Lee}
(\protect\hyperlink{ref-lundbergUnifiedApproachInterpreting2017}{2017}),
\protect\hyperlink{ref-shrikumarLearningImportantFeatures2019}{Shrikumar
\emph{et al.}}
(\protect\hyperlink{ref-shrikumarLearningImportantFeatures2019}{2019})
and \protect\hyperlink{ref-aasExplainingIndividualPredictions2020}{Aas
\emph{et al.}}
(\protect\hyperlink{ref-aasExplainingIndividualPredictions2020}{2020}).
None of these methods adjust the model architecture, but represent
\emph{a posteriori} analyses of the outputs, implying that they are
model-agnostic and can be used with a wide range of black box
algorithms.

Interpreting machine learning algorithms such as tree-based methods or
neural networks is challenging primarily due to the rich feature
interactions captured by these models. As will be shown in Section
\ref{simulation}, popular \emph{post hoc} algorithms --- which
superimpose interpretable (and thus inevitably simpler) models onto
complex ones --- do not necessarily perform well in the presence of
feature interactions and dependencies. Since the difficulty in
interpreting machine learning models stems from their complexity,
another approach to interpretability has been to simplify the methods
themselves (\protect\hyperlink{ref-rudinStopExplainingBlack2019}{Rudin,
2019}). The most common example is the generalized additive model (GAM)
proposed in
\protect\hyperlink{ref-hastieGeneralizedAdditiveModels1986}{Hastie \&
Tibshirani}
(\protect\hyperlink{ref-hastieGeneralizedAdditiveModels1986}{1986}),
\protect\hyperlink{ref-hastieGeneralizedAdditiveModels1987}{Hastie \&
Tibshirani}
(\protect\hyperlink{ref-hastieGeneralizedAdditiveModels1987}{1987}), and
\protect\hyperlink{ref-hastieGeneralizedAdditiveModels1999}{Hastie \&
Tibshirani}
(\protect\hyperlink{ref-hastieGeneralizedAdditiveModels1999}{1999}). The
GAM estimates nonlinear functions of the covariates, which are
additively combined to produce predictions. Unless accounted for
explicitly, the model is therefore not able to capture dependencies
between two or more covariates.

Some examples of applications of the GAM framework to machine learning
algorithms exist. The nonlinear covariate-specific functions of the GAM
have been represented, for instance, by neural networks
(\protect\hyperlink{ref-lisboaEfficientEstimationGeneral2020}{Lisboa
\emph{et al.}, 2020};
\protect\hyperlink{ref-pottsGeneralizedAdditiveNeural1999}{Potts,
1999}), or by a random forest
(\protect\hyperlink{ref-caruanaIntelligibleModelsHealthCare2015}{Caruana
\emph{et al.}, 2015}). This results in an additive version of the
respective underlying machine learning algorithm, which is appealing due
to the inherent simplicity, but unsatisfactory when interaction effects
are expected to exist, or the researcher wishes to remain agnostic about
their existence.

The PENN model proposed in this paper modifies a neural network
architecture, to approximate local posterior parameter distributions,
and to generate interpretable outputs without imposing an additive
dependence structure. The approach is most closely related to
\protect\hyperlink{ref-al-shedivatContextualExplanationNetworks2017}{Al-Shedivat
\emph{et al.}}
(\protect\hyperlink{ref-al-shedivatContextualExplanationNetworks2017}{2017})
and
\protect\hyperlink{ref-melisRobustInterpretabilitySelfExplaining2018}{Melis
\& Jaakkola}
(\protect\hyperlink{ref-melisRobustInterpretabilitySelfExplaining2018}{2018}),
both of which propose conceptually related self-explaining frameworks.
The PENN can be seen as a variant of the self-explaining neural network
framework proposed by
\protect\hyperlink{ref-melisRobustInterpretabilitySelfExplaining2018}{Melis
\& Jaakkola}
(\protect\hyperlink{ref-melisRobustInterpretabilitySelfExplaining2018}{2018}),
but is distinct in a few important respects: (i) It utilizes Bayesian
inference techniques to produce parameter distributions, rather than
point estimates; (ii) The PENN is conceived specifically as an
econometric tool, where the emphasis of comparable studies has been on
computer vision tasks; (iii) The concepts of stability and
regularization are derived directly from a posterior distribution of the
parameters, differing from the gradient-regularized objective proposed
in
\protect\hyperlink{ref-melisRobustInterpretabilitySelfExplaining2018}{Melis
\& Jaakkola}
(\protect\hyperlink{ref-melisRobustInterpretabilitySelfExplaining2018}{2018}).
An implication of this final point is that, as an approach to model
explainability, the gradient-regularized self-explaining network ---
like the \emph{post hoc} algorithms described above --- embeds an
assumption of feature independence, while the PENN does not.

Econometric inference using machine learning models is in its infancy
with comparatively few examples in the literature. Those approaches
proposed to date either build on existing explainability algorithms
(primarily Shapley values)
(\protect\hyperlink{ref-lundbergUnifiedApproachInterpreting2017}{Lundberg
\& Lee, 2017};
\protect\hyperlink{ref-shapleyValueNPersonGames1953}{Shapley, 1953};
\protect\hyperlink{ref-strumbeljExplainingPredictionModels2014}{Štrumbelj
\& Kononenko, 2014}) or modify machine learning models directly, with
the PENN falling into the latter of these two categories. The former
group includes, for instance,
\protect\hyperlink{ref-josephShapleyRegressionsFramework2019a}{Joseph}
(\protect\hyperlink{ref-josephShapleyRegressionsFramework2019a}{2019}),
who proposes a framework to conduct statistical inference in machine
learning models using standard regression analysis with Shapley values
as inputs.
\protect\hyperlink{ref-brackeMachineLearningExplainability2019}{Bracke
\emph{et al.}}
(\protect\hyperlink{ref-brackeMachineLearningExplainability2019}{2019})
show how a Shapley-based algorithm developed in
\protect\hyperlink{ref-dattaAlgorithmicTransparencyQuantitative2016}{Datta
\emph{et al.}}
(\protect\hyperlink{ref-dattaAlgorithmicTransparencyQuantitative2016}{2016}),
which can account for some degree of feature dependence, can be used to
obtain a systematic analytical framework for explainability in financial
and econometric applications. The authors apply the algorithm to
determine key drivers of mortgage default. Finally,
\protect\hyperlink{ref-tiffinMachineLearningCausality2019}{Tiffin}
(\protect\hyperlink{ref-tiffinMachineLearningCausality2019}{2019})
demonstrates how Shapley values can be used to quantify the impact of
financial crises on growth.

Arguably the most prominent application of machine learning methods to
econometric inference is in post machine learning semiparametric
inference, typically for the calculation of average treatment effects
(\protect\hyperlink{ref-atheyApproximateResidualBalancing2018}{Athey
\emph{et al.}, 2018};
\protect\hyperlink{ref-belloniInferenceTreatmentEffects2014}{Belloni
\emph{et al.}, 2014},
\protect\hyperlink{ref-belloniProgramEvaluationCausal2017}{2017};
\protect\hyperlink{ref-chernozhukovDoubleDebiasedMachine2018}{Chernozhukov
\emph{et al.}, 2018};
\protect\hyperlink{ref-farrellRobustInferenceAverage2015}{Farrell,
2015}; \protect\hyperlink{ref-farrellDeepNeuralNetworks2021}{Farrell
\emph{et al.}, 2021}). This literature studies various methods of
obtaining valid causal inference on a static parameter (the average
treatment effect), when the first-stage machine learning method is
subject to regularization bias. The methods require notional
modifications to the underlying machine learning techniques, in the form
of partially linear designs.

Other approaches that modify machine learning algorithms directly
include
\protect\hyperlink{ref-wagerEstimationInferenceHeterogeneous2018}{Wager
\& Athey}
(\protect\hyperlink{ref-wagerEstimationInferenceHeterogeneous2018}{2018}),
\protect\hyperlink{ref-atheyGeneralizedRandomForests2019}{Athey \emph{et
al.}} (\protect\hyperlink{ref-atheyGeneralizedRandomForests2019}{2019})
and \protect\hyperlink{ref-friedbergLocalLinearForests2020}{Friedberg
\emph{et al.}}
(\protect\hyperlink{ref-friedbergLocalLinearForests2020}{2020}), who
propose random forest based algorithms to learn causal effects. The
authors introduce the concept of a causal forest to estimate
heterogeneous treatment effects.
\protect\hyperlink{ref-mullainathanMachineLearningApplied2017}{Mullainathan
\& Spiess}
(\protect\hyperlink{ref-mullainathanMachineLearningApplied2017}{2017})
propose constructing an explicit correspondence between an econometric
and a machine learning approach, by treating a decision tree like a
regression with multiple interaction terms. Interpretability frameworks
to test for nonlinear Granger causality have been put forward by
\protect\hyperlink{ref-tankInterpretableSparseNeural2018}{Tank \emph{et
al.}} (\protect\hyperlink{ref-tankInterpretableSparseNeural2018}{2018}),
\protect\hyperlink{ref-nautaCausalDiscoveryAttentionBased2019}{Nauta
\emph{et al.}}
(\protect\hyperlink{ref-nautaCausalDiscoveryAttentionBased2019}{2019}),
\protect\hyperlink{ref-wuDiscoveringNonlinearRelations2020}{Wu \emph{et
al.}}
(\protect\hyperlink{ref-wuDiscoveringNonlinearRelations2020}{2020}),
\protect\hyperlink{ref-loweAmortizedCausalDiscovery2020}{Löwe \emph{et
al.}} (\protect\hyperlink{ref-loweAmortizedCausalDiscovery2020}{2020}),
\protect\hyperlink{ref-khannaEconomyStatisticalRecurrent2020}{Khanna \&
Tan}
(\protect\hyperlink{ref-khannaEconomyStatisticalRecurrent2020}{2020})
and
\protect\hyperlink{ref-marcinkevicsInterpretableModelsGranger2021}{Marcinkevičs
\& Vogt}
(\protect\hyperlink{ref-marcinkevicsInterpretableModelsGranger2021}{2021}).
\protect\hyperlink{ref-marcinkevicsInterpretableModelsGranger2021}{Marcinkevičs
\& Vogt}
(\protect\hyperlink{ref-marcinkevicsInterpretableModelsGranger2021}{2021})
is noteworthy in that the authors employ a variant of a self-explaining
neural network architecture to examine Granger causality. Finally,
\protect\hyperlink{ref-horelSignificanceTestsNeural2020}{Horel \&
Giesecke}
(\protect\hyperlink{ref-horelSignificanceTestsNeural2020}{2020}) propose
a general method of conducting significance tests in a nonlinear setting
using neural networks, with an application to economic data. For further
detailed reviews of the role of machine learning in economics and
finance, the reader is referred to
\protect\hyperlink{ref-tiffinMachineLearningCausality2019}{Tiffin}
(\protect\hyperlink{ref-tiffinMachineLearningCausality2019}{2019}) and
\protect\hyperlink{ref-varianBigDataNew2014}{Varian}
(\protect\hyperlink{ref-varianBigDataNew2014}{2014}).

The PENN methodology introduced in the following sections differs from
most of the aforementioned approaches in that it is capable of capturing
heterogeneous effect structures, in the form of a locally parameterized
difference equation, without reducing the flexibility of the underlying
machine learning algorithm. Much of the current literature focusses on
methods to accommodate the black-box property of neural networks (by
approximating the gradient of a fitted model, or by correcting the bias
inherent to a semiparametric framework). The PENN instead aims to
circumvent the black-box property entirely, and to permit a degree of
statistical inference on the effects revealed by the DNN.

\hypertarget{parameter-encoder-neural-network}{%
\section{\texorpdfstring{Parameter encoder neural network
\label{method}}{Parameter encoder neural network }}\label{parameter-encoder-neural-network}}

\hypertarget{theoretical-foundations}{%
\subsection{Theoretical foundations}\label{theoretical-foundations}}

For a standard DNN, the objective is to learn a probabilistic function
\(p_{\boldsymbol{\theta}}(y|\boldsymbol{x})\), where \(y\) is the
dependent variable with \(y = \{y_i\}_{i = 1,...,N} \in \mathbb{R}\),
\(\boldsymbol{x}\) is a matrix of covariates with
\(\boldsymbol{x} = \{\boldsymbol{x}_i\}_{i = 1,...,N} \in \mathbb{R}_K\),
and \(\boldsymbol{\theta}\) is a vector of neural network weights. While
its multilayer structure coupled with a nonlinear activation function
can capture nonlinearity with a high degree of flexibility, the
predictions (\(\hat{y}\)) generated by the neural network are not
immediately interpretable. For regression tasks, \(\boldsymbol{\theta}\)
is optimized using a gradient descent algorithm, by minimizing the loss
in Eq. \ref{eq:nn_loss}
(\protect\hyperlink{ref-goodfellowDeepLearning2016}{Goodfellow \emph{et
al.}, 2016}):\footnote{A more detailed discussion of the structure of
  feed-forward neural networks is omitted here and has been covered
  extensively in the related literature.}

\begin{equation}
\mathcal{L}(\boldsymbol{x}, y) = - \sum_{i=1}^N \log p_{\boldsymbol{\theta}}(y_i|\boldsymbol{x}_i) \propto \sum_{i=1}^N(y_i - \hat{y}_i)^2.
\label{eq:nn_loss}
\end{equation}

In contrast to the above, econometric analysis typically employs some
form of a linear regression or classification framework, parameterizing
the problem with a vector of coefficients that results in the
conditional data likelihood \(p(y|\boldsymbol{\beta}, \boldsymbol{x})\),
which is maximized with respect to \(\boldsymbol{\beta}\). The
coefficients uniquely map predictions to covariates, and can --- subject
to a set of assumptions --- be treated as causal effects
(\protect\hyperlink{ref-hansenEconometrics2019}{Hansen, 2019}). The cost
of this interpretability is the supposition that the underlying data
generating process (DGP) is linear in \(\boldsymbol{\beta}\). A standard
Gaussian likelihood for regression tasks results in the loss function in
Eq. \ref{eq:lin_loss}:

\begin{equation}
\mathcal{L}(\boldsymbol{x}, y) = -\sum_{i=1}^N \log p(y_i|\boldsymbol{\beta}, \boldsymbol{x}_i) \propto \sum_{i=1}^N (y_i - \boldsymbol{x}_i' \boldsymbol{\beta})^2.
\label{eq:lin_loss}
\end{equation}

The PENN proposes a synthesis of the flexible neural network and the
interpretable linear regression, uniting both approaches in the context
of an encoder-decoder framework.\footnote{Encoder-decoder frameworks are
  neural network architectures that consist of two separate entities ---
  an encoder and a decoder ---, which are chained and trained together
  using a combined loss. Examples of encoder-decoder frameworks are
  autoencoders for data compression applications or sequence-to-sequence
  models, often used in natural language translation tasks
  (\protect\hyperlink{ref-goodfellowDeepLearning2016}{Goodfellow
  \emph{et al.}, 2016}).} The encoder is an inference network that
generates posterior densities for a vector of local regression
parameters \(\boldsymbol{\beta}_i\), \(i \in 1,...N\), and is denoted
\(q_{\boldsymbol{\theta}}(\boldsymbol{\beta}| \boldsymbol{x})\). The
decoder uses the posterior densities to form predictions over \(y_i\)
based on a parameterized likelihood. This framework retains the rich
flexibility of a DNN, but compels the neural network to encode
predictions via an interpretable locally linear model. Combining the
neural network with the linear likelihood function results in the
(conceptual) loss in Eq. \ref{eq:penn_loss_conceptual}, which represents
the expected linear likelihood with parameters generated by the
inference network:

\begin{equation}
\mathcal{L}(\boldsymbol{x}, y) = -\mathbb{E}_{\boldsymbol{\beta} \sim q_{\boldsymbol{\theta}}(\boldsymbol{\beta}|\boldsymbol{x})} \log p(y|\boldsymbol{\beta}, \boldsymbol{x}).
\label{eq:penn_loss_conceptual}
\end{equation}

The aim in this and subsequent sections is to convert Eq.
\ref{eq:penn_loss_conceptual} into a loss function that can be used to
train a neural network, and to obtain parameters for the locally linear
regression model.

Each local linear model, while interpretable, can be useful only insofar
as the coefficients are uniquely identified. A logical starting point
for the derivation of a loss function is therefore the unknown true
density distribution of the coefficients,
\(p(\boldsymbol{\beta}|y, \boldsymbol{x})\), which the encoder
\(q_{\boldsymbol{\theta}}(\boldsymbol{\beta}| \boldsymbol{x})\) aims to
approximate, such that
\(q_{\boldsymbol{\theta}}(\boldsymbol{\beta}| \boldsymbol{x}) \approx p(\boldsymbol{\beta}|y, \boldsymbol{x})\).
This manner of framing the objective is closely related to variational
inference problems, where the divergence between a latent and an
approximating density distribution is minimized.

Variational inference is a Bayesian technique of approximating the
posterior density of a latent variable, typically denoted \(z\), and has
become an important tool in several branches of machine learning
(\protect\hyperlink{ref-bleiVariationalInferenceReview2017}{Blei
\emph{et al.}, 2017};
\protect\hyperlink{ref-jordanIntroductionVariationalMethods1999}{Jordan
\emph{et al.}, 1999};
\protect\hyperlink{ref-zhangAdvancesVariationalInference2018}{Zhang
\emph{et al.}, 2018}). Its objective is to infer a posterior
distribution of model parameters given data. In contrast to Markov Chain
Monte Carlo (MCMC) sampling, variational inference converts the process
of obtaining a posterior from a sampling problem into an optimization
problem. The aim in variational inference is to determine a density,
\(q(z)\), that is as similar as possible to the posterior
\(p(z|\boldsymbol{x})\) --- i.e.~that minimizes the divergence between
the distributions. The approximate posterior \(q(z)\) is represented
using a parametric distribution whose parameters are optimized with the
objective of resembling \(p(z|\boldsymbol{x})\). For instance, the
latent variable could be assumed to follow a normal distribution, with
\(z\sim\mathcal{N}(\mu_z, \sigma_z^2)\), where the mean and variance
(\(\mu_z\) and \(\sigma_z^2\)) are optimized to ensure that
\(q(z) \approx p(z|\boldsymbol{x})\).

In the context of the PENN model, the latent variable \(z\) is replaced
by \(\boldsymbol{\beta}\),
\(q_{\boldsymbol{\theta}}(\boldsymbol{\beta}| \boldsymbol{x})\) is the
approximating function, and \(p(\boldsymbol{\beta}|y, \boldsymbol{x})\)
is the posterior. In order to make the problem tractable, it is common
to assume that the approximating function follows a mean field
variational distribution, with independently distributed latent
variables
(\protect\hyperlink{ref-bleiVariationalInferenceReview2017}{Blei
\emph{et al.}, 2017}), such that:

\begin{equation}
q_{\boldsymbol{\theta}}(\boldsymbol{\beta}| \boldsymbol{x}) = \prod_{i=1}^N q_{\boldsymbol{\theta}}(\boldsymbol{\beta}_i | \boldsymbol{x}_i).
\label{eq:mean_field}
\end{equation}

The vector \(\boldsymbol{\beta}_i\) is furthermore assumed to follow an
amortized multivariate Gaussian distribution, with

\begin{equation}
q_{\boldsymbol{\theta}}(\boldsymbol{\beta}_i | \boldsymbol{x}_i) = \mathcal{N}\left(\mu_{\boldsymbol{\theta}}(\boldsymbol{x}_i), \boldsymbol{\Sigma}_{\boldsymbol{x}_i}\right),
\label{eq:q_distribution}
\end{equation}

where
\(\boldsymbol{\Sigma}_{\boldsymbol{x}_i} = \text{Diag}\left(\sigma_{\boldsymbol{\theta}}^2(\boldsymbol{x}_i)\right)\).
The distribution is referred to as amortized, since it depends on shared
parameter functions \(\mu_{\boldsymbol{\theta}}(\cdot)\) and
\(\sigma_{\boldsymbol{\theta}}^2(\cdot)\), substantially reducing the
complexity of the problem. The functions take as input a vector
\(\boldsymbol{x}_i\) and infer from it parameters for the approximate
posterior of \(\boldsymbol{\beta}_i\). The parameter functions are
estimated using a DNN, referred to as an inference network, since it
infers parameters for the posterior based on the data. This removes the
need to parameterize \(N \times K\) distributions, and instead requires
a single neural network --- the inference network --- that predicts
\(K\) parameterizations given an arbitrary input.

This general methodology has found wide application in machine learning,
most prominently in the optimization of variational autoencoders (VAE)
(\protect\hyperlink{ref-kingmaAutoEncodingVariationalBayes2014}{Kingma
\& Welling, 2014};
\protect\hyperlink{ref-rezendeStochasticBackpropagationApproximate2014}{Rezende
\emph{et al.}, 2014}). The assumption of an amortized Gaussian mean
field variational distribution for the latent variables is also standard
in machine learning applications
(\protect\hyperlink{ref-zhangAdvancesVariationalInference2018}{Zhang
\emph{et al.}, 2018}). A contribution of the PENN is to interpret the
latent variables as the posteriors of local regression parameters and
training a supervised inference network using \(y\).

\hypertarget{objective-function-of-the-penn}{%
\subsection{Objective function of the
PENN}\label{objective-function-of-the-penn}}

The similarity between the approximating and posterior densities can be
captured using a Kullback-Leibler divergence, \(D_{KL}\), which measures
the expectation of the information difference between any two
distributions and is defined, for the general case, as:

\begin{equation}
D_{KL}(q(x) || p(x)) = \mathbb{E}_{q(x)} \left[ \log \frac{q(x)}{p(x)} \right] = - \int q(x) \log \frac{p(x)}{q(x)} dx \geq 0.
\end{equation}

Note that \(D_{KL}\) is not symmetrical
(i.e.~\(D_{KL}(q || p) \neq D_{KL}(p || q)\)). Reversing the arguments
can be more intuitive, however in the context of variational inference
--- where \(p\) is unknown ---, integrating over \(q\) is preferred
since it allows the expectation to be computed
(\protect\hyperlink{ref-zhangAdvancesVariationalInference2018}{Zhang
\emph{et al.}, 2018}). Several approaches exist to ``symmetrize''
\(D_{KL}\) (e.g.
\protect\hyperlink{ref-puAdversarialSymmetricVariational2017}{Pu
\emph{et al.}}
(\protect\hyperlink{ref-puAdversarialSymmetricVariational2017}{2017}),
\protect\hyperlink{ref-chenSymmetricVariationalAutoencoder2017}{Chen
\emph{et al.}}
(\protect\hyperlink{ref-chenSymmetricVariationalAutoencoder2017}{2017}),
and
\protect\hyperlink{ref-arjovskyPrincipledMethodsTraining2017}{Arjovsky
\& Bottou}
(\protect\hyperlink{ref-arjovskyPrincipledMethodsTraining2017}{2017})),
however, these are not explored here.

Substituting the approximated and true posteriors for any parameter
vector \(\boldsymbol{\beta}_{i} \in \boldsymbol{\beta}\) into a
multivariate \(D_{KL}\) (where \(i = 1,...,N\)) yields Eq. \ref{eq:kl}:

\begin{equation}
D_{KL}(q_{\boldsymbol{\theta}}(\boldsymbol{\beta}_{i}| \boldsymbol{x}_i) || p(\boldsymbol{\beta}_{i}|y_i, \boldsymbol{x}_i)) = - \int q_{\boldsymbol{\theta}}(\boldsymbol{\beta}_{i}| \boldsymbol{x}_i) \log \frac{p(\boldsymbol{\beta}_{i}|y_i, \boldsymbol{x}_i)}{q_{\boldsymbol{\theta}}(\boldsymbol{\beta}_{i}| \boldsymbol{x}_i)} d\boldsymbol{\beta}_{i} \geq 0.
\label{eq:kl}
\end{equation}

Given the distributional assumption in Eq. \ref{eq:q_distribution}, the
additive property implies that the multivariate \(D_{KL}\) in Eq.
\ref{eq:kl} is equal to the sum of the univariate Kullback-Leibler
divergences for the individual parameters
\(\beta_{ik} \in \boldsymbol{\beta}_i\). This property is used in
subsequent discussions to simplify the exposition and derivation of the
closed form solution.

A composite learning objective of the PENN can be defined by combining
the Kullback-Leibler divergence in Eq. \ref{eq:kl} with Eq.
\ref{eq:nn_loss}. The aim is to minimize the aggregate \(D_{KL}\), while
simultaneously maximizing the likelihood of the data,
\(p(y|\boldsymbol{x})\). Since
\(q_{\boldsymbol{\theta}}(\boldsymbol{\beta}| \boldsymbol{x})\) follows
a mean field variational distribution with independently distributed
parameters, the aggregate \(D_{KL}\) can be found by summing over \(N\),
leading to the objective captured in Eq. \ref{eq:objective}:

\begin{equation}
\min_{\boldsymbol{\theta}} \mathcal{L}(\boldsymbol{x}, y) = \min_{\boldsymbol{\theta}} \Big[ -\log p(y|\boldsymbol{x}) + \sum_{i=1}^N D_{KL}(q_{\theta}(\boldsymbol{\beta}_{i}| \boldsymbol{x}_i) || p(\boldsymbol{\beta}_{i}|y_i, \boldsymbol{x}_i)) \Big].
\label{eq:objective}
\end{equation}

Eq. \ref{eq:objective} is not yet a computable loss function since it
contains the unknown density
\(p(\boldsymbol{\beta}_{i}|y_i, \boldsymbol{x}_i)\). Instead,
Proposition \ref{prop:penn} operationalizes the estimation of the PENN
in a manner related to the VAE loss function, by applying Bayes' rule to
Eq. \ref{eq:kl} (see \ref{appa} for a proof):

\begin{proposition}
Where $\boldsymbol{\theta}$ is a vector of neural network weights, $\boldsymbol{\beta}$ is a matrix of local regression coefficients with $\boldsymbol{\beta}_{i}$ representing one coefficient vector, $q_{\boldsymbol{\theta}}(\boldsymbol{\beta}_{i}| \boldsymbol{x}_i)$ is an approximation function of the latent posterior density given by $p(\boldsymbol{\beta}_{i}|y_i, \boldsymbol{x}_i)$, and $p(\boldsymbol{\beta}_{i}|\boldsymbol{x}_i)$ is a conditional prior:

\begin{align}
\log p(y | \boldsymbol{x}) - &\sum_{i=1}^N D_{KL}(q_{\boldsymbol{\theta}}(\boldsymbol{\beta}_{i}| \boldsymbol{x}_i) || p(\boldsymbol{\beta}_{i}|y_i, \boldsymbol{x}_i)) = \nonumber \\
&\mathbb{E}_{\boldsymbol{\beta} \sim q_{\boldsymbol{\theta}}(\boldsymbol{\beta}| \boldsymbol{x})} \log p(y | \boldsymbol{\beta}, \boldsymbol{x}) - \sum_{i=1}^N  D_{KL}(q_{\boldsymbol{\theta}}(\boldsymbol{\beta}_{i}| \boldsymbol{x}_i) || p(\boldsymbol{\beta}_{i}|\boldsymbol{x}_i)).
\nonumber
\end{align}
\label{prop:penn}
\end{proposition}

Proposition \ref{prop:penn} warrants detailed examination. The LHS of
the equality is identical to the (negative) objective function of the
PENN in Eq. \ref{eq:objective}. The RHS is analogous to the evidence
lower bound (ELBO) in variational inference, which is equal to the
original Kullback-Leibler divergence in Eq. \ref{eq:kl} upto a constant
term --- \(\log p(y | \boldsymbol{x})\) (constant in the sense that it
does not depend on \(q_{\boldsymbol{\theta}}\)), and can therefore be
used as a proxy for optimizing \(D_{KL}\).

Dissecting the ELBO further, the first term encapsulates the
encoder-decoder framework (Eq. \ref{eq:penn_loss_conceptual}), and is
the expected parametric likelihood
\(\log p(y | \boldsymbol{\beta}, \boldsymbol{x})\) (decoder), with
\(\boldsymbol{\beta}\) generated by
\(q_{\boldsymbol{\theta}}(\boldsymbol{\beta}| \boldsymbol{x})\)
(encoder). The second term introduces a conditional prior on the
parameters, \(p(\boldsymbol{\beta}|\boldsymbol{x})\), with
\(D_{KL}(q_{\boldsymbol{\theta}}(\boldsymbol{\beta}_{i}|\boldsymbol{x}_i) || p(\boldsymbol{\beta}_{i}|\boldsymbol{x}_i))\)
acting as a regularization loss, which penalizes any solution that
diverges from a stable process. The role of the prior is discussed in
more detail below. The Kullback-Leibler divergence facilitates a form of
identification of the local parameter estimates, since it restricts
estimates of \(\boldsymbol{\beta}_i\) to the most stable --- in the
extreme case, globally static --- path, thus ensuring a solution that is
both unique and meaningful.

\hypertarget{loss-function-of-the-penn}{%
\subsection{Loss function of the PENN}\label{loss-function-of-the-penn}}

The RHS of the equality in Proposition \ref{prop:penn} can be converted
into a neural network loss function by making a few assumptions about
the shape of the likelihood, the coefficients and the prior. The
parametric log-likelihood
\(\log p(y|\boldsymbol{\beta}, \boldsymbol{x})\) can be defined using
any linear regression or classification model. For the applications in
this paper, the simple Gaussian regression is considered, with:

\begin{equation}
\log p(y | \boldsymbol{\beta}, \boldsymbol{x}) \propto -\sum_{i = 1}^{N}(y_i - \hat{y}_i)^2,
\end{equation}

where \(\hat{y}_i = \boldsymbol{x}_i'\boldsymbol{\beta}_i^m\), and
\(\boldsymbol{\beta}_i^m\) is a draw from the estimated posterior for
the \(i\)th observation. The prediction mean squared error is obtained
by computing the expectation over \(M\) Monte Carlo draws from the
parameter posterior:

\begin{equation}
\mathbb{E}_{\boldsymbol{\beta} \sim q_{\boldsymbol{\theta}}(\boldsymbol{\beta}| \boldsymbol{x})} \log p(y | \boldsymbol{\beta}, \boldsymbol{x}) \propto -\frac{1}{M}\sum_{m = 1}^M \sum_{i = 1}^{N} (y_i - \boldsymbol{x}_i' \boldsymbol{\beta}_i^m)^2.
\end{equation}

The second term in the loss function, the Kullback-Leibler loss, can be
computed in closed form given the Gaussian assumption for
\(\beta_{ik}\):

\begin{equation}
-D_{KL}(q_{\boldsymbol{\theta}}(\boldsymbol{\beta}_{i}| \boldsymbol{x}_i) || p(\boldsymbol{\beta}_{i}|\boldsymbol{x}_i)) = \sum_{k = 1}^K \left[ \log \Big(\frac{\sigma^q_{ik}}{\sigma^p_{ik}}\Big) - \frac{(\sigma^q_{ik})^2 + (\mu^q_{ik} - \mu^p_{ik})^2}{2(\sigma^p_{ik})^2} + \frac{1}{2} \right],
\label{eq:dkl}
\end{equation}

where \(\mu^p_{ik}\), \(\mu^q_{ik}\), \(\sigma^p_{ik}\) and
\(\sigma^q_{ik}\) parameterize the prior and posterior distributions,
such that

\begin{align}
&q_{\boldsymbol{\theta}}(\beta_{ik}| \boldsymbol{x}_i) = \mathcal{N}\left(\mu^q_{ik}, \left[\sigma^q_{ik}\right]^2\right), \text{ and} \\ &p(\beta_{ik}| \boldsymbol{x}_i) = \mathcal{N}\left(\mu^p_{ik}, \left[\sigma^p_{ik}\right]^2\right). 
\end{align}

The parameters \(\mu^q_{ik}\) and \(\sigma^q_{ik}\) are inferred by the
neural network given data at observation \(i\). The parameters of the
prior are discussed below. The derivation of Eq. \ref{eq:dkl} is
provided in \ref{appb}.

Finally, for the estimation of nonlinear regression parameters, it is
desirable to control the relative importance of the Kullback-Leibler
penalty within the loss function using a hyperparameter. This permits
the extent of nonlinearity encoded in the PENN to be determined in a
data-driven manner. For instance, when increasing the weight of the
prior, the parameters vary less over \(i\), converging to a static
solution as the weight goes to infinity. The trade-off between prior and
likelihood is achieved using the hyperparameter \(\lambda\).\footnote{This
  makes \(q_{\boldsymbol{\theta}}(\boldsymbol{\beta}| \boldsymbol{x})\)
  analogous to the encoder architecture of \(\beta\)-VAE networks, which
  introduce a hyperparameter weight on the prior of a VAE to encourage
  disentangled latent states
  (\protect\hyperlink{ref-higginsVVAELearningBasic2017}{Higgins \emph{et
  al.}, 2017}).}

Assembling the various components, the overall objective of the PENN is
given in Eq. \ref{eq:penn_loss}:

\begin{align}
\min_{\boldsymbol{\theta}} \mathcal{L}(\boldsymbol{x}, y) = \min_{\boldsymbol{\theta}} &\Bigg[ \frac{1}{M}\sum_{m = 1}^M \sum_{i = 1}^N (y_i - \boldsymbol{x}_i' \boldsymbol{\beta}_i^m)^2 + \nonumber \\
&\lambda \sum_{k=1}^K \sum_{i = 1}^N \left( \log \Big(\frac{\sigma^q_{ik}}{\sigma^p_{ik}}\Big) - \frac{(\sigma^q_{ik})^2 + (\mu^q_{ik} - \mu^p_{ik})^2}{2(\sigma^p_{ik})^2} + \frac{1}{2} \right)
\Bigg].
\label{eq:penn_loss}
\end{align}

The loss function in Eq. \ref{eq:penn_loss} is readily computable and
can be used to train a neural network --- in this case the inference
network of the PENN framework.

\hypertarget{kernel-weighted-prior}{%
\subsection{Kernel weighted prior}\label{kernel-weighted-prior}}

The prior \(p(\boldsymbol{\beta}|\boldsymbol{x})\) is not expressed in
absolute terms, but rather encodes an expected conditional relationship
with \(\boldsymbol{x}\). Given the underlying independent Gaussian
assumption, the prior density
\(p(\boldsymbol{\beta}_i|\boldsymbol{x}_i)\) is fully described by its
conditonal moments \(\boldsymbol{\mu}_i^p|\boldsymbol{x}_i\) and
\(\boldsymbol{\sigma}_i^p|\boldsymbol{x}_i\), for any vector of
parameters \(\boldsymbol{\beta}_i\). An important function of the prior
is to induce stability in the estimates, shrinking the conditional
moments of the posterior towards static parameters in the neighborhood
of \(\boldsymbol{x}_i\). With an appropriate definition of the prior
moments, the inference network can be trained to infer posterior
parameterizations in adherence to ``stability rules'', ensuring that
similar neighborhoods in \(\boldsymbol{x}\) result in similar
posteriors. Or, expressed differently, the gradient
\(\triangledown_{\boldsymbol{x}} \boldsymbol{\beta}_i \approx \boldsymbol{0}\)
(the gradient of the parameters with respect to the inputs approximates
zero).

A parameter gradient approximating zero leads naturally to interpretable
parameters that resemble local marginal effects, as shown (for the
independent features case) in Proposition \ref{prop:prior}:

\begin{proposition}
Let $\triangledown_{\boldsymbol{x}} \boldsymbol{\beta}_i$ be a matrix of partial derivatives of the inference network $q_{\boldsymbol{\theta}}(\boldsymbol{\beta}_i|\boldsymbol{x}_i)$, with respect to independent input features. Furthermore, let $f(\boldsymbol{x})$ describe a PENN architecture that generates predictions, such that $f_m(\boldsymbol{x}_i) = \boldsymbol{x}_i'\boldsymbol{\beta}_i^m$, where $\boldsymbol{\beta}_i^m$ is the $m$th sample from the parameter posterior, with $m \in 1,...,M$ and $i\in 1,...,N$. The gradient of $f_m(\boldsymbol{x}_i)$ with respect to the inputs is denoted $\triangledown_{\boldsymbol{x}}f_m(\boldsymbol{x}_i)$.

Now, when $\triangledown_{\boldsymbol{x}} \boldsymbol{\beta}_i = \boldsymbol{0}$, it holds that $\boldsymbol{\beta}_i^m = \triangledown_{\boldsymbol{x}}f_m(\boldsymbol{x}_i)$.
\label{prop:prior}
\end{proposition}

The proof of Proposition \ref{prop:prior}, as well as a discussion of
the dependent features case, is provided in \ref{app:prop_prior}.

Apart from suggesting that stable parameters are indeed interpretable,
Proposition \ref{prop:prior} also implies an approach to their
estimation. By constructing a loss function that penalizes the deviation
of the parameters from backpropagated network gradients (i.e.~penalizing
the deviation from \(\triangledown_{\boldsymbol{x}}f(\boldsymbol{x})\)),
parameter estimates are encouraged to act as locally stable
gradients.\footnote{\protect\hyperlink{ref-melisRobustInterpretabilitySelfExplaining2018}{Melis
  \& Jaakkola}
  (\protect\hyperlink{ref-melisRobustInterpretabilitySelfExplaining2018}{2018})
  present such an approach, in the form of the gradient-regularized
  self-explaining neural network (SENN).} However, since the
backpropagated gradient is composed of partial derivatives of
\(f(\boldsymbol{x})\) with respect to \(\boldsymbol{x}\), the approach
produces biased estimates when the assumption of feature independence
fails. The point is illustrated using the example of the
gradient-regularized self-explaining network (SENN) of
\protect\hyperlink{ref-melisRobustInterpretabilitySelfExplaining2018}{Melis
\& Jaakkola}
(\protect\hyperlink{ref-melisRobustInterpretabilitySelfExplaining2018}{2018})
in Section \ref{simulation}, and in \ref{app:prop_prior}. Given its
implicit assumption of feature independence, the explanations generated
by a SENN are closely related to those produced by popular \emph{post
hoc} algorithms such as SHAP or LIME, which also assume feature
independence.

In the case of the PENN, \(p(\boldsymbol{\beta}|\boldsymbol{x})\) is
instead obtained using multivariate predictions of the prior conditional
moments, denoted \(\boldsymbol{\hat{\mu}}_i^p\) and
\(\boldsymbol{\hat{\sigma}}_i^p\), such that

\begin{equation}
p(\boldsymbol{\beta}_i|\boldsymbol{x}_i)=
\mathcal{N}\left(\boldsymbol{\hat{\mu}}_{i}^p, \boldsymbol{\hat{\Sigma}}^p_i \right), \;\;\; \boldsymbol{\hat{\Sigma}}^p_i = \text{Diag}\left( [\boldsymbol{\hat{\sigma}}_{i}^p]^2\right).
\end{equation}

Any estimator of the prior conditional moments should satisfy three
properties: (i) The estimator should induce stability, with
\(\lim_{\lambda\rightarrow\infty}\triangledown_{\boldsymbol{x}} \boldsymbol{\beta}_i = \boldsymbol{0} \;\forall \; i\)
in the empirical joint data distribution. (ii) The estimator should be
nonparametric. By using an estimator with minimal distributional
assumptions, the more rigorous assumptions required by other
explainability algorithms, such as feature independence, can be avoided.
(iii) The extent of regularization should be variable. The estimator
should accommodate shrinkage towards global static parameters, by
allowing the degree of prior support to vary based on a hyperparameter.

A stable and nonparametric method that satisfies the above criteria and
captures the relational character of the prior is the \(k\)-nearest
neighbors estimator. Formally, the prior is a function that computes
expected moments of
\(q_{\boldsymbol{\theta}}(\boldsymbol{\beta}|\boldsymbol{x})\) over any
neighborhood, \(\mathcal{I}\), of points in \(\boldsymbol{x}\), such
that:

\begin{align}
p(\boldsymbol{\beta}_{i}|\boldsymbol{x}_i) = \mathcal{N}&\left(\mathbb{E}_{i\in\mathcal{I}}[\boldsymbol{\mu}_i^q], \mathbb{E}_{i\in\mathcal{I}}[\boldsymbol{\Sigma}^q_{i}]\right) \;\;\; \forall \;\;\; i \in \mathcal{I} \nonumber \\
&\;\;\; \text{where $\mathcal{I}$ is defined such that} \;\;\; ||\boldsymbol{x}_i - \boldsymbol{x}_j|| < D \;\;\; \forall \; i,j \in \mathcal{I},
\label{eq:cluster_condition}
\end{align}

and \(\mathcal{I}\subset [1,N]\) indexes a neighborhood of vectors in
\(\boldsymbol{x}\) that lie in close mutual proximity. Here
\(\boldsymbol{\Sigma}^q_{i} = \text{Diag}\left([\boldsymbol{\sigma}_i^q]^2\right)\).
The expectation is computed using an \(N\times N\) kernel weighting
matrix, \(\boldsymbol{\pi}_{\boldsymbol{x};D}\), that indexes blocks of
neighboring points in \(\boldsymbol{x}\), and whose rows sum to unity.
This is closely related to conventional kernel weighting methods,
particularly compact support kernels, with positive weights over a
region of neighboring points. The definition of \(\mathcal{I}\) in Eq.
\ref{eq:cluster_condition} results in disjoint neighborhoods (defined
based on the distance threshold \(D\)) that can be interpreted as
parameter regimes. This custom compact support kernel is found to permit
more nuanced control over the support range (and hence over the extent
and manner of regularization in the inference network). A comparison to
more traditional compact support kernels is provided in
\ref{app:kernel}.

Given \(\boldsymbol{\pi}_{\boldsymbol{x};D}\), the prior moments are
obtained from
\(q_{\boldsymbol{\theta}}(\boldsymbol{\beta}| \boldsymbol{x})\), as

\[
\boldsymbol{\hat{\mu}}^p = \boldsymbol{\pi}_{\boldsymbol{x};D}\boldsymbol{\mu}^q \;\;\; \text{and}\;\;\; \boldsymbol{\hat{\sigma}}^p = \boldsymbol{\pi}_{\boldsymbol{x};D}\boldsymbol{\sigma}^q,
\]

where \(\boldsymbol{\mu}^q\) and \(\boldsymbol{\sigma}^q\) are generated
in the inference network. When \(D\) is large, the condition in Eq.
\ref{eq:cluster_condition} is satisfied for all \(i\) and \(j\),
resulting in a single (static) prior parameter vector. As \(D\)
decreases, the number of regimes encoded in the prior increases. When
\(D=0\), \(\boldsymbol{\pi}_{\boldsymbol{x};D}\) collapses to an
identity matrix, with the regularization loss equal to zero.

Apart from ensuring that the gradient
\(\triangledown_{\boldsymbol{x}} \boldsymbol{\beta}_i \approx \boldsymbol{0}\),
this definition of the prior has the intuitive appeal that the broader a
neighborhood in \(\boldsymbol{x}\) is defined, the more the posterior is
shrunken towards a globally static parameter value, with --- in the
extreme case ---
\(\boldsymbol{\beta}_i = \boldsymbol{\beta}^* \; \forall \; i \in 1,...,N\),
where \(\boldsymbol{\beta}^*\) is the least-squares optimal static
parameter vector. In addition, regularizing the overall gradient of the
inference network with respect to its inputs results in a network that
is less sensitive to the chosen topology, and is therefore simpler to
train.\footnote{Gradient regularization has played an important role
  recently in improving the adversarial robustness of neural networks
  (i.e.~the extent to which small perturbations can lead to large
  changes in prediction). A good overview of current research is
  provided by
  \protect\hyperlink{ref-finlayScaleableInputGradient2021}{Finlay \&
  Oberman}
  (\protect\hyperlink{ref-finlayScaleableInputGradient2021}{2021}).} By
constraining nonlinearity to a stable range of solutions that are
meaningful for purposes of interpretation, the regression parameters
\(\boldsymbol{\beta}_i\) can be uniquely identified in the inference
network. The prior permits a single unique solution to exist, which
maximizes model fit under the condition that similar input vectors lead
to a similarly parameterized posterior.

An equivalent, but more intuitive alternative to the scalar \(D\), is to
define the number of separate neighborhoods in \(\boldsymbol{x}\) over
which expectations are computed. The normalized number of neighborhoods
is denoted \(\delta\), where \(\delta = 0\) is equivalent to a single
regime (\(D=\infty\)), and \(\delta = 1\) is equivalent to \(N\) regimes
(\(D=0\)). \(\delta\) is treated as a second hyperparameter in the PENN
model (in addition to \(\lambda\)), and governs the number of distinct
parameter regimes (or neighborhoods) encoded in the prior.

The optimal composition of neighborhoods can conveniently be obtained
using complete linkage agglomerative clustering (see
\protect\hyperlink{ref-kaufmanFindingGroupsData2005}{Kaufman \&
Rousseeuw} (\protect\hyperlink{ref-kaufmanFindingGroupsData2005}{2005})
for an overview). Complete linkage clustering begins by placing each
observation vector into a cluster of its own, and merging the clusters
with the minimum cluster distance, \(D_{\mathcal{I}\mathcal{J}}\), where
\(\mathcal{I}\) and \(\mathcal{J}\) denote clusters. Clusters are merged
until exactly \(1 + \delta(N-1)\) clusters remain. The cluster distance
is formally defined as the largest distance between any two sample
vectors in the respective clusters, such that:

\begin{equation}
D_{\mathcal{I}\mathcal{J}} = \max \{d_{ij}\}_{i\in \mathcal{I},j\in \mathcal{J}},\;\;\; \text{where} \;d_{ij}=||\boldsymbol{x}_i-\boldsymbol{x}_j||_2.
\end{equation}

Complete linkage clustering ensures that for any value of \(\delta\),
the distance threshold \(D\) that satisfies the condition \(d_{ij}<D\),
\(i,j\in \mathcal{I}\) for all neighborhoods, is minimized. Thus, there
is no cluster arrangement that meets the condition set forth in Eq.
\ref{eq:cluster_condition} with a smaller number of clusters. Treating
\(\delta\) as a hyperparameter in conjunction with complete linkage
agglomerative clustering is therefore equivalent to the use of \(D\),
with \(\delta\) representing a choice that is at once more intuitive and
can easily be combined with standard software implementations of
hierarchical clustering algorithms.

\hypertarget{exploring-the-role-of-hyperparameters}{%
\subsection{\texorpdfstring{Exploring the role of hyperparameters
\label{role_of_hyperparameters}}{Exploring the role of hyperparameters }}\label{exploring-the-role-of-hyperparameters}}

The interplay between the hyperparameters \(\lambda\) and \(\delta\)
permits a great deal of flexibility in controlling the strength of
shrinkage of the coefficients. Where \(\lambda\) governs the overall
weight of the prior in relation to the training mean squared error,
\(\delta\) controls the nuance in the nonlinear patterns captured by the
model. If \(\delta = (N-1)^{-1}\) (resulting exactly two neighborhoods
in \(\boldsymbol{x}\)), the parameters are shrunken towards a two-regime
representation. As \(\delta\) increases, so does the number of regimes.
\(\delta\) can therefore be described as defining the resolution of the
static estimates. At maximum resolution (\(\delta=1\)), the problem is
unregularized.

Increasing \(\lambda\) limits the extent to which parameters can vary
around the regime-specific static parameter values. As \(\lambda\) grows
very large, parameters become static, with the number of static
parameters driven by \(\delta\). When \(\lambda = 0\), the problem is
unregularized.

Table \ref{tbl:hyperparameters} summarizes the effects of the
hyperparameters on the PENN estimates:

\begin{table}
\centering
\begin{tabular}{ | m{1.5cm} | m{4.7cm}| m{4.7cm} | } 
\hline
& $\lambda = 0$ & $\lambda \rightarrow \infty$ \\
\hline
$\delta \rightarrow 1$ & No shrinkage & Multiple parameter regimes \\ 
\hline
$\delta = 0$ & No shrinkage & Static parameters \\ 
\hline
\end{tabular}
\caption{Different combinations of hyperparameters of the PENN and the resulting regularization profile.}
\label{tbl:hyperparameters}
\end{table}

\begin{figure}
\centering
\includegraphics{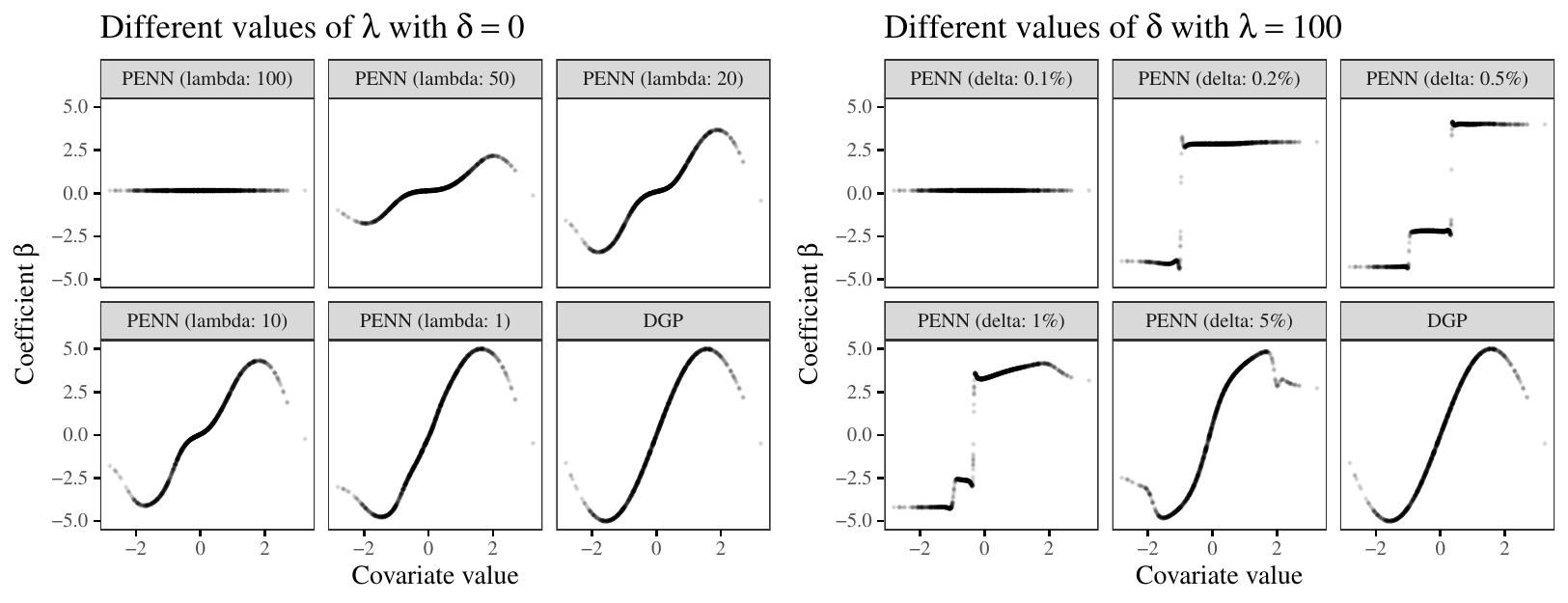}
\caption{Effect of hyperparameters \(\lambda\) (left panel) and
\(\delta\) (right panel). In each case the respective other
hyperparameter is held constant at \(\delta = 0\) and \(\lambda = 100\).
The true coefficient function is plotted in the bottom-right panel
(DGP). \label{fig:sim_hyperparameters}}
\end{figure}

The simulations introduced in Section \ref{simulation} provide a good
setting within which to explore the role of the hyperparameters
\(\lambda\) and \(\delta\). Fig. \ref{fig:sim_hyperparameters} uses the
example of the coefficient function \(\mathcal{B}_1\) in the simulation
(refer to Section \ref{simulation} for a complete description of the
simulated DGP), and plots the estimated coefficients against the
covariate \(x_1\).\footnote{The coefficients are estimated by simulating
  data from a DGP identical with that described in Eq. \ref{eq:dgp}, but
  excluding the covariates \(x_2\) and \(x_3\).} The left panel shows
the effect of decreasing \(\lambda\) while holding \(\delta\) constant
at \(\delta = 0\). The coefficients begin at the static solution, and
gradually increase the extent of nonlinearity to resemble the DGP more
closely. The right panel plots the effect of \(\delta\), where
\(\lambda\) is held constant at a high value of \(\lambda = 100\). As
the number of individual regimes is increased, the coefficient estimates
increase in resolution, starting with a single regime (static
coefficient), and increasing gradually to resemble the DGP as more
regimes are included.

The parameter prior enables the inference network to learn a set of
relational rules governing the manner in which it infers posterior
parameterizations from the input data. When optimal hyperparameters are
very restrictive, the resulting rules dictate identical parameters
inferred from all input vectors (the static solution). Conversely, as
the support for nonlinearity increases, the nature of the rules can
become highly complex, permitting a nuanced trade-off between model fit
and parameter stability. Since the rules are encoded in the weights of
the inference network, out-of-sample parameter inference becomes a
simple matter of predicting posterior parameterizations using an
appropriately trained network.

\hypertarget{network-architecture}{%
\subsection{\texorpdfstring{Network architecture
\label{method:network_architecture}}{Network architecture }}\label{network-architecture}}

Having derived a loss function, the following section proposes a network
architecture for the PENN. Significant discretion exists in the choice
of the network topology for the inference network (e.g.~the number of
hidden layers, node types etc.). The layout presented below utilizes a
standard feed forward network with two hidden layers, which is
appropriate for both the simulated and empirical applications included
in this paper. However, the type of architecture used should be chosen
to suit the specific application.

Fig. \ref{fig:network} depicts the network topology using a simplified
flow diagram:

\begin{figure}[ht]
\centering
\includegraphics[width=10cm,keepaspectratio]{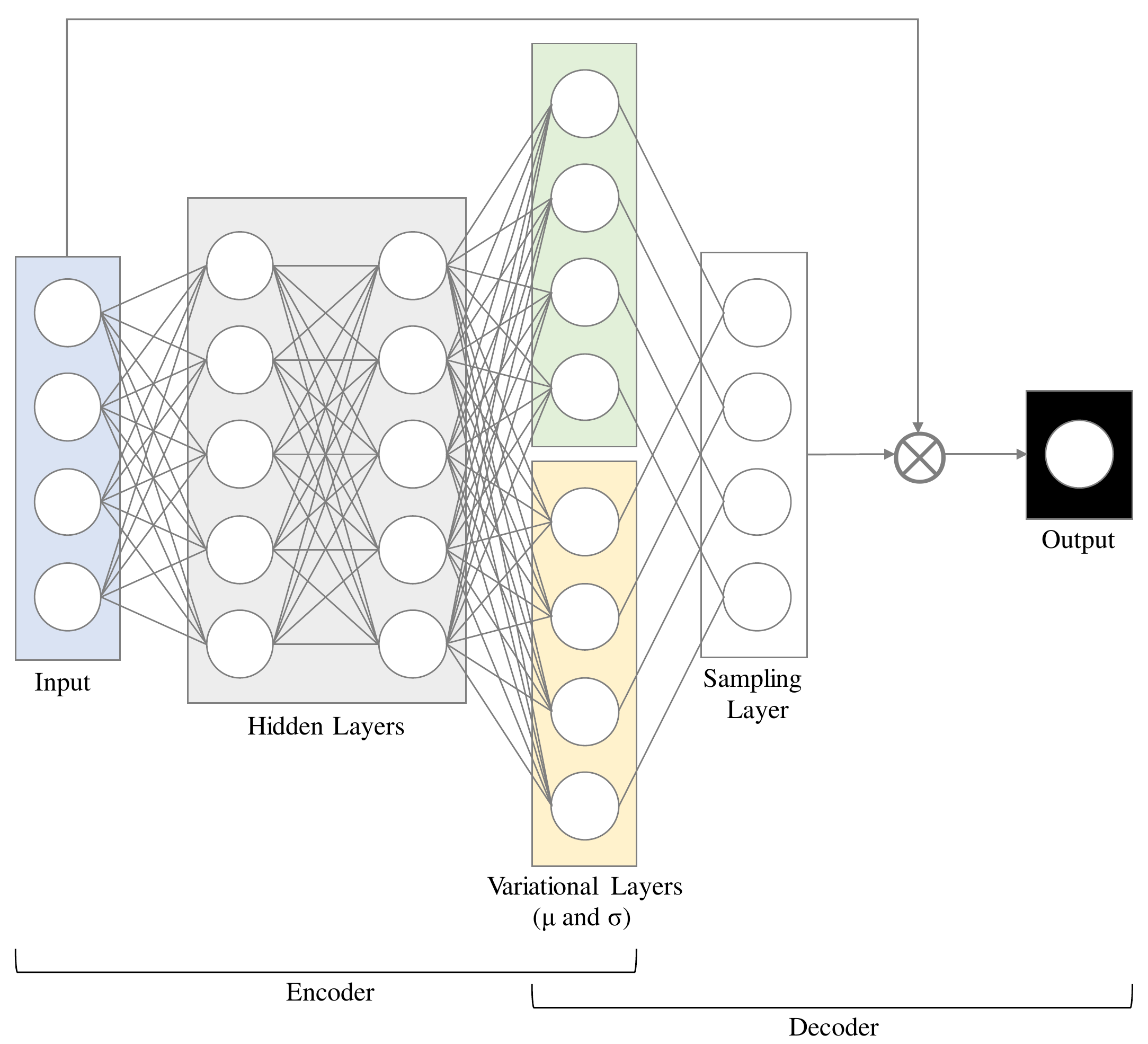}
\caption{PENN network architecture. Nodes are represented by circles. Nodes with trainable input weights also include a bias.}
\label{fig:network}
\end{figure}

The network consists of five components. Standardized input features are
passed to a stack of fully-connected feed forward hidden layers. The
sigmoid activation function is used to introduce nonlinearity within the
hidden layers.\footnote{Recently, the rectified linear unit (ReLU) has
  gained popularity as an activation function --- particularly for
  classification problems
  (\protect\hyperlink{ref-farrellDeepNeuralNetworks2021}{Farrell
  \emph{et al.}, 2021}). Convergence in the PENN can be achieved with
  both sigmoid and ReLU functions, and a comparison is omitted since no
  indication of a systematic difference in performance between the
  methods was found.} The hidden layers feed into two variational layers
of equal dimension as the inputs, which learn \(\boldsymbol{\mu}^q\) and
\(\boldsymbol{\sigma}^q\), respectively. The purpose of the variational
layers is to infer parameters (mean and variance) for the distribution
\(q_{\boldsymbol{\theta}}(\boldsymbol{\beta}| \boldsymbol{x})\),
permitting direct sampling from the approximated posterior. The variance
layer utilizes an exponential activation function to ensure that the
output is limited to \(\mathbb{R}_{\geq0}\).

Since it is not possible to backpropagate the gradient through a
stochastic process, the so-called ``reparameterization trick'' is used
in the sampling layer to connect the encoder and decoder sections of the
PENN (see
\protect\hyperlink{ref-kingmaAutoEncodingVariationalBayes2014}{Kingma \&
Welling}
(\protect\hyperlink{ref-kingmaAutoEncodingVariationalBayes2014}{2014})).
During each training epoch, \(M\) random samples are drawn from a
multivariate standard normal distribution, such that each sample
\(\boldsymbol{s}^m \sim \mathcal{N}(\boldsymbol{0}, \boldsymbol{I}_K), \; m = 1,...,M\),
where \(\boldsymbol{I}_K\) is a \(K\times K\) dimensional identity
matrix. The variational layers are subsequently used to generate the
corresponding \(M\) Monte Carlo samples from the approximate posterior,
with
\(\boldsymbol{\beta}^m = \boldsymbol{\mu}^q + \boldsymbol{\sigma}^q \odot \boldsymbol{s}^m\)
(where \(\odot\) is element-wise multiplication). Note that the sampling
layer therefore consists of a three dimensional tensor with dimensions
\(N\times M \times K\).

Finally, the samples from the parameter posterior,
\(\boldsymbol{\beta}^m, \; m = 1,...,M\), feed into a non-trainable
output layer that simply calculates the dot product
\(\hat{y}_{i}^m = \boldsymbol{x}_i'\boldsymbol{\beta}_{i}^m\), to
generate \(M\) draws from the predictive density. The input layer is an
auxiliary input to the final layer, which requires both
\(\boldsymbol{\beta}^m\) and \(\boldsymbol{x}\), and the output layer is
again a three dimensional tensor with dimensions \(N\times M\times 1\).

In the applications in Sections \ref{simulation} and \ref{application},
the PENN is implemented using \texttt{keras} and \texttt{tensorflow}
frameworks
(\protect\hyperlink{ref-allaireKerasInterfaceKeras2020}{Allaire \&
Chollet, 2020};
\protect\hyperlink{ref-allaireTensorflowInterfaceTensorFlow2019}{Allaire
\& Tang, 2019}). Models are trained using the popular adaptive moment
(Adam) algorithm introduced in
\protect\hyperlink{ref-kingmaAdamMethodStochastic2017}{Kingma \& Ba}
(\protect\hyperlink{ref-kingmaAdamMethodStochastic2017}{2017}), which is
a stochastic gradient descent algorithm with adaptive learning rates.
The number of Monte Carlo draws \(M\) is set to 100 across all
applications. The size of the hidden layers is determined individually
for each application, with larger specifications preferred, in order to
permit a high capacity for nonlinearity in the network. A more
comprehensive overview of the neural network hyperparameters is provided
in \ref{appc}.

\hypertarget{extensions}{%
\subsection{Extensions}\label{extensions}}

A useful variation on the standard PENN architecture in econometric or
financial applications may be to constrain some parameters to be static.
This is desirable, for instance, when the data availability is limited
and a full neural network is impractical, or when the DGP is assumed to
be partially linear. Static parameters are easily accommodated in the
framework, by removing connections between the nodes associated with
linear parameters in the variational layers, and the hidden layer. By
removing the connections between variational and hidden layers, only the
bias term is retained, resulting in a posterior with static mean and
variance. Fig. \ref{fig:network_lin} visualizes the partially linear
network:

\begin{figure}[ht]
\centering
\includegraphics[width=10cm,keepaspectratio]{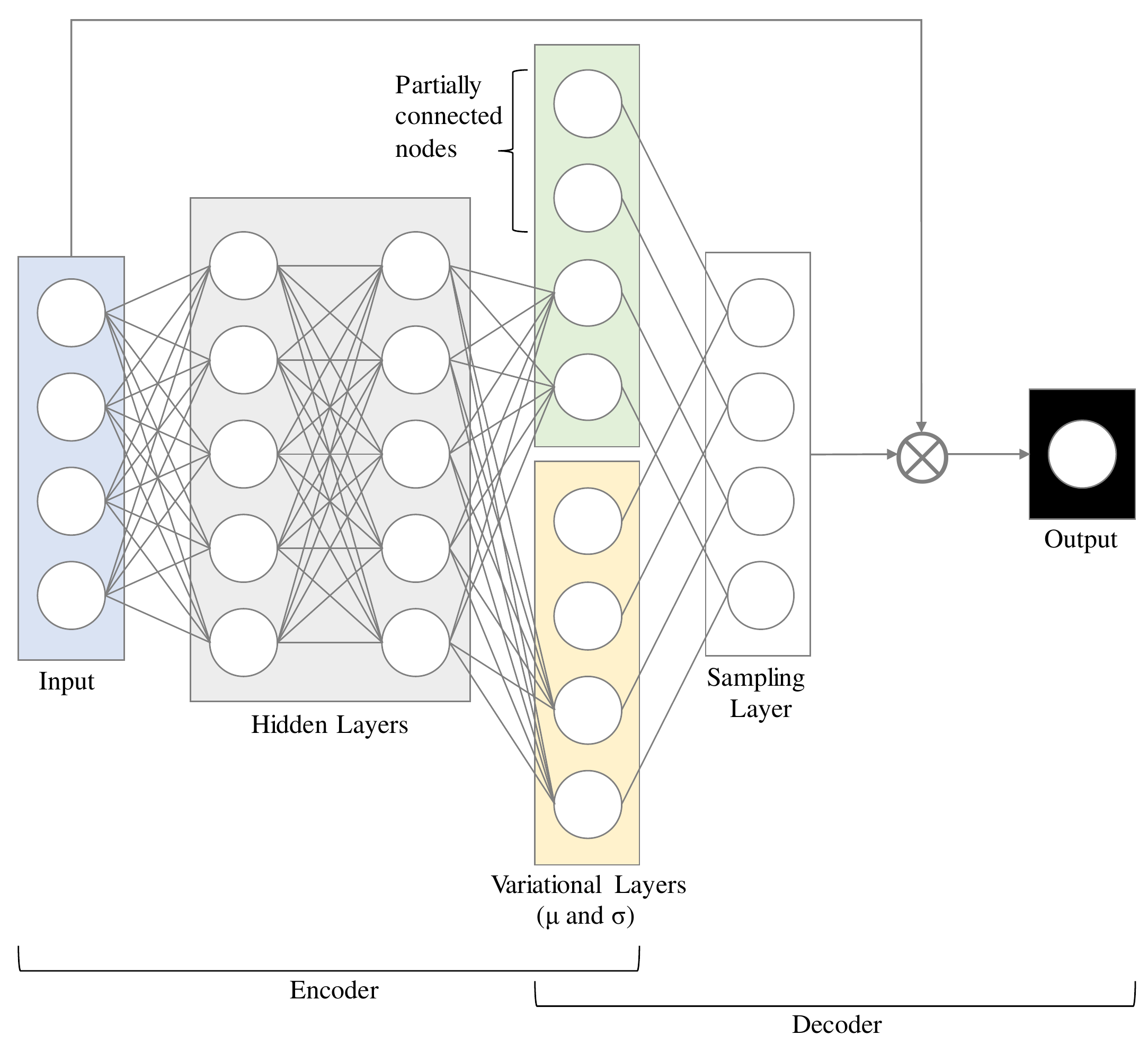}
\caption{PENN network architecture with partially static parameters. Nodes are represented by circles. Nodes with trainable input weights also include a bias.}
\label{fig:network_lin}
\end{figure}

\begin{figure}[ht]
\centering
\includegraphics[width=10cm,keepaspectratio]{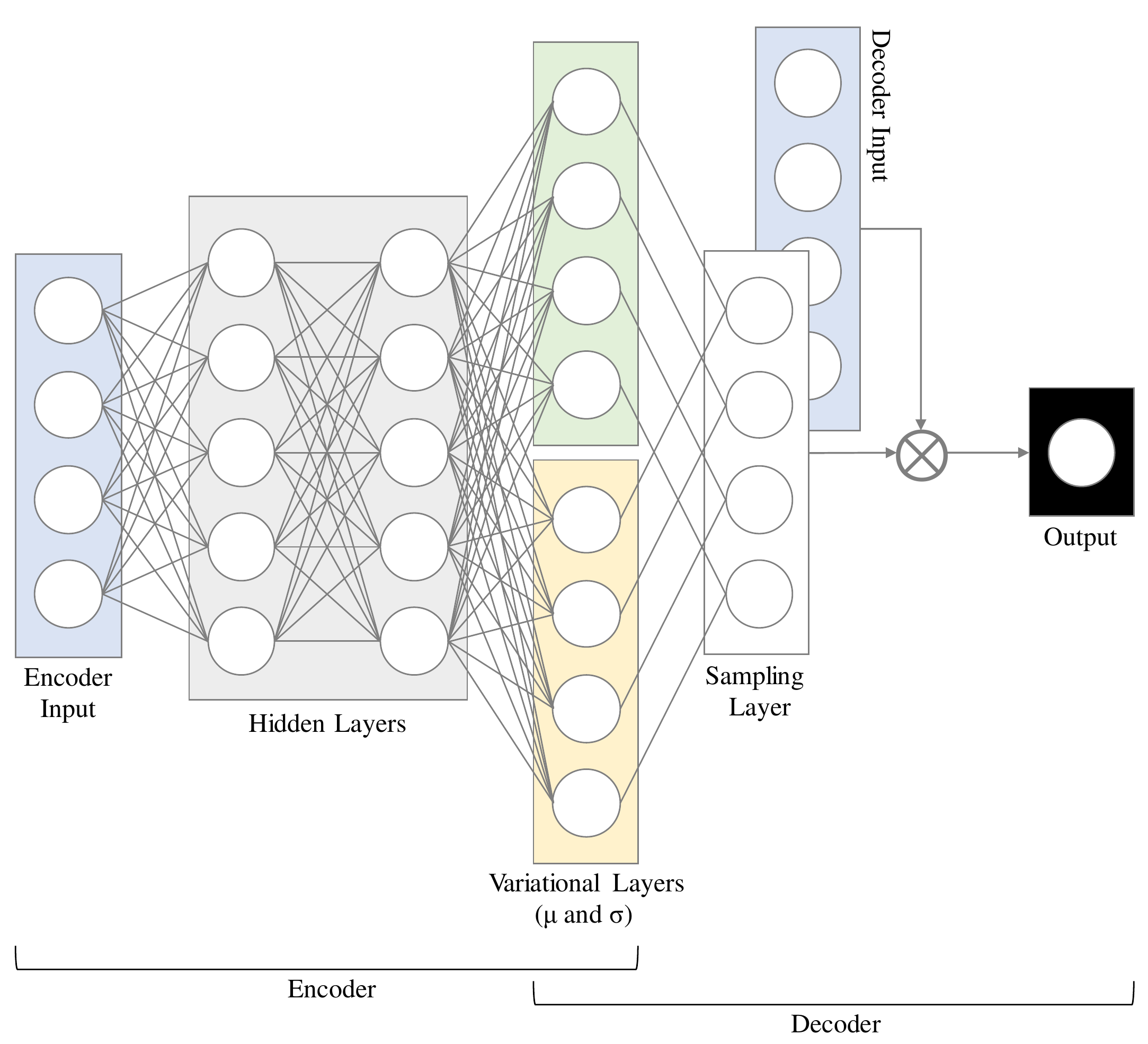}
\caption{PENN network architecture with distinct input data sets for the encoder and decoder. Nodes are represented by circles. Nodes with trainable input weights also include a bias.}
\label{fig:network_full}
\end{figure}

More generally, the input data to the inference network does not by
necessity have to correspond to the variables included in the model
specification and for which parameter posteriors are estimated. Letting
the model of interest be given by
\(\hat{y}_i = \boldsymbol{x}_i'\boldsymbol{\beta}_i\), the parameters
\(\boldsymbol{\beta}_i\) could be inferred using
\(q_{\boldsymbol{\theta}}(\boldsymbol{\beta}| \boldsymbol{z})\), where
\(\boldsymbol{z}\) differs from \(\boldsymbol{x}\), as demonstrated in
Fig. \ref{fig:network_full}.

This network architecture allows nonlinear dynamics to draw on a
distinct information set (e.g.~a larger set of variables capturing the
macroeconomic state), while parameters are estimated for a smaller model
(e.g.~a model motivated by theoretical considerations). Section
\ref{application} demonstrates how this setup can be applied in an
empirical setting.

Finally, a local mean can be included in the PENN, by adding a column of
ones to \(\boldsymbol{x}\). A global mean is computed by holding the
local mean constant using the architecture described in Fig.
\ref{fig:network_lin}, or simply by adding a trainable bias to the
decoder.

\hypertarget{simulations}{%
\section{\texorpdfstring{Simulations
\label{simulation}}{Simulations }}\label{simulations}}

\hypertarget{setup}{%
\subsection{Setup}\label{setup}}

The role of the PENN as an explainability method in the presence of a
nonlinear and non-additive DGP can be demonstrated using a simple
simulation. Consider a DGP consisting of three covariates,
\(\boldsymbol{x} = \begin{bmatrix} x_1&x_2&x_3 \end{bmatrix}\), a
dependent variable, \(y\), and an error term,
\(\epsilon \sim \mathcal{N}(0, \sigma^2_{\epsilon})\):

\begin{equation}
y_i = \mathcal{B}_{1}(x_{i1}) \cdot x_{i1} + 0 \cdot x_{i2} + \mathcal{B}_{3}(x_{i2}) \cdot x_{i3} + \epsilon_i \;\;\; i\in 1,...,N.
\label{eq:dgp}
\end{equation}

The marginal effects are nonlinear and denoted by the coefficient
functions \(\mathcal{B}_k(x_{ij})\). For the first covariate, the
coefficient follows a sine curve, with
\(\mathcal{B}_1(x_{i1}) = 5 \sin x_{i1}\). The second and third
covariates exhibit a simple interaction, with
\(\mathcal{B}_3(x_{i2}) = \tau(x_{i2})\). Notice that \(x_2\) has no
effect on \(y\) directly, but influences the output via a threshold
function \(\tau(x_2)\), such that

\begin{equation}
\tau(x_2) = \begin{cases}
      5 & \text{ if } x_2 > 0.5 \\
      0 & \text{ if } -0.5\leq x_2\leq 0.5 \\
      -5 & \text{ if } x_2 < -0.5
   \end{cases}.
\nonumber
\end{equation}

The first marginal effect function, \(\mathcal{B}_1(x_1)\), evaluates
the method's ability to capture arbitrary nonlinearity. Conversely,
\(\mathcal{B}_3(x_2)\) tests whether the PENN is capable of identifying
the true effect in the presence of an interaction. The latter task is
challenging particularly for \emph{post hoc} explainability algorithms
that assume feature independence. Such an approach usually fails to
disentangle the indirect effect of \(x_2\) from the effect of \(x_3\).

Finally, the covariates are independent samples from a correlated
multivariate normal distribution,
\(\boldsymbol{x} \sim \mathcal{N}(\boldsymbol{0}, \boldsymbol{\Sigma})\),
where

\begin{equation}
\boldsymbol{\Sigma} = \begin{bmatrix} 1 & \rho & \rho \\ \rho & 1 & \rho \\ \rho & \rho & 1 \end{bmatrix}.
\nonumber
\end{equation}

\hypertarget{benchmark-methods}{%
\subsection{Benchmark methods}\label{benchmark-methods}}

The estimates of \(\mathcal{B}_k\) generated by the PENN are compared to
a standard DNN (using \emph{post hoc} explainability algorithms), a DNN
with an interpretable architecture in the form of the
gradient-regularized SENN of
\protect\hyperlink{ref-melisRobustInterpretabilitySelfExplaining2018}{Melis
\& Jaakkola}
(\protect\hyperlink{ref-melisRobustInterpretabilitySelfExplaining2018}{2018}),
and a GAM. Hyperparameters for the DNN, SENN and the PENN are selected
using a validation sample with \(N = 1000\) by minimizing the validation
mean squared error (MSE).\footnote{Hyperparameters include the number of
  hidden nodes, as well as an \(\ell_2\) penalty to regularize weights
  in the DNN, \(\lambda\) in the SENN (the weight of the gradient
  penalty term in the loss function), and \(\delta\) and \(\lambda\) in
  the PENN model. DNN, SENN and PENN contain the same number of nodes
  and hidden layers in all cases.} All neural networks comprise two
hidden layers, with the DNN and SENN emulating the architecture of the
encoder component of the PENN.

While the additive structure of the GAM and the self-explaining
character of the SENN make the methods interpretable, the DNN is, by
default, a black box. Drawing on recent developments in the field of
explainable machine learning, two popular \emph{post hoc} explainability
algorithms based on Shapley values (SHAP) and local linear surrogate
models (LIME) are used to approximate covariate contributions for the
DNN. This class of \emph{post hoc} algorithms estimates a contribution
(\(\phi_{ik}\)) of \(x_{ik}\) to the \(i\)th predicted value
\(\hat{y}_i\), with the aim of decomposing the prediction into \(K\)
variable-specific components, such that

\begin{equation}
\hat{y}_i = \mathbb{E}[\hat{y}] + \sum_k \hat{\phi}_{ik},
\end{equation}

where \(\hat{\phi}_{ik}\) is the contribution estimate. The contribution
is defined using a linear model, with

\begin{equation}
\hat{\phi}_{ik} = \beta_{ik} x_{ik} - \mathbb{E}\left[\beta_{ik} x_{ik}\right],
\label{eq:phi_k}
\end{equation}

where \(\beta_{ik}\) is a coefficient or weight. Note that \(\phi_{ik}\)
is not equivalent to a marginal effect, but can be computed using
\(\beta_{ik}\). In the case of a local linear model, with
\(\hat{y}_i = \beta_0 + \sum_k \beta_{ik}x_{ik}\), summing the
contributions yields the predicted value for the \(i\)th observation,
normalized by the average prediction:

\begin{align}
\sum_{k} \hat{\phi}_{ik} &= \sum_{k} \left(\beta_{ik} x_{ik} -  \mathbb{E}\left[\beta_{ik} x_{ik}\right]\right) + (\beta_0 - \beta_0), \label{eq:phi} \\ 
&= (\beta_0 + \sum_{k} \beta_{ik} x_{ik}) - (\beta_0 + \sum_{k}\mathbb{E}\left[\beta_{ik} x_{ik}\right]),  \\
&= \hat{y}_i - \mathbb{E}\left[\hat{y}\right]. 
\end{align}

The two explainability methods used in this application are among the
most widely applied algorithms and warrant more detailed discussion. The
first is based on the game theoretic concept of Shapley values
(\protect\hyperlink{ref-shapleyValueNPersonGames1953}{Shapley, 1953}),
and conceptually determines a contribution for each covariate \(x_k\),
by approximating

\begin{equation}
\phi_{k} = \mathbb{E}[y|\boldsymbol{x}_{-k}, x_k] - \mathbb{E}[y|\boldsymbol{x}_{-k}],
\end{equation}

where \(\boldsymbol{x}_{-k}\) comprises a set of covariates excluding
\(x_k\)
(\protect\hyperlink{ref-lundbergUnifiedApproachInterpreting2017}{Lundberg
\& Lee, 2017};
\protect\hyperlink{ref-strumbeljExplainingPredictionModels2014}{Štrumbelj
\& Kononenko, 2014}). The expectation is calculated over all possible
subsets of covariates \(\boldsymbol{x}_{-k}\), making the method
computationally expensive. Several method-specific incarnations
introduced in
\protect\hyperlink{ref-lundbergUnifiedApproachInterpreting2017}{Lundberg
\& Lee}
(\protect\hyperlink{ref-lundbergUnifiedApproachInterpreting2017}{2017})
have recently gained popularity, which increase computational efficiency
by exploiting specific features of the underlying machine learning
algorithm. However, these approaches have only limited applicability to
DNN architectures. For this simulation, SHAP values are obtained using
the \texttt{R}-package \texttt{iml}
(\protect\hyperlink{ref-molnarImlPackageInterpretable2018}{Molnar,
2018}), which implements the method according to
\protect\hyperlink{ref-strumbeljExplainingPredictionModels2014}{Štrumbelj
\& Kononenko}
(\protect\hyperlink{ref-strumbeljExplainingPredictionModels2014}{2014}).

The second explainability method used as a benchmark for the PENN is the
local interpretable model-agnostic explanations (LIME) algorithm,
introduced in \protect\hyperlink{ref-ribeiroWhyShouldTrust2016}{Ribeiro
\emph{et al.}}
(\protect\hyperlink{ref-ribeiroWhyShouldTrust2016}{2016}). The algorithm
fits an interpretable model (e.g.~a linear regression) for each
observation, by perturbing the data set at \(\boldsymbol{x}_i\) to
generate a simulated data set \(\boldsymbol{z}\), and subsequently
minimizing

\begin{equation}
\mathcal{L} (f, g, \pi_x) = \sum_j \pi_x(\boldsymbol{z}_j)\left[f(\boldsymbol{z}_j) - g(\boldsymbol{z}_j)\right]^2,
\end{equation}

where \(f\) is the black-box model, \(g\) is a linear regression, and
\(\pi_x\) is a weighting function measuring the proximity between
\(\boldsymbol{x}_i\) and \(\boldsymbol{z}\) (based on the Euclidean
distance in this application). Sampling \(\boldsymbol{z}\) uniformly as
suggested in \protect\hyperlink{ref-ribeiroWhyShouldTrust2016}{Ribeiro
\emph{et al.}} (\protect\hyperlink{ref-ribeiroWhyShouldTrust2016}{2016})
and implemented in various software packages is found to work poorly in
this context. Instead \(\boldsymbol{z}\) is drawn randomly from a narrow
region around \(\boldsymbol{x}_i\), with
\(\boldsymbol{z} \sim \mathcal{N}(\boldsymbol{x}_i, \sigma^2_{\boldsymbol{z}} \boldsymbol{I}_K)\),
where \(\sigma^2_{\boldsymbol{z}}\) is treated as a hyperparameter.

Finally, a GAM is estimated, which generates contribution functions
using smoothing splines, and is computed using the \texttt{R}-package
\texttt{mgcv}
(\protect\hyperlink{ref-woodFastStableRestricted2011}{Wood, 2011}). Note
that GAM and SHAP offer limited comparability to the PENN, since the
methods only produce contributions \(\hat{\phi}_{ik}\) and no local
coefficients. Conversely, \(\hat{\phi}_{ik}\) can easily be calculated
from the estimates generated by PENN, SENN and LIME using Eq.
\ref{eq:phi_k}.

The accuracy of the various methods is computed formally using the mean
absolute error (MAE) of the contributions:

\begin{equation}
\text{MAE}_{\phi}^q = \frac{1}{NK} \sum_{k = 1}^K \sum_{i = 1}^N |\phi_{ik}^{\text{DGP}} - \hat{\phi}_{ik}^{q}|.
\end{equation}

\hypertarget{results}{%
\subsection{Results}\label{results}}

Fig. \ref{fig:sim_main} displays the estimated coefficient functions
(upper panel) and contributions (lower panel) for the various methods,
where \(N = 1000\), \(\sigma^2_{\epsilon} = 1\) and \(\rho = 0\):

\begin{figure}
\centering
\includegraphics{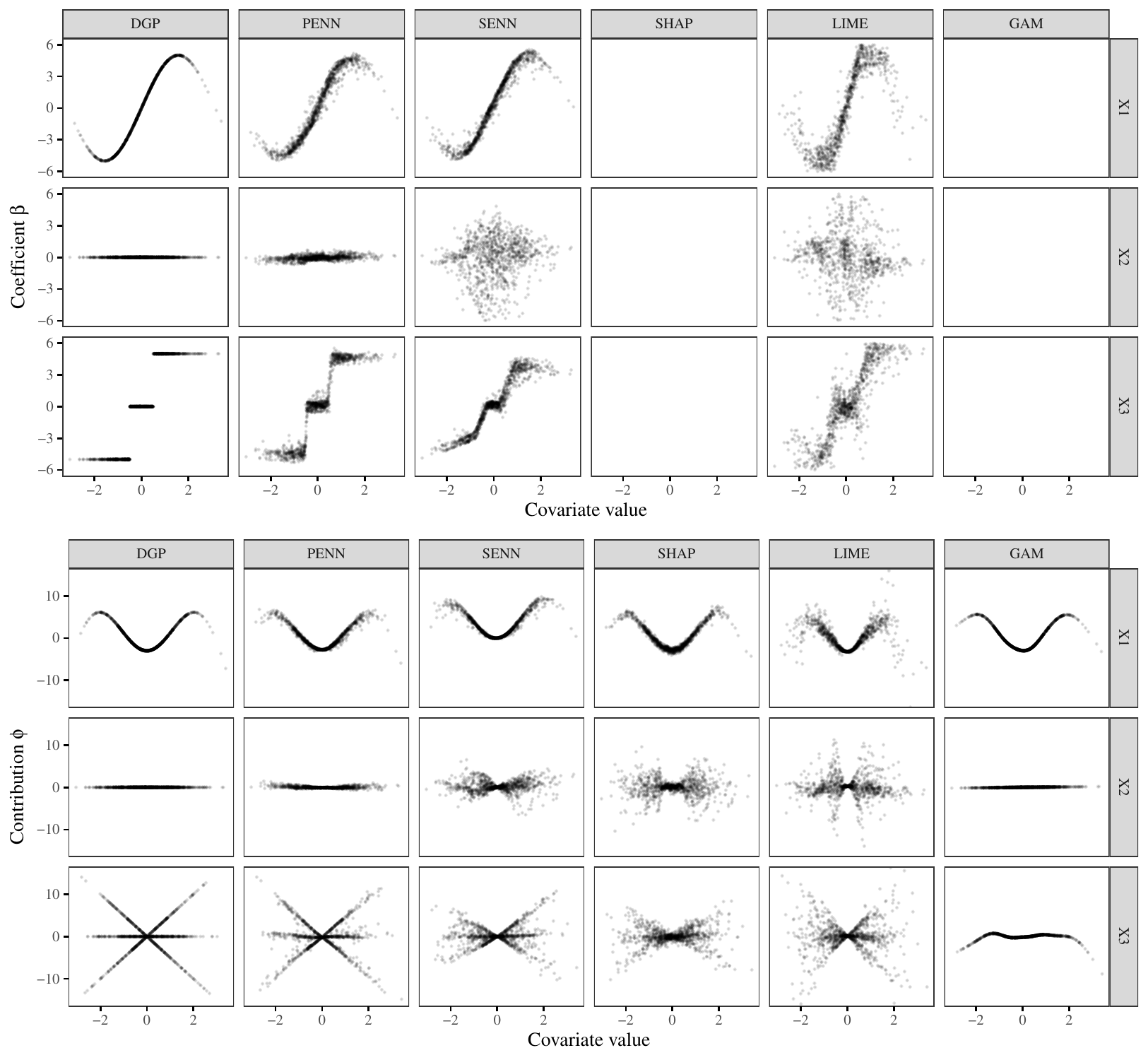}
\caption{Coefficient estimates (upper panel) and contribution estimates
(lower panel) for the DGP, PENN, SENN, SHAP, LIME and GAM methods.
Estimates for the PENN are posterior means. The \(x\)-axis plots the
relevant covariate values, thus the coefficient \(\beta_3\) is plotted
against the values for \(x_2\). \label{fig:sim_main}}
\end{figure}

The PENN model generates coefficients and contributions that closely
resemble the DGP, outperforming all benchmark methods. Particularly the
values associated with \(x_2\) and \(x_3\) suggest that the PENN is
capable of identifying the effects correctly in the presence of feature
dependencies. All benchmark methods produce reasonable results for
\(\phi_1\), but fail to identify the correct contributions for the
remaining covariates, due to an implicit assumption of feature
independence. Since SHAP values are calculated by measuring the average
impact that the inclusion of a covariate has on predictions, it tends to
allocate contribution to both \(x_2\) and \(x_3\) equally. The
additivity underlying the model structures of LIME and GAM is similarly
incapable of capturing feature dependencies. Finally, by forcing the
parameter vectors to act as gradients, the SENN embeds feature
independence into its concept of explainable parameters and
contributions, inducing smoothness of \(\mathcal{B}_k\) over the range
of \(x_k\) and hence penalizing zero values of \(\mathcal{B}_2\).

\begin{figure}
\centering
\includegraphics{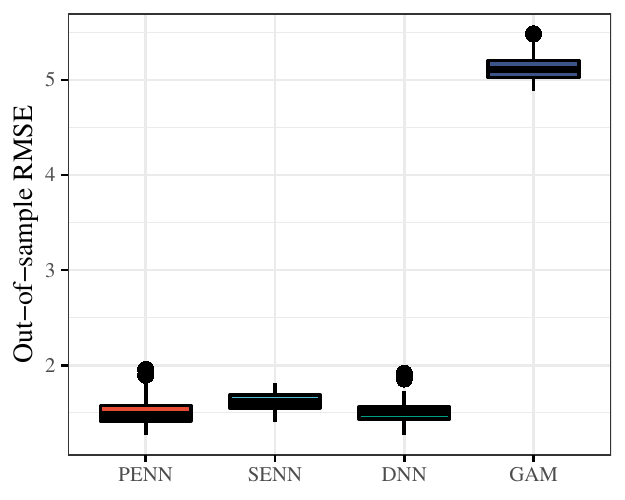}
\caption{Out-of-sample RMSE of the PENN, SENN, DNN and GAM, using 50
simulation runs at the default values of \(N = 1000\),
\(\sigma^2_{\epsilon} = 1\) and \(\rho = 0\). Hyperparameters:
\(\lambda = 0.1\), \(\delta = 0.2\), \(\ell_2\)-norm weight penalty in
the DNN of \(0.001\).\label{fig:sim_mse}}
\end{figure}

Fig. \ref{fig:sim_mse} plots the out-of-sample prediction error of the
PENN, SENN, DNN and GAM for 50 simulation runs with the settings used in
Fig. \ref{fig:sim_main} above. Given its additive structure, the GAM
performs comparatively poorly. The three neural networks capture the DGP
almost identically well, illustrating that the structure of the PENN is
capable of retaining the complexity of an equally-sized DNN. In the same
vein, the poor explanatory performance of the SHAP and LIME algorithms,
as well as the SENN parameters observed above, does not stem from an
insufficiently flexible underlying neural network, but is the result of
the theoretical particularities of the explainability methods
themselves.

In order to examine the relative performance across different simulated
scenarios, Fig. \ref{fig:sim_scen} plots the accuracy of the methods in
terms of \(\text{MAE}_{\phi}\) for different values of \(N\), \(\rho\)
and \(\sigma\) (holding the respective remaining values equal at the
default specified above):

\begin{figure}
\centering
\includegraphics{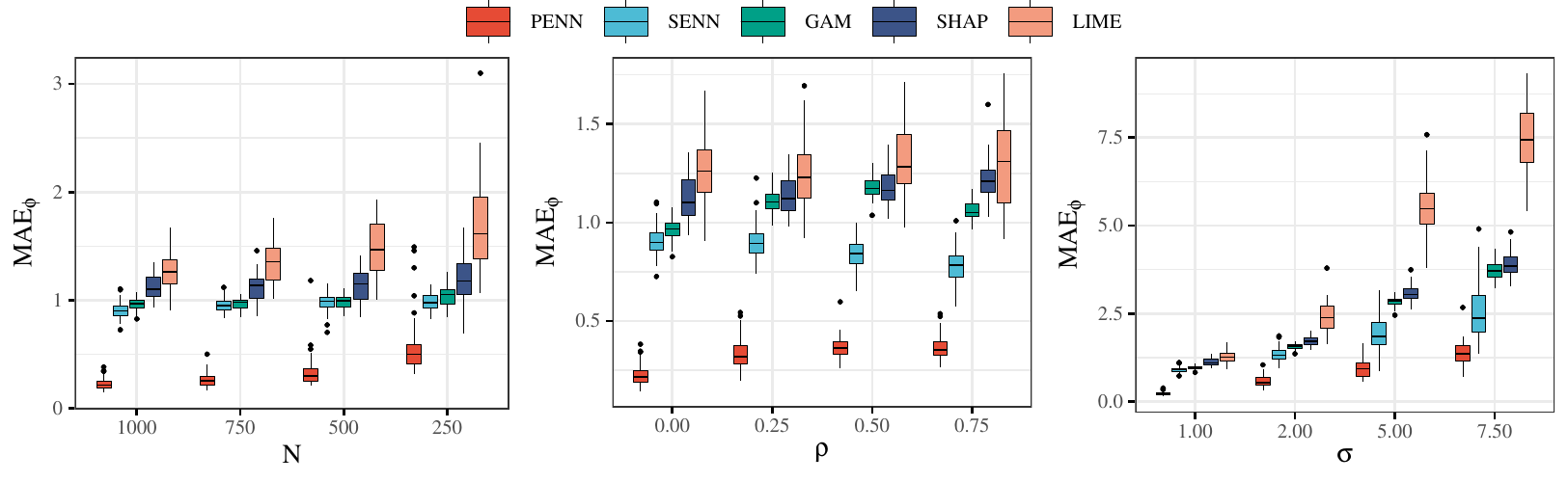}
\caption{Accuracy (\(\text{MAE}_{\phi}\)) for different values of \(N\)
(left), \(\rho\) (middle) and \(\sigma\) (right). Boxplots show the
results of 50 simulation runs for each scenario. \label{fig:sim_scen}}
\end{figure}

Overall, the PENN achieves the highest accuracy in all scenarios. The
left panel plots the effect of decreasing the sample size on
\(\text{MAE}_{\phi}\), the middle panel the role of multicollinearity,
and the right panel of the signal-to-noise ratio. The PENN is
comparatively robust in small sample sizes, which results from the
regularizing effect of the parameter prior. Multicollinearity is not a
substantial problem for any of the methods, with relatively minor
changes to the accuracy. In fact, the SENN even improves slightly as
\(\rho\) increases. Decreasing the signal-to-noise ratio results in a
lower accuracy across all methods. Of the benchmark methods, the SENN
performs best, which may reflect an advantage in embedding
explainability directly into the neural network architecture as opposed
to adding a \emph{post hoc} explainability layer (with an associated
additional source for errors).

\begin{figure}
\centering
\includegraphics{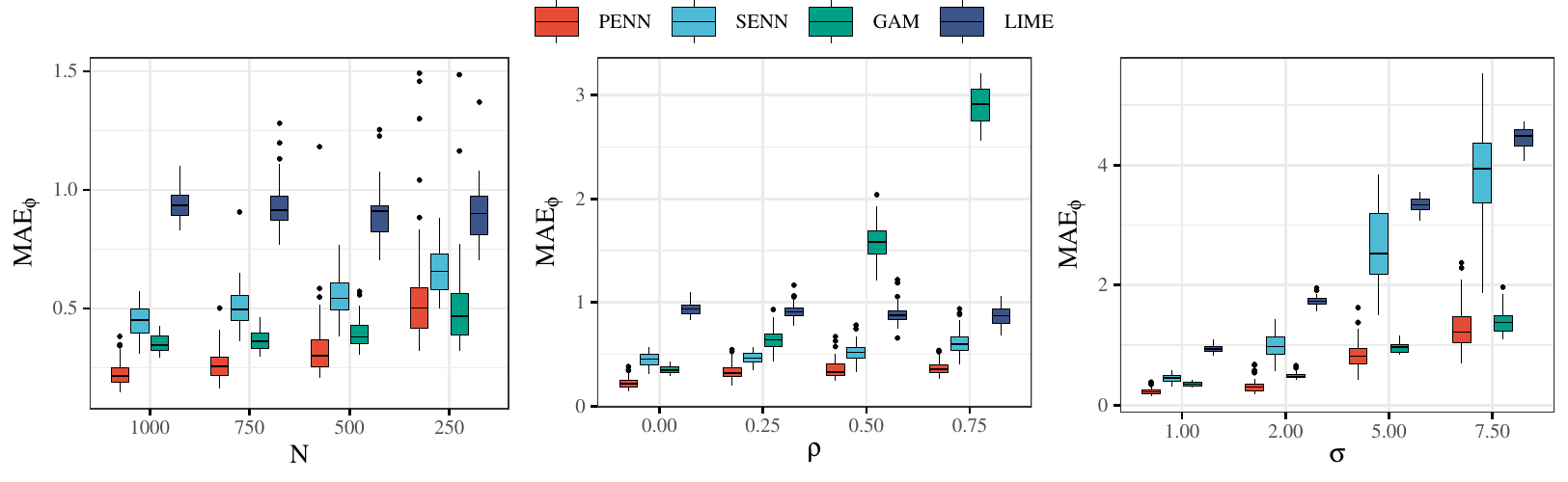}
\caption{Accuracy (\(\text{MAE}_{\phi}\)) for different values of \(N\)
(left), \(\rho\) (middle) and \(\sigma\) (right). The interaction effect
between \(x_2\) and \(x_3\) is modelled explicitly in the GAM, LIME and
SENN methods. Boxplots show the results of 50 simulation runs for each
scenario. \label{fig:sim_scen_interaction}}
\end{figure}

The key comparative advantage of the PENN is its ability to infer the
interaction between \(x_2\) and \(x_3\) without prior knowledge. The
interaction could be modeled explicitly by the benchmark methods, under
the assumption that the existence of the interaction is known \emph{a
priori}. In the case of the GAM, LIME and SENN this is simply a matter
of adding the appropriate interaction term to the regression equations,
and adding the interaction to the input feature set of the SENN. For
SHAP it is more complicated and currently not implemented in the
software packages used here.\footnote{\protect\hyperlink{ref-aasExplainingIndividualPredictions2020}{Aas
  \emph{et al.}}
  (\protect\hyperlink{ref-aasExplainingIndividualPredictions2020}{2020})
  and
  \protect\hyperlink{ref-dattaAlgorithmicTransparencyQuantitative2016}{Datta
  \emph{et al.}}
  (\protect\hyperlink{ref-dattaAlgorithmicTransparencyQuantitative2016}{2016})
  provide discussions of SHAP with dependent features and interactions.
  However, proposed alterations can increase the computational overhead
  of the method dramatically.} The interaction is therefore only
examined for the PENN, SENN, GAM and LIME. Fig.
\ref{fig:sim_scen_interaction} plots the resulting
\(\text{MAE}_{\phi}\), illustrating that the PENN achieves superior or
closely matched accuracy even when the interaction is modeled with prior
knowledge by the benchmark methods. The GAM performs particularly poorly
when the correlation between the interacting variables is high. In this
case, differentiating between \(x_2\) (which has no effect) and the
interaction between \(x_2\) and \(x_3\) is far more difficult in the
GAM, but does not pose as large a challenge for the PENN. The LIME,
while substantially more accurate than in Fig. \ref{fig:sim_scen},
continues to capture the effects poorly.

The PENN model has several advantages over alternative explainability
algorithms. The primary aspect explored in this section is its ability
to estimate parameters consistently in the presence of dependent effects
structures, and without reducing model interpretability. This is an
important achievement, since interpretability is usually obtained at the
cost of complexity --- more specifically at the cost of imposing
additivity and removing any feature interactions. Aside from the ability
to generate explainable results without reducing complexity, the
approach is less computationally costly when compared to DNN-based
alternatives --- particularly SHAP --- which requires extensive
simulations and quickly becomes unfeasible as the number of covariates
grows. Furthermore, the posterior densities of the estimates produced by
the PENN incorporate a measure of confidence in the parameter values,
and permit parameter inference in highly nonlinear settings. This latter
point is explored in detail using the empirical example in the following
section.

\hypertarget{real-time-risk-measurement-in-equity-markets}{%
\section{\texorpdfstring{Real-time risk measurement in equity markets
\label{application}}{Real-time risk measurement in equity markets }}\label{real-time-risk-measurement-in-equity-markets}}

The application introduced in this section illustrates how a PENN
architecture can be used (i) to explore nonlinear behavior in asset
markets, and (ii) to conduct parameter inference. I estimate a nonlinear
version of the capital asset pricing model (CAPM) for 10 global sector
classifications, with the nonlinear dynamics of systematic risk driven
by the economic regime. The application is sufficiently simple to
illustrate key aspects of the PENN methodology, while simultaneously
providing a novel perspective on risk in equity markets. The underlying
neural network facilitates the prediction of systematic and
idiosyncratic risk components in real-time, addressing two of the most
important empirical critiques to the CAPM: its static and its
backward-looking character.

The application is embedded within the theoretical literature on the
conditional CAPM, which posits that risk premia vary based on the state
that the economy resides in at a given point in time (see
\protect\hyperlink{ref-lewellenConditionalCAPMDoes2006}{Lewellen \&
Nagel} (\protect\hyperlink{ref-lewellenConditionalCAPMDoes2006}{2006})
for a review). The proposed PENN model can be viewed as a nonlinear
rendition of the conditional CAPM introduced in
\protect\hyperlink{ref-jagannathanConditionalCAPMCrossSection1996}{Jagannathan
\& Wang}
(\protect\hyperlink{ref-jagannathanConditionalCAPMCrossSection1996}{1996})
and \protect\hyperlink{ref-petkovaValueRiskierGrowth2005}{Petkova \&
Zhang} (\protect\hyperlink{ref-petkovaValueRiskierGrowth2005}{2005}),
where the economic state is described using a data set of external
macroeconomic instruments. The results reveal substantial variation in
risk premia over time, with a closer examination of risk dynamics
suggesting that the nonlinear character of the proposed conditional CAPM
is indeed appropriate.

\hypertarget{discussion-of-the-capm}{%
\subsection{Discussion of the CAPM}\label{discussion-of-the-capm}}

The capital asset pricing model, developed by
\protect\hyperlink{ref-sharpeCapitalAssetPrices1964}{Sharpe}
(\protect\hyperlink{ref-sharpeCapitalAssetPrices1964}{1964}) and
\protect\hyperlink{ref-lintnerValuationRiskAssets1965}{Lintner}
(\protect\hyperlink{ref-lintnerValuationRiskAssets1965}{1965}),
represents an important cornerstone of academic financial theory. It is
constructed based on the assumptions of efficient markets with rational
risk-averse investors, and characterizes a trade-off between risk and
return of financial assets. Despite several prominent empirical and
theoretical critiques, the CAPM continues to play an important role for
financial and investment practitioners as an objective and intuitive
approach to cost of capital estimation, business valuation and
performance measurement.

The investment decision was originally framed by
\protect\hyperlink{ref-markowitzPortfolioSelection1952a}{Markowitz}
(\protect\hyperlink{ref-markowitzPortfolioSelection1952a}{1952}) as a
trade-off between expected risk and return. For any level of expected
return, a rational investor attempts to minimize risk, and for any level
of risk, the investor aims to maximize return. The feasible set of
portfolios emerging from these decision criteria traces a pareto
efficient frontier, which optimally trades off risk and return. By
introducing a risk-free asset to the investable universe, the efficient
frontier collapses to a linear frontier --- the security market line
(SML) --- between the risk-free asset and a tangency portfolio situated
on the Markowitz risk-return frontier
(\protect\hyperlink{ref-sharpeCapitalAssetPrices1964}{Sharpe, 1964}).
Given appropriate assumptions, the linear SML is described
mathematically by the CAPM, which formulates the expected return on
asset \(k\) as a function of the risk-free return (\(r_f\)) and the
return on the tangency portfolio (given by the value-weighted market
portfolio, \(r_m\)):\footnote{If investors can borrow and lend freely at
  the risk-free rate, expected utility is maximized by holding only the
  tangency portfolio and the risk-free asset, with the weights
  determined by the investor's degree of risk aversion. Under the
  assumptions of information efficiency (all market participants form
  equivalent risk and return expectations) and market clearing (all
  assets have an owner), the tangency portfolio must be a value-weighted
  portfolio of the entire investable universe.}

\begin{equation}
\mathbb{E}[r_k] = r_f + (r_m - r_f)\beta_k.
\label{eq:capm}
\end{equation}

Thus, asset \(k\) earns the risk-free rate of return plus a risk
premium. The risk premium depends on the expected excess return earned
by the market portfolio, as well as the asset's market beta
(\(\beta_k\)). The market beta measures the overall systematic risk
exposure of asset \(k\) and is typically estimated using a time series
regression of excess asset returns on a proxy for the market risk
premium (\protect\hyperlink{ref-rossiCapitalAssetPricing2016}{Rossi,
2016}):

\begin{equation}
\tilde{r}_{tk} = \alpha_k + \beta_{k} \tilde{r}_{tm} + \epsilon_t,
\label{eq:capm_reg}
\end{equation}

where \(\tilde{r}_{tk} = r_{tk} - r_{tf}\) and
\(\tilde{r}_{tm} = r_{tm} - r_{tf}\) are the excess returns on the asset
and market, respectively, and \(t\) indexes time. Idiosyncratic
variation (\(\epsilon_t\)) is argued to be diversifiable and hence does
not earn a return premium, with \(\mathbb{E}[\epsilon_t] = 0\). Finally,
\(\alpha_k\) (or simply ``alpha'') captures market risk-adjusted excess
return. In the CAPM, the market return represents the only systematic
risk factor, and it must hold that \(\alpha_k = 0\). The case when
\(\alpha_k \neq 0\) can be viewed as evidence against the validity of
the CAPM, suggesting either the existence of additional uncaptured risk
sources, or the failure of another related assumption (e.g.~the
assumption of a static \(\beta_k\)).

There have been several important theoretical and empirical challenges
to the CAPM (see
\protect\hyperlink{ref-brownAnalysisInvestmentsManagement2012}{Brown \&
Reilly}
(\protect\hyperlink{ref-brownAnalysisInvestmentsManagement2012}{2012})
for an overview). Observed failures of the assumption that
\(\alpha_k = 0\) have been explained most prominently in two separate
strands of the literature: (i)
\protect\hyperlink{ref-famaCrossSectionExpectedStock1992}{Fama \&
French}
(\protect\hyperlink{ref-famaCrossSectionExpectedStock1992}{1992}),
\protect\hyperlink{ref-famaCommonRiskFactors1993}{Fama \& French}
(\protect\hyperlink{ref-famaCommonRiskFactors1993}{1993}), and
\protect\hyperlink{ref-famaSizeBooktoMarketFactors1995}{Fama \& French}
(\protect\hyperlink{ref-famaSizeBooktoMarketFactors1995}{1995})
criticize the supposition of a single systematic risk source, and posit
the existence of several additional risk premia using a multi-factor
version of the CAPM; (ii) The observed instability of market beta over
time has given rise to a conditional version of the CAPM, that permits
variation in \(\beta_k\) conditional on the state of the economy.

As highlighted in
\protect\hyperlink{ref-brownAnalysisInvestmentsManagement2012}{Brown \&
Reilly}
(\protect\hyperlink{ref-brownAnalysisInvestmentsManagement2012}{2012}),
empirical studies have generally found the market beta to be unstable
over time with static estimates highly sensitive to the chosen sample
period. Estimates of \(\beta_{k}\) obtained from historical data
therefore tend to be poor predictors of future risk
(\protect\hyperlink{ref-rossiCapitalAssetPricing2016}{Rossi, 2016}). A
theoretical literature explores a conditional version of the CAPM,
arguing that \(\beta_k\) should be expected to vary over the course of
the business cycle, as market participants demand a higher hurdle rate
of return to compensate equity risk during recessionary periods (see for
instance
\protect\hyperlink{ref-jensenPerformanceMutualFunds1968}{Jensen}
(\protect\hyperlink{ref-jensenPerformanceMutualFunds1968}{1968}),
\protect\hyperlink{ref-dybvigDifferentialInformationPerformance1985}{Dybvig
\& Ross}
(\protect\hyperlink{ref-dybvigDifferentialInformationPerformance1985}{1985}),
and \protect\hyperlink{ref-hansenRoleConditioningInformation1987}{Hansen
\& Richard}
(\protect\hyperlink{ref-hansenRoleConditioningInformation1987}{1987})
for a discussion of the theoretical models). The conditional CAPM
emerged both in response to the empirical observation of unstable market
betas, and (with mixed success) in response to the failure of the zero
alpha assumption, suggesting that unexplained excess return
(\(\alpha_k \neq 0\)) stems not from unaccounted risk premia, but from
the existence of multiple beta-regimes.

Examples of empirical estimates of conditional CAPM include
\protect\hyperlink{ref-jagannathanConditionalCAPMCrossSection1996}{Jagannathan
\& Wang}
(\protect\hyperlink{ref-jagannathanConditionalCAPMCrossSection1996}{1996}),
\protect\hyperlink{ref-adrianLearningBetaTimeVarying2004}{Adrian \&
Franzoni}
(\protect\hyperlink{ref-adrianLearningBetaTimeVarying2004}{2004}),
\protect\hyperlink{ref-petkovaValueRiskierGrowth2005}{Petkova \& Zhang}
(\protect\hyperlink{ref-petkovaValueRiskierGrowth2005}{2005}),
\protect\hyperlink{ref-lustigHousingCollateralConsumption2005}{Lustig \&
Van Nieuwerburgh}
(\protect\hyperlink{ref-lustigHousingCollateralConsumption2005}{2005})
and \protect\hyperlink{ref-santosLaborIncomePredictable2006}{Santos \&
Veronesi}
(\protect\hyperlink{ref-santosLaborIncomePredictable2006}{2006}). A
particularly interesting pendant to the approach presented here is found
in \protect\hyperlink{ref-petkovaValueRiskierGrowth2005}{Petkova \&
Zhang} (\protect\hyperlink{ref-petkovaValueRiskierGrowth2005}{2005}),
who determine market beta based on a data set of macroeconomic variables
(denoted \(\boldsymbol{z}_t\)):

\begin{align}
&\mathbb{E}[r_{tk}] = r_{tf} + \hat{\beta}_{tk}\tilde{r}_{tm}, \;\;\; \text{where} \label{eq:lin_cCAPM_1} \\
&\hat{\beta}_{tk} = \gamma_0 + \boldsymbol{z}_{t-1} \boldsymbol{\gamma} \label{eq:lin_cCAPM_2}.
\end{align}

Here \(\gamma_0\) and \(\boldsymbol{\gamma}\) are coefficients. In the
case of \protect\hyperlink{ref-petkovaValueRiskierGrowth2005}{Petkova \&
Zhang} (\protect\hyperlink{ref-petkovaValueRiskierGrowth2005}{2005}),
\(r_{tk}\) is the return on high and low book-to-market value
portfolios, and the authors attempt to explain the apparent existence of
a value premium using the economic state. Substituting Eq.
\ref{eq:lin_cCAPM_2} into Eq. \ref{eq:lin_cCAPM_1} allows for estimation
of \(\hat{\beta}_{tk}\) in a regression framework with interaction terms
between \(\tilde{r}_{tm}\) and \(\boldsymbol{z}_t\).\footnote{Note that
  \protect\hyperlink{ref-petkovaValueRiskierGrowth2005}{Petkova \&
  Zhang} (\protect\hyperlink{ref-petkovaValueRiskierGrowth2005}{2005})
  use a conditional estimate of the excess market return,
  \(\mathbb{E}\left[\tilde{r}_{tm}|\boldsymbol{z}_{t-1}\right]\),
  determined by regressing the observed \(\tilde{r}_{tm}\) on
  \(\boldsymbol{z}_{t-1}\). Since the scope of this application is
  limited to the study of dynamic time-varying betas, the observed
  excess market return is employed directly instead.}

A PENN model can be used to estimate a nonlinear version of the above
framework, where parameter estimates are obtained using:

\begin{equation}
\begin{bmatrix} \hat{\alpha}_{tk} & \hat{\beta}_{tk} \end{bmatrix} \sim q_{\boldsymbol{\theta};k}([ \alpha_{tk} \;\;\; \beta_{tk} ] |\boldsymbol{z}_{t-1}), \label{eq:nn_cCAPM_2}
\end{equation}

and
\(q_{\boldsymbol{\theta};k}([ \alpha_{tk} \;\;\; \beta_{tk} ] |\boldsymbol{z}_{t-1})\)
is the inference network. This results in a PENN architecture, with
local regression parameters \(\alpha_{tk}\) and \(\beta_{tk}\), whose
parameterizations are inferred based on an encoder input data set
\(\boldsymbol{z}_{t-1}\), a decoder input \(\tilde{r}_{tm}\), and the
output \(\tilde{r}_{tk}\). The setup corresponds to the architecture
described in Fig. \ref{fig:network_full}.

Other approaches to the nonlinear estimation of CAPM typically employ
time-varying parameter frameworks, such as rolling sample regressions
(e.g.
\protect\hyperlink{ref-koutmosEstimatingSystematicRisk2002}{Koutmos \&
Knif}
(\protect\hyperlink{ref-koutmosEstimatingSystematicRisk2002}{2002}),
\protect\hyperlink{ref-adrianLearningBetaTimeVarying2004}{Adrian \&
Franzoni}
(\protect\hyperlink{ref-adrianLearningBetaTimeVarying2004}{2004}), and
\protect\hyperlink{ref-glovaTimeVaryingCAPMIts2015}{Glova}
(\protect\hyperlink{ref-glovaTimeVaryingCAPMIts2015}{2015})). The
approach proposed here differs in the important respect that, instead of
inferring the evolution of \(\beta_{tk}\) over the temporal dimension,
the PENN learns correlative associations between \(\beta_{tk}\) and the
regime that the economy and financial markets reside in at time \(t\).
Since the regimes are inferred from input data, estimates are less
dependent on their immediate history, and the model yields better
real-time forecasts than time-varying parameter alternatives. Section
\ref{app:results} highlights this distinction by comparing PENN
estimates to rolling OLS estimates over different sample periods.

\hypertarget{data}{%
\subsection{Data}\label{data}}

\begin{figure}
\centering
\includegraphics{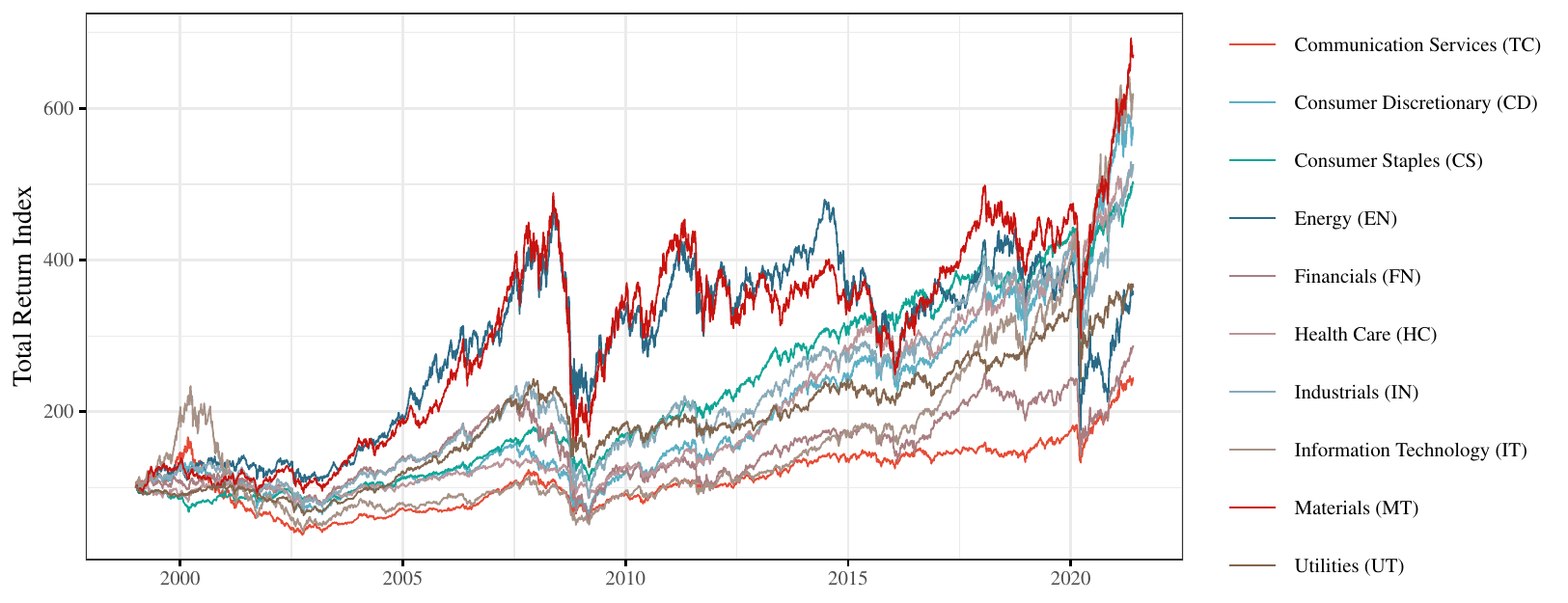}
\caption{Total return index development of sector indices. Data source:
Bloomberg.\label{fig:index_data}}
\end{figure}

Equity risk premia are estimated for each of 10 sectors using a global
universe of stock returns.\footnote{A complete set of global sectors
  according to the global industry classifiction standard (GICS)
  includes 11 sectors. The real estate sector is excluded here due to
  limited data availability.} The market return (\(r_{tm}\)) is proxied
using the MSCI All Countries World Index (ACWI), which is a
value-weighted index of global large and mid-cap stocks comprising over
3000 constituents. The sector returns (\(r_{tk}\)) are given by the
associated sector-specific sub-indices of the MSCI ACWI. Returns are
calculated over a rolling one month period for each index and adjusted
for dividends and distributions. Individual sectors are referred to by
the two letter acronyms listed in Fig. \ref{fig:index_data} (e.g.~EN =
Energy).

The risk-free rate (\(r_{tf}\)) is taken to be the 3-month US Treasury
bill rate, as is typically assumed in the related literature. US
economic variables are used to proxy for both the risk-free rate and the
macroeconomic state. While this implies a regional mismatch to the MSCI
ACWI index, the high percentage of US equities in the index and the
important role of the US financial system suggest that the
generalization is reasonable. The sample consists of daily data for the
period January 2000 -- May 2021, and thus encompasses three recessions:
the dot-com bubble in 2000, the global financial crisis of 2008 and the
COVID-19 pandemic of 2020-21 (ongoing at time of writing).

The inputs to the encoder network of the PENN model govern the regime
dynamics of alpha and beta estimates. A natural choice of input
variables are measures typically considered to be regime determinants in
financial markets. The information set \(\boldsymbol{z}_{t-1}\)
comprises 6 such regime determinants as inputs to the encoder,
augmenting the data set used in
\protect\hyperlink{ref-petkovaValueRiskierGrowth2005}{Petkova \& Zhang}
(\protect\hyperlink{ref-petkovaValueRiskierGrowth2005}{2005}) by two
additional variables. The monetary stance is captured by short-term US
interest rates (3-month Treasury bill rate). A US term premium (10-year
minus 2-year maturity Treasury bond yields) and a default premium
(10-year corporate bond yields minus 10-year Treasury bond yields)
summarize risk dynamics in the economy.\footnote{The Moody's Seasoned
  10-year Baa Corporate Bond Index is used to measure US corporate bond
  yields.} The MSCI ACWI dividend yield is a proxy for the overall
return on equity investments and profitability of firms in the
underlying index. Finally, long-term real and nominal economic
expectations are expressed using the 10-year US real yield (yield on US
inflation indexed bonds), and long-term inflation expectations (10-year
US breakeven inflation). All variables enter the model as year-on-year
differences and are standardized.

\begin{figure}
\centering
\includegraphics{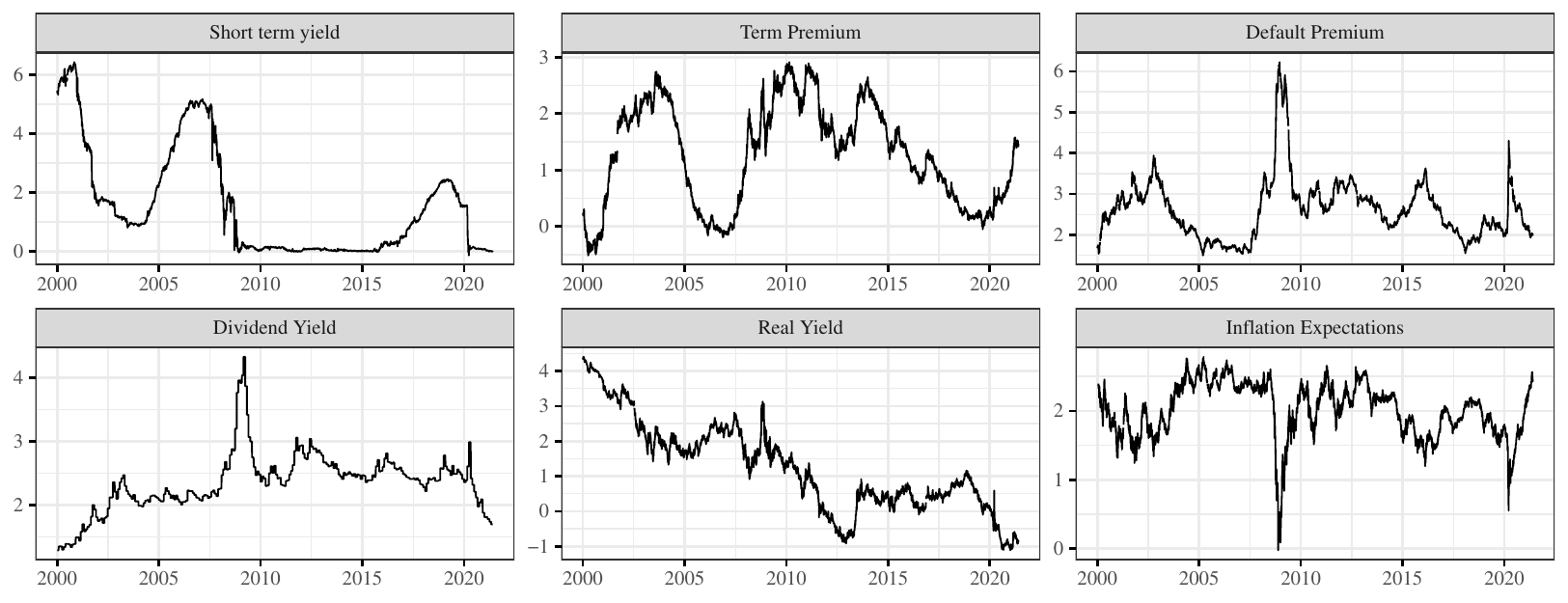}
\caption{Macroeconomic regime determinants (untransformed) used as input
variables to the encoder network (\(\boldsymbol{z}_t\)). Data source:
Bloomberg, author's calculations.}
\end{figure}

\hypertarget{model-selection}{%
\subsection{\texorpdfstring{Model selection
\label{app:cv}}{Model selection }}\label{model-selection}}

The networks are regularized using the two hyperparameters \(\lambda\)
and \(\delta\), with linear constant parameters resulting as
\(\lambda \rightarrow \infty\) and \(\delta \rightarrow 0\). Model
selection for machine learning algorithms (i.e.~selecting optimal
hyperparameter values) is most commonly performed using some form of
cross-validation (CV) algorithm. CV algorithms perform pseudo
out-of-sample model evaluations by training the network several times on
different subsets of the data, and evaluating each fit using validation
samples that are withheld during training. Hyperparameters are chosen
based on a measure of predictive accuracy (validation MSE).

To accommodate the time-dependence structure of time series data, model
selection in this special case usually involves expanding or rolling
window CV procedures. Since these methods discard a substantial portion
of the data during training (e.g.~recent data is only used in a single
training slice), I employ the \(hv\)-block CV algorithm described in
\protect\hyperlink{ref-racineConsistentCrossValidatoryModelSelection2000}{Racine}
(\protect\hyperlink{ref-racineConsistentCrossValidatoryModelSelection2000}{2000}).
Data is divided into \(v\) consecutive validation blocks that retain
their ordering. A margin of \(h\) samples before and after each
validation set is masked from training, to prevent data leakage between
training and validation sets, which may occur due to time dependencies
and autocorrelation.
\protect\hyperlink{ref-bergmeirUseCrossValidationTime2012}{Bergmeir \&
Benítez}
(\protect\hyperlink{ref-bergmeirUseCrossValidationTime2012}{2012}) use
rigorous empirical tests to demonstrate that the benefit of the more
efficient use of data during training outweighs the theoretical
inconsistency of \(hv\)-block CV, that results from an evaluation with
past data. The authors find that \(hv\)-block CV achieves significantly
better results than expanding window alternatives.

In the application, I set \(v = 10\) and \(h = 10\), and evaluate a grid
of candidate values of \(\lambda\) and \(\delta\) that includes both the
static and the unregularized extremes, with the optimal model minimizing
the validation error. Hyperparameter tuning is performed individually
for each of the sector indices.

\hypertarget{discussion-of-results}{%
\subsection{\texorpdfstring{Discussion of results
\label{app:results}}{Discussion of results }}\label{discussion-of-results}}

\begin{figure}
\centering
\includegraphics{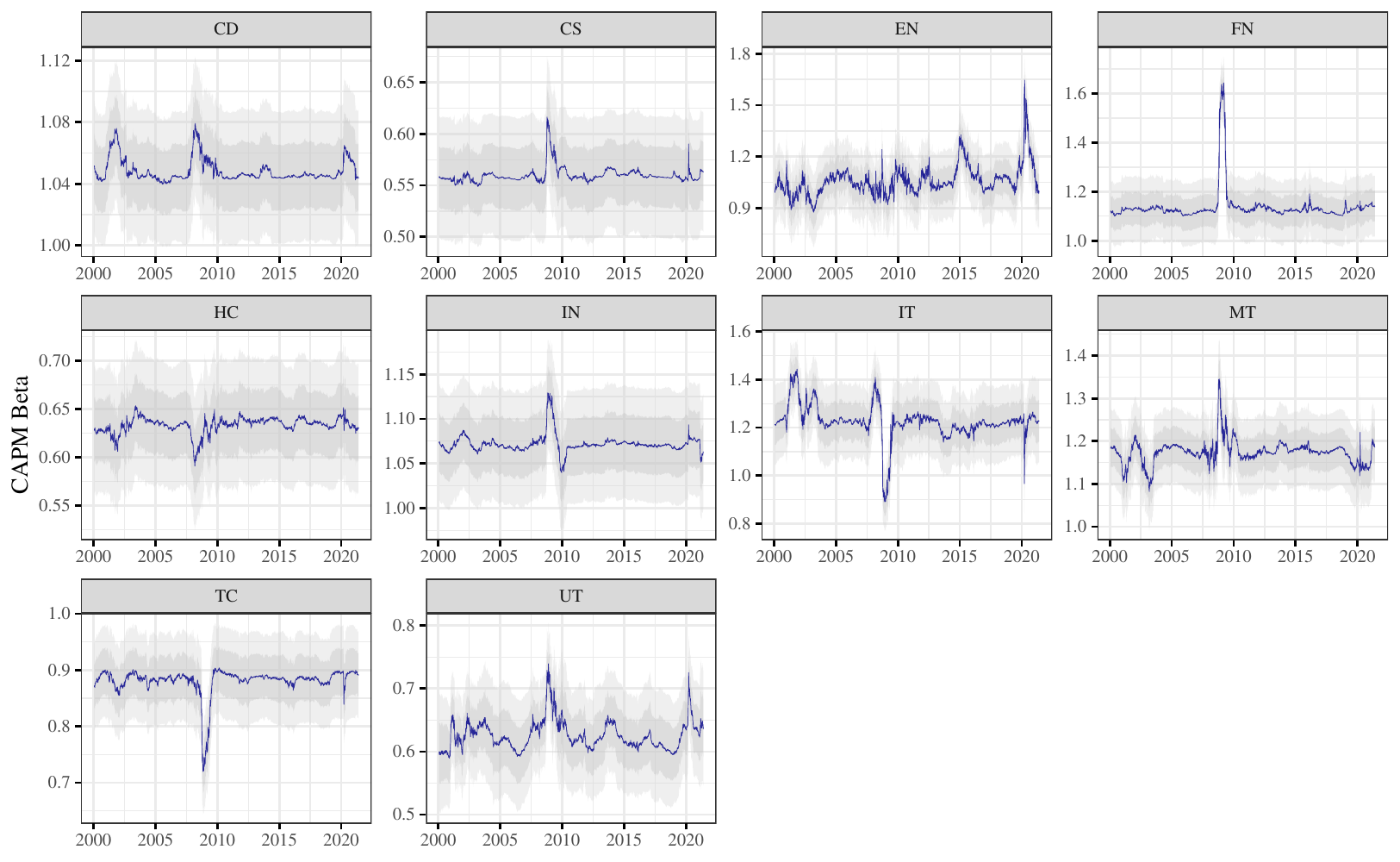}
\caption{PENN estimates of \(\beta_{tk}\) for 10 GICS sectors. Shaded
areas indicate two standard deviations of the posterior
density.\label{fig:opt_beta}}
\end{figure}

Fig. \ref{fig:opt_beta} plots PENN estimates of the market beta for each
sector over the sample period. Most sectors present distinctly nonlinear
patterns, confirming the empirical finding of unstable market betas. In
general, defensive sectors (CS, HC, TC and UT) reflect lower systematic
risk, with \(\hat{\beta}_{tk} < 1.0\). Cyclical sectors exhibit a higher
sensitivity to the market, demanding a higher overall risk premium.

\begin{figure}
\centering
\includegraphics{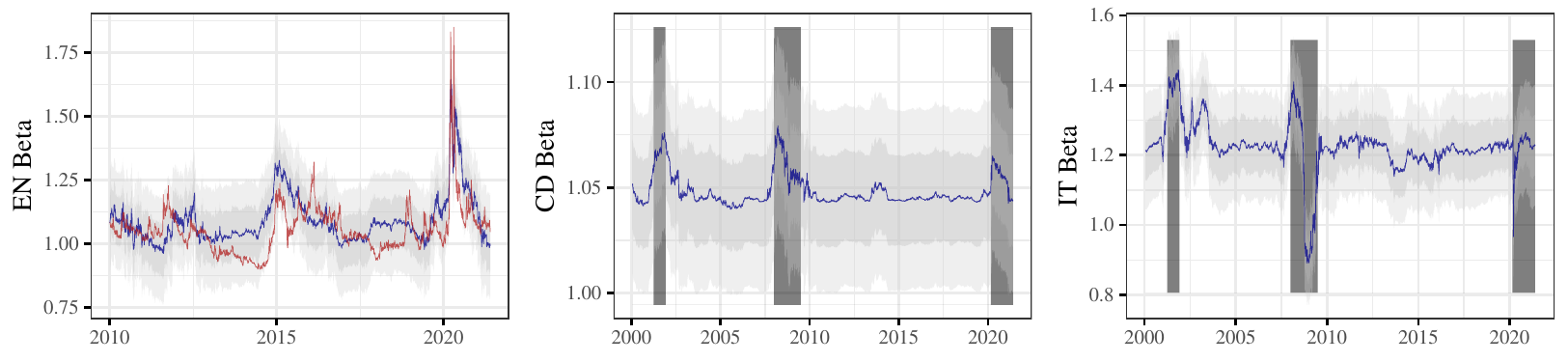}
\caption{PENN estimates of \(\beta_{tk}\) for the EN, CD an IT sectors
(blue), overlaid with 3-month Crude Oil implied volatility (red,
adjusted range). Data is shown only for the last decade in the left
panel, since the measure of implied volatility is not available for the
full sample period. The center and right panels include shaded areas
indicating US recessions. Data sources: Bloomberg, NBER, author's
calculations.\label{fig:opt_en_beta}}
\end{figure}

The spikes in systematic risk exposure are concentrated around periods
of economic crisis, suggesting that market participants require a higher
compensation for holding equity during tumultuous market phases. This
finding aligns with theoretical notions underpinning the development of
the conditional CAPM. Distinct differences between sectors' cyclical
responses can be observed. For instance, the financial sector exhibits a
large risk increase during the financial crisis of 2008, while the IT
sector realizes a peak during the dot-com bust. Other sectors, such as
CD or UT spike during all recessionary periods (see, for instance, the
middle panel in Fig. \ref{fig:opt_en_beta}).

Systematic risk in the energy sector is closely related to the oil
price. Fig. \ref{fig:opt_en_beta} plots EN market beta alongside the
3-month implied Brent Crude oil price volatility, suggesting a strong
correlation. The market beta clearly reflects the oil price collapse in
2014-16, as well as the energy market turmoil during the COVID-19
pandemic of 2020. The plot illustrates the neural network's ability to
learn meaningful behavior without explicitly including oil market
volatility in the input feature set.

\begin{figure}
\centering
\includegraphics{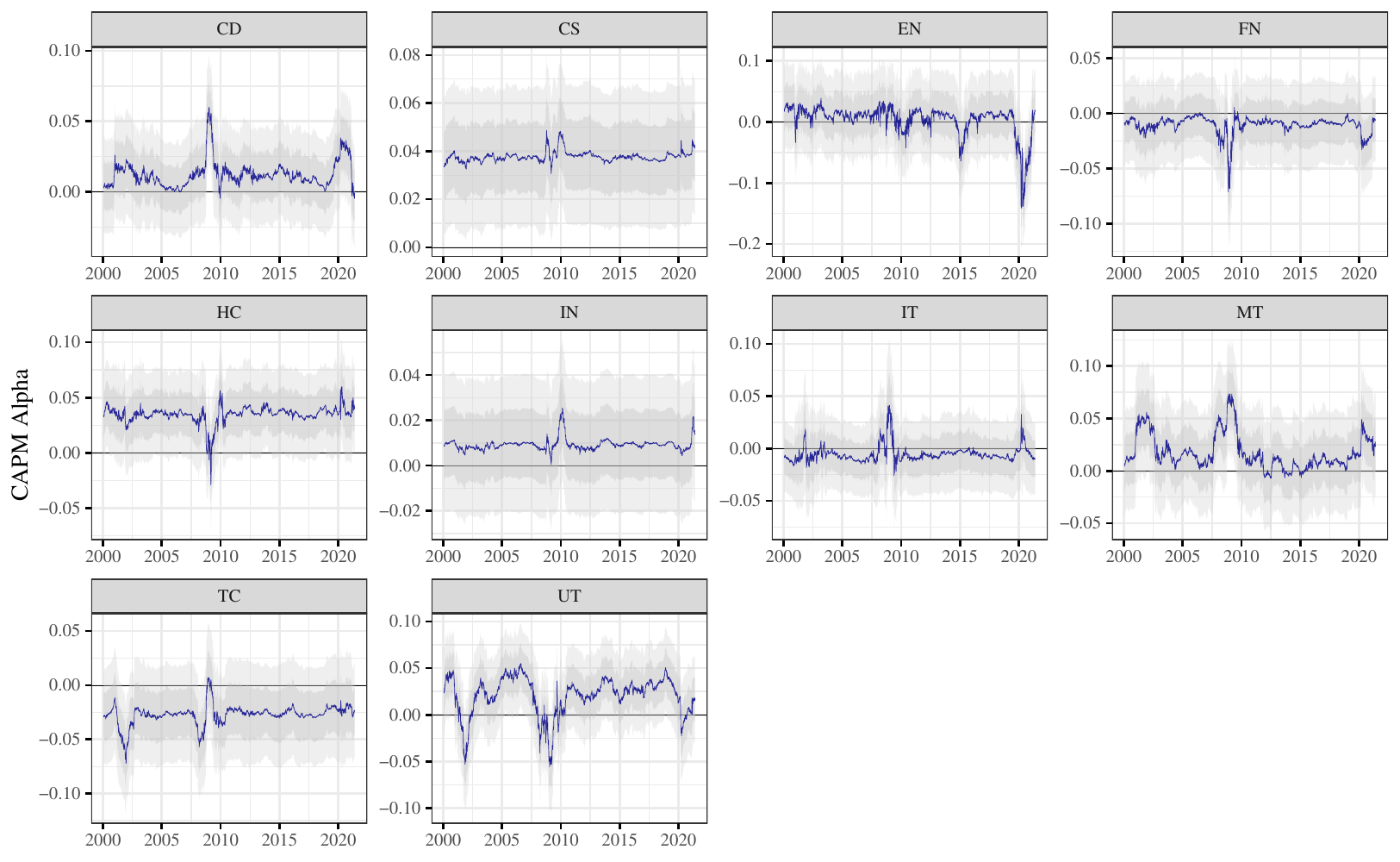}
\caption{PENN estimates of \(\alpha_{tk}\) for 10 GICS sectors. Shaded
areas indicate two standard deviations of the posterior density.
\label{fig:opt_alpha}}
\end{figure}

Fig. \ref{fig:opt_alpha} plots the nonlinear estimates of
\(\alpha_{tk}\) for each sector. Recall that the parameter measures the
market risk-adjusted excess return. In standard CAPM theory, the
parameter is assumed to equal zero. The plots show that it is indeed
statistically equal to zero in many cases based on the 95\% confidence
interval. Several sectors exhibit a higher excess return during crises
(e.g.~IT, IN, MT, CD), and some exhibit a lower excess return (e.g.~FN,
UT, EN). This suggests that, apart from its impact on risk, the business
cycle also affects the extent to which the CAPM can be used to explain
excess return. During downturns other risk factors (not captured by
CAPM) play an important explanatory role. Note that some sectors
(e.g.~CS, HC) exhibit positive values of alpha throughout the sample,
suggesting that the CAPM omits important sources of risk and return in
these sectors.

\begin{figure}
\centering
\includegraphics{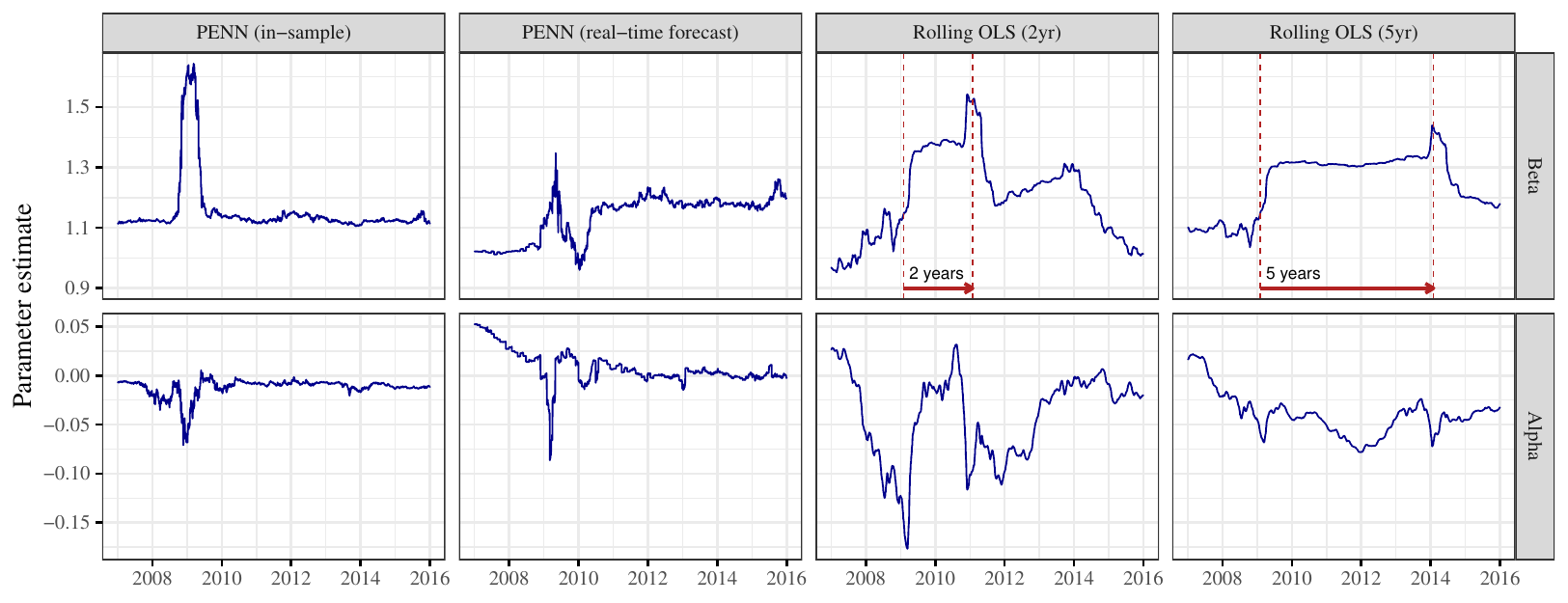}
\caption{Market alpha and beta for MSCI ACWI financials sector,
estimated using a PENN model (both in-sample and out-of-sample), and two
rolling-window OLS regressions (2-years and 5-years). Real-time
estimates of the PENN are obtained by forecasting the parameter value
using \(\boldsymbol{z}_{t-1}\) with an expanding window estimation of
the neural network, and retraining on a monthly basis.
\label{fig:penn_vs_ols}}
\end{figure}

The estimates of beta and alpha presented in the preceding figures can
be forecast in real-time conditional on \(\boldsymbol{z}_{t-1}\) and are
not subject to the backward-looking bias of conventional rolling sample
estimates of the CAPM. The distinction can easily be understood by
visualizing the effect of a large outlier event on the estimates. Fig.
\ref{fig:penn_vs_ols} plots the FN estimates of alpha and beta produced
by the PENN (both in-sample and real-time) alongside rolling OLS
regressions with 2 and 5-year estimation windows, for the period
following the global financial crisis of 2008. The real-time estimates
of the PENN are out-of-sample forecasts of \(\beta_{t+1,\text{FN}}\) and
\(\alpha_{t+1,\text{FN}}\) in an expanding window estimation. The
real-time estimates are broadly similar to the in-sample estimates, with
a more noisy response to the crisis in 2008 on an out-of-sample basis,
as is to be expected given only historical data.

The observations associated with the 2008 crisis have high leverage in
the OLS regressions resulting in sustained high estimated levels of the
market risk premium and reduced estimated levels of excess return that
correspond exactly to the length of the sample window. Once the crisis
rolls out of the backward-looking sample period, estimates adjust
abruptly. The PENN by contrast exhibits a spike associated with the
crisis, but returns to a more moderate level by 2010 both in the
in-sample and the real-time estimates.

This well-known problem of backward-looking estimation frameworks
implies that time-varying market betas obtained in this manner are
poorly suited to gauge true systematic risk exposure. In the years
following the global financial crisis, estimates of the cost of equity
would have remained artificially inflated for a period as long as the
historical sample drawn upon in the estimation. The PENN offers a unique
solution to this problem, by generating predictive real-time forecasts
of the regression parameters.

\hypertarget{how-is-risk-inferred-from-the-economic-state}{%
\subsection{How is risk inferred from the economic
state?}\label{how-is-risk-inferred-from-the-economic-state}}

As stated at the outset, the PENN model represents a nonlinear version
of a conditional CAPM. In order to study the manner in which variation
in \(\hat{\beta}_{tk}\) and \(\hat{\alpha}_{tk}\) is embedded in the
economic state, a further layer of explanation can be added to the PENN
model. The inference network, which parameterizes the posterior of alpha
and beta based on \(\boldsymbol{z}_{t}\), remains a black-box.
Conventional explainability algorithms can shed light on the correlative
structure of regime inference occurring in the network in a type of
meta-explainability model. To do so, I estimate the local contributions
of each variable in \(\boldsymbol{z}_{t-1}\) to the predicted posterior
mean of \(\hat{\beta}_{tk}\) and \(\hat{\alpha}_{tk}\) using the SHAP
algorithm introduced in Section \ref{simulation}.

\begin{figure}
\centering
\includegraphics{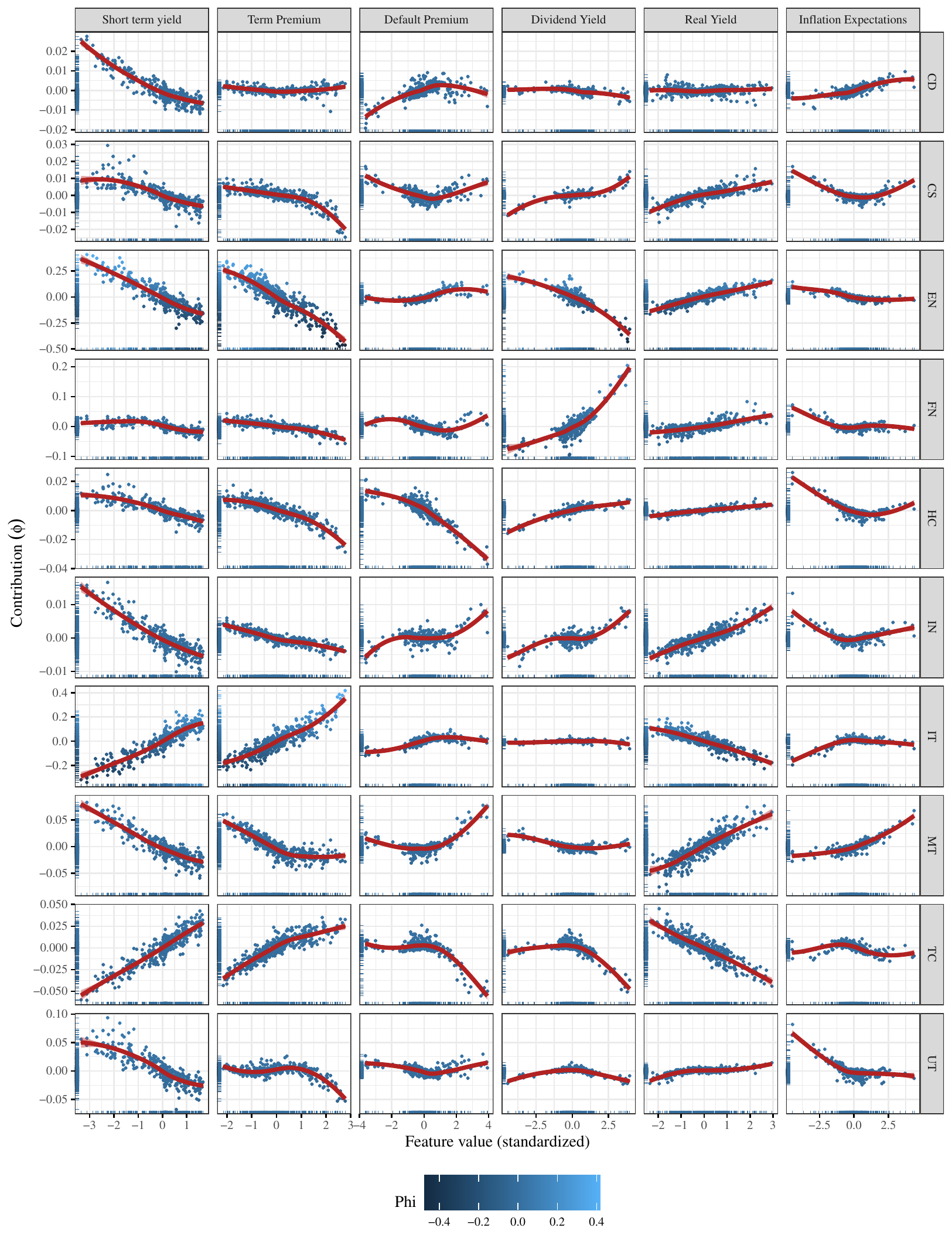}
\caption{Dependence plots of \(\beta_{tk}\) parameter estimates.
Contributions are calculated using Shapley values.\label{fig:capm_shap}}
\end{figure}

Fig. \ref{fig:capm_shap} visualizes the Shapley values for beta. The
explainability algorithm is applied only to the encoder portion of the
PENN, thus explaining comovement between the posterior mean of the
parameters and the input variables. The figure plots contributions
(\(\phi\)) against the standardized feature values. The presence of
nonlinear dependencies could be taken to indicate that the use of the
nonlinear version of the conditional CAPM is warranted, and that the
specification of the economic state is driven by nonlinear, potentially
interactive combinations of the macroeconomic variables.

The Shapley values reveal substantial nonlinear behavior for several
variables. The effects are not easily interpreted, since they do not
represent causal relationships. Generally, however, a reduction in
short-term yields and a flattening of the yield curve --- patterns often
accompanying economic crises --- are associated with increases in
systematic risk exposure (with some exceptions). The default premium,
dividend yield and inflation expectations have a mixed effect with
nonlinear threshold-type behavior in several instances. The real yield
has a comparatively low, linear effect.

\begin{figure}
\centering
\includegraphics{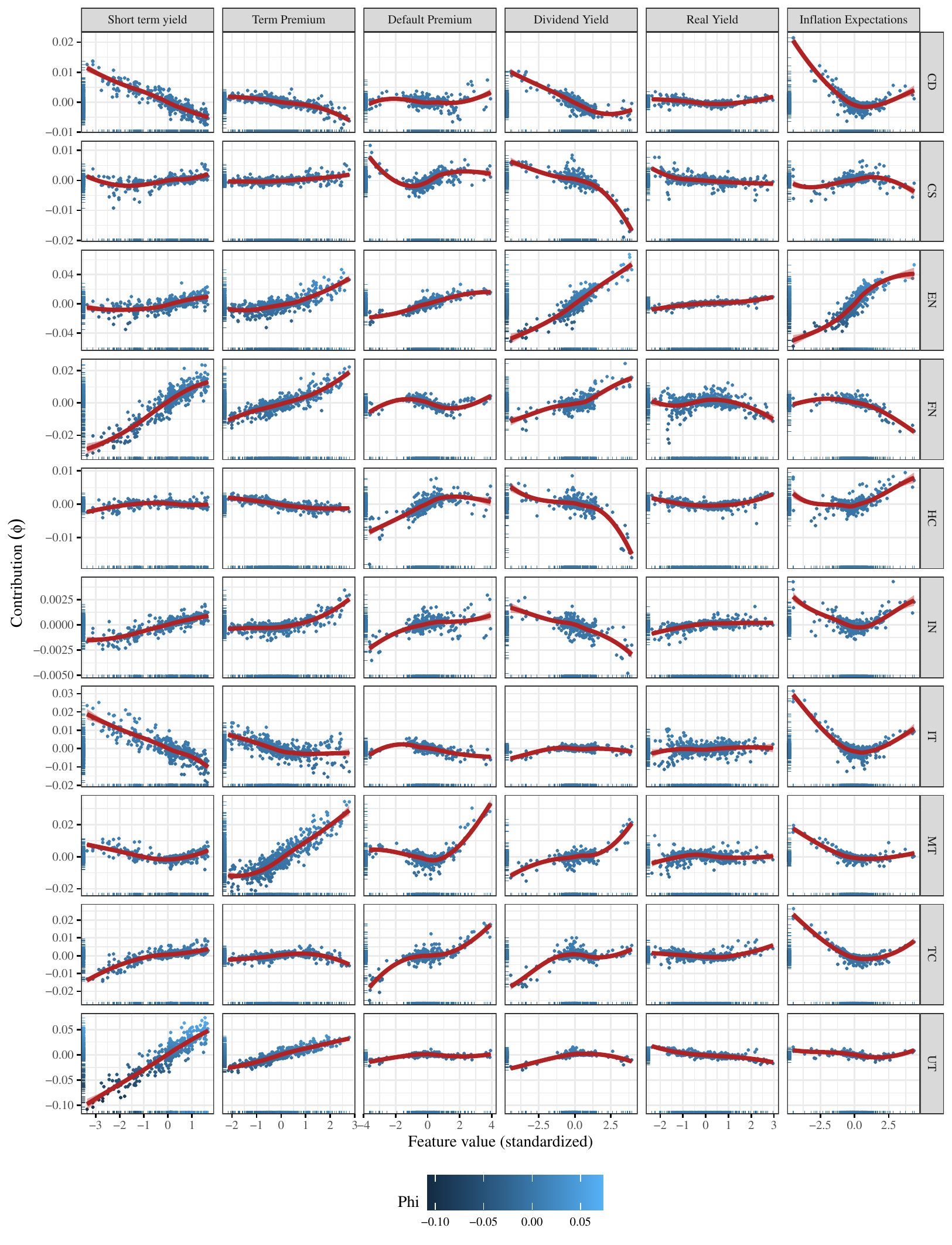}
\caption{Dependence plots of \(\alpha_{tk}\) parameter estimates.
Contributions are calculated using Shapley
values.\label{fig:capm_shap_alpha}}
\end{figure}

Fig. \ref{fig:capm_shap_alpha} plots the contributions for the posterior
mean of alpha. Some interesting observations include the large increase
in CD alpha associated with declines in short-term yields, or the
increase in FN alpha when dividend yields increase. Once again, the
contributions can not be interpreted causally, but confirm the nonlinear
shape of regime embedding observed in many cases and suggest that the
flexible neural network based estimate of the conditional CAPM is
warranted.

\hypertarget{concluding-remarks}{%
\section{\texorpdfstring{Concluding remarks
\label{conclusion}}{Concluding remarks }}\label{concluding-remarks}}

Model interpretability is oftentimes presented as an inverse function of
model complexity. On the one extreme lies the linear regression,
characterized by complete interpretability. On the other are machine
learning algorithms such as neural networks which --- while highly
flexible --- are opaque. \emph{Post hoc} algorithms that attempt to
understand the inner workings of uninterpretable models, typically shift
along the same continuum by superimposing a less complex and thus more
interpretable model onto the black box. This paper has introduced a
method that breaks with such a straightforward portrayal by generating
interpretable outcomes similar to those of a linear regression, but in
the context of a neural network. The PENN does not simply aim to
interpret a fitted model, but rather aims to understand an underlying
DGP, estimating posterior densities for a locally linear model
parameterization, that are capable of accounting for the rich feature
interactivity encoded in the neural network architecture.

Simulations illustrate that the PENN is capable of producing consistent
coefficient estimates and feature contributions in the presence of
dependent features, thus achieving a high degree of interpretability
without imposing an additive structure. In addition, an empirical
application demonstrates how the method can be deployed in the context
of econometric analysis, both to explore nonlinear parameter behavior
and to conduct parameter inference. A nonlinear version of the
conditional CAPM is estimated, which is capable of producing real-time
predictions of systematic and idiosyncratic risk components of financial
assets that are embedded in the economic regime. The results suggest a
substantial amount of nonlinear variation in the risk structure of
equity returns depending on the economic regime, as well as variation in
the extent to which the assumptions underlying the CAPM are met.
Specifically, the theoretical assumptions underlying the CAPM appear to
be violated with a high probability during economic downturns and
financial crises.

As an explainability concept, the PENN is interesting insofar as it
achieves interpretability without inherently sacrificing complexity. The
trend in explainability algorithms arguably goes towards model-agnostic
algorithms, such as SHAP or LIME
(\protect\hyperlink{ref-molnarInterpretableMachineLearning2020}{Molnar,
2020}). Nonetheless, by embedding the concept of interpretability into
the neural network loss function, the need to impose a simpler
explainability model is obviated. While the difference may appear
trivial, it is precisely to capture non-additive behavior that a neural
network is generally deployed. If the econometrician's objective is to
explore a presumed nonlinear additive DGP, a sufficient toolkit of
interpretable and more readily implemented methods exist that are fit
for the task. Using a neural network to explore a complex DGP must be
accompanied by an explainability approach that is equally complex.

As an econometric technique, the PENN furthermore offers the ability to
perform inference in an extremely flexible environment, capable of
capturing asymmetries, thresholds, regime-changes and many other types
of nonlinear behavior. The flexibility of the underlying inference
network facilitates the estimation of local parameters without imposing
onerous assumptions on the DGP. The approach to regularization is highly
intuitive in the form of shrinkage towards the static linear solution.
This permits the PENN to be employed when data availability is
comparatively limited, and makes the task of model training more
transparent, with the complexity of the neural network directly
observable in the heterogeneity of the posterior densities.

\newpage

\hypertarget{references}{%
\section*{References}\label{references}}
\addcontentsline{toc}{section}{References}

\hypertarget{refs}{}
\begin{CSLReferences}
\leavevmode\hypertarget{ref-aasExplainingIndividualPredictions2020}{}%
Aas, K., Jullum, M. \& Løland, A. 2020. \emph{Explaining {Individual
Predictions When Features Are Dependent}: {More Accurate Approximations}
to {Shapley Values}}. (Paper 1903.10464). {arXiv.org}.

\leavevmode\hypertarget{ref-adrianLearningBetaTimeVarying2004}{}%
Adrian, T. \& Franzoni, F. 2004. \emph{Learning about {Beta}:
{Time}-{Varying Factor Loadings}, {Expected Returns}, and the
{Conditional CAPM}}. (Staff Report 193). {Federal Reserve Bank of New
York}.

\leavevmode\hypertarget{ref-allaireKerasInterfaceKeras2020}{}%
Allaire, J. \& Chollet, F. 2020. \emph{Keras: {R Interface} to
"{Keras}"}. (R Package Version 2.3.0.0).

\leavevmode\hypertarget{ref-allaireTensorflowInterfaceTensorFlow2019}{}%
Allaire, J. \& Tang, Y. 2019. \emph{Tensorflow: {R Interface} to
"{TensorFlow}"}. (R Package Version 2.0.0).

\leavevmode\hypertarget{ref-al-shedivatContextualExplanationNetworks2017}{}%
Al-Shedivat, M., Dubey, A. \& Xing, E.P. 2017. \emph{Contextual
{Explanation Networks}}. (Paper 1705.10301). {arXiv.org}.

\leavevmode\hypertarget{ref-alvarez-melisRobustnessInterpretabilityMethods2018}{}%
Alvarez-Melis, D. \& Jaakkola, T.S. 2018. \emph{On the {Robustness} of
{Interpretability Methods}}. (Paper 1806.08049). {arXiv.org}.

\leavevmode\hypertarget{ref-arjovskyPrincipledMethodsTraining2017}{}%
Arjovsky, M. \& Bottou, L. 2017. \emph{Towards {Principled Methods} for
{Training Generative Adversarial Networks}}. (Paper 1701.04862).
{arXiv.org}.

\leavevmode\hypertarget{ref-atheyApproximateResidualBalancing2018}{}%
Athey, S., Imbens, G.W. \& Wager, S. 2018. Approximate {Residual
Balancing}: {Debiased Inference} of {Average Treatment Effects} in {High
Dimensions}. \emph{Journal of the Royal Statistical Society: Series B
(Statistical Methodology)}. 80(4):597--623.

\leavevmode\hypertarget{ref-atheyGeneralizedRandomForests2019}{}%
Athey, S., Tibshirani, J. \& Wager, S. 2019. Generalized {Random
Forests}. \emph{The Annals of Statistics}. 47(2).

\leavevmode\hypertarget{ref-belloniInferenceTreatmentEffects2014}{}%
Belloni, A., Chernozhukov, V. \& Hansen, C. 2014. Inference on
{Treatment Effects} after {Selection} among {High}-{Dimensional
Controls}. \emph{The Review of Economic Studies}. 81(2):608--650.

\leavevmode\hypertarget{ref-belloniProgramEvaluationCausal2017}{}%
Belloni, A., Chernozhukov, V., Fernandez-Val, I. \& Hansen, C. 2017.
Program {Evaluation} and {Causal Inference With High}-{Dimensional
Data}. \emph{Econometrica}. 85(1):233--298.

\leavevmode\hypertarget{ref-bergmeirUseCrossValidationTime2012}{}%
Bergmeir, C. \& Benítez, J.M. 2012. On the {Use} of {Cross}-{Validation}
for {Time Series Predictor Evaluation}. \emph{Information Sciences}.
191:192--213.

\leavevmode\hypertarget{ref-bleiVariationalInferenceReview2017}{}%
Blei, D.M., Kucukelbir, A. \& McAuliffe, J.D. 2017. Variational
{Inference}: {A Review} for {Statisticians}. \emph{Journal of the
American Statistical Association}. 112(518):859--877.

\leavevmode\hypertarget{ref-brackeMachineLearningExplainability2019}{}%
Bracke, P., Datta, A., Jung, C. \& Sen, S. 2019. \emph{Machine {Learning
Explainability} in {Finance}: {An Application} to {Default Risk
Analysis}}. (Staff Working Paper 816). {Bank of England}.

\leavevmode\hypertarget{ref-breimanRandomForests2001}{}%
Breiman, L. 2001. Random {Forests}. \emph{Machine Learning}. 45:5--32.

\leavevmode\hypertarget{ref-brownAnalysisInvestmentsManagement2012}{}%
Brown, K.C. \& Reilly, F.K. 2012. \emph{Analysis of {Investments} and
{Management} of {Portfolios}}. 10th ed., International ed ed. {Mason,
Ohio}: {South-Western}.

\leavevmode\hypertarget{ref-caruanaIntelligibleModelsHealthCare2015}{}%
Caruana, R., Lou, Y., Gehrke, J., Koch, P., Sturm, M. \& Elhadad, N.
2015. Intelligible {Models} for {HealthCare}: {Predicting Pneumonia
Risk} and {Hospital} 30-day {Readmission}. in \emph{Proceedings of the
21th {ACM SIGKDD International Conference} on {Knowledge Discovery} and
{Data Mining} - {KDD} '15} {Sydney, NSW, Australia}: {ACM Press}.
1721--1730.

\leavevmode\hypertarget{ref-chenSymmetricVariationalAutoencoder2017}{}%
Chen, L., Dai, S., Pu, Y., Li, C., Su, Q. \& Carin, L. 2017.
\emph{Symmetric {Variational Autoencoder} and {Connections} to
{Adversarial Learning}}. (Paper 1709.01846). {arXiv.org}.

\leavevmode\hypertarget{ref-chernozhukovDoubleDebiasedMachine2018}{}%
Chernozhukov, V., Chetverikov, D., Demirer, M., Duflo, E., Hansen, C.,
Newey, W. \& Robins, J. 2018. Double/{Debiased Machine Learning} for
{Treatment} and {Structural Parameters}. \emph{The Econometrics
Journal}. 21(1):C1--C68.

\leavevmode\hypertarget{ref-dattaAlgorithmicTransparencyQuantitative2016}{}%
Datta, A., Sen, S. \& Zick, Y. 2016. Algorithmic {Transparency} via
{Quantitative Input Influence}: {Theory} and {Experiments} with
{Learning Systems}. in \emph{2016 {IEEE Symposium} on {Security} and
{Privacy} ({SP})} {San Jose, CA}: {IEEE}. 598--617.

\leavevmode\hypertarget{ref-dybvigDifferentialInformationPerformance1985}{}%
Dybvig, P.H. \& Ross, S.A. 1985. Differential {Information} and
{Performance Measurement Using} a {Security Market Line}. \emph{The
Journal of Finance}. 40(2):383--399.

\leavevmode\hypertarget{ref-euRegulationEU20162016}{}%
EU. 2016. Regulation ({EU}) 2016/679 of the {European Parliament},
{Directive} 95/46/{EC} ({General Data Protection Regulation}).
\emph{Official Journal of the European Union}. (L119):1--88.

\leavevmode\hypertarget{ref-famaCrossSectionExpectedStock1992}{}%
Fama, E.F. \& French, K.R. 1992. The {Cross}-{Section} of {Expected
Stock Returns}. \emph{The Journal of Finance}. 47(2):427--465.

\leavevmode\hypertarget{ref-famaCommonRiskFactors1993}{}%
Fama, E.F. \& French, K.R. 1993. Common {Risk Factors} in the {Returns}
on {Stocks} and {Bonds}. \emph{Journal of Financial Economics}.
33(1):3--56.

\leavevmode\hypertarget{ref-famaSizeBooktoMarketFactors1995}{}%
Fama, E.F. \& French, K.R. 1995. Size and {Book}-to-{Market Factors} in
{Earnings} and {Returns}. \emph{The Journal of Finance}. 50(1):131--155.

\leavevmode\hypertarget{ref-farrellRobustInferenceAverage2015}{}%
Farrell, M.H. 2015. Robust {Inference} on {Average Treatment Effects}
with {Possibly} more {Covariates} than {Observations}. \emph{Journal of
Econometrics}. 189(1):1--23.

\leavevmode\hypertarget{ref-farrellDeepNeuralNetworks2021}{}%
Farrell, M.H., Liang, T. \& Misra, S. 2021. Deep {Neural Networks} for
{Estimation} and {Inference}. \emph{Econometrica}. 89(1):181--213.

\leavevmode\hypertarget{ref-fernandez-delgadoWeNeedHundreds2014}{}%
Fernandez-Delgado, M., Cernadas, E., Barro, S. \& Amorim, D. 2014. Do we
{Need Hundreds} of {Classifiers} to {Solve Real World Classification
Problems}? \emph{Journal of Machine Learning Research}. 15:3133--3181.

\leavevmode\hypertarget{ref-finlayScaleableInputGradient2021}{}%
Finlay, C. \& Oberman, A.M. 2021. Scaleable {Input Gradient
Regularization} for {Adversarial Robustness}. \emph{Machine Learning
with Applications}. 3.

\leavevmode\hypertarget{ref-fisherAllModelsAre2019}{}%
Fisher, A., Rudin, C. \& Dominici, F. 2019. \emph{All {Models Are
Wrong}, but {Many Are Useful}: {Learning} a {Variable}'s {Importance} by
{Studying} an {Entire Class} of {Prediction Models Simultaneously}}.
(Paper 1801.01489). {arXiv.org}.

\leavevmode\hypertarget{ref-friedbergLocalLinearForests2020}{}%
Friedberg, R., Tibshirani, J., Athey, S. \& Wager, S. 2020. \emph{Local
{Linear Forests}}. (Paper 1807.11408). {arXiv.org}.

\leavevmode\hypertarget{ref-friedmanGreedyFunctionApproximation2001}{}%
Friedman, J. 2001. Greedy {Function Approximation}: {A Gradient Boosting
Machine}. \emph{The Annals of Statistics}. 29(5):1189--1232.

\leavevmode\hypertarget{ref-glovaTimeVaryingCAPMIts2015}{}%
Glova, J. 2015. Time-{Varying CAPM} and {Its Applicability} in {Cost} of
{Equity Determination}. \emph{Procedia Economics and Finance}.
32:60--67.

\leavevmode\hypertarget{ref-goldsteinPeekingBlackBox2014}{}%
Goldstein, A., Kapelner, A., Bleich, J. \& Pitkin, E. 2014.
\emph{Peeking {Inside} the {Black Box}: {Visualizing Statistical
Learning} with {Plots} of {Individual Conditional Expectation}}. (Paper
1309.6392v2). {arXiv.org}.

\leavevmode\hypertarget{ref-goodfellowDeepLearning2016}{}%
Goodfellow, I., Bengio, Y. \& Courville, A. 2016. \emph{Deep
{Learning}}. Illustrated ed. (Adaptive {Computation} and {Machine
Learning}). {Cambridge, Massachusetts}: {The MIT Press}.

\leavevmode\hypertarget{ref-hansenEconometrics2019}{}%
Hansen, B.E. 2019. \emph{Econometrics}. Draft ed. {University of
Wisconsin}.

\leavevmode\hypertarget{ref-hansenRoleConditioningInformation1987}{}%
Hansen, L.P. \& Richard, S.F. 1987. The {Role} of {Conditioning
Information} in {Deducing Testable Restrictions Implied} by {Dynamic
Asset Pricing Models}. \emph{Econometrica}. 55(3):587--613.

\leavevmode\hypertarget{ref-hastieGeneralizedAdditiveModels1986}{}%
Hastie, T. \& Tibshirani, R. 1986. Generalized {Additive Models}.
\emph{Statistical Science}. 1(3):297--310.

\leavevmode\hypertarget{ref-hastieGeneralizedAdditiveModels1987}{}%
Hastie, T. \& Tibshirani, R. 1987. Generalized {Additive Models}: {Some
Applications}. \emph{Journal of the American Statistical Association}.
82(398):371--386.

\leavevmode\hypertarget{ref-hastieGeneralizedAdditiveModels1999}{}%
Hastie, T. \& Tibshirani, R. 1999. \emph{Generalized {Additive Models}}.
First ed. {Boca Raton, Fla}: {Chapman \& Hall/CRC}.

\leavevmode\hypertarget{ref-higginsVVAELearningBasic2017}{}%
Higgins, I., Matthey, L., Pal, A., Burgess, C., Glorot, X., Botvinick,
M., Mohamed, S. \& Lerchner, A. 2017. {\(\beta\)}-{VAE}: {Learning Basic
Visual Concepts} with a {Constrained Variational Framework}. \emph{ICLR
Conference Paper}.

\leavevmode\hypertarget{ref-horelSignificanceTestsNeural2020}{}%
Horel, E. \& Giesecke, K. 2020. Significance {Tests} for {Neural
Networks}. \emph{Journal of Machine Learning Research}. 21:1--29.

\leavevmode\hypertarget{ref-hornikMultilayerFeedforwardNetworks1989}{}%
Hornik, K., Stinchcombe, M. \& White, H. 1989. Multilayer {Feedforward
Networks Are Universal Approximators}. \emph{Neural Networks}.
2(5):359--366.

\leavevmode\hypertarget{ref-jagannathanConditionalCAPMCrossSection1996}{}%
Jagannathan, R. \& Wang, Z. 1996. The {Conditional CAPM} and the
{Cross}-{Section} of {Expected Returns}. \emph{The Journal of Finance}.
51(1):3--53.

\leavevmode\hypertarget{ref-jensenPerformanceMutualFunds1968}{}%
Jensen, M.C. 1968. The {Performance} of {Mutual Funds} in the {Period}
1945-1964. \emph{The Journal of Finance}. 23(2):389--416.

\leavevmode\hypertarget{ref-jordanIntroductionVariationalMethods1999}{}%
Jordan, M.I., Ghahramani, Z., Jaakkola, T.S. \& Saul, L.K. 1999. An
{Introduction} to {Variational Methods} for {Graphical Models}.
\emph{Machine Learning}. 37:183--233.

\leavevmode\hypertarget{ref-josephShapleyRegressionsFramework2019a}{}%
Joseph, A. 2019. \emph{Shapley {Regressions}: {A Framework} for
{Statistical Inference} on {Machine Learning Models}}. (Staff Working
Paper 784). {Bank of England}.

\leavevmode\hypertarget{ref-kaufmanFindingGroupsData2005}{}%
Kaufman, L. \& Rousseeuw, P.J. 2005. \emph{Finding {Groups} in {Data}:
{An Introduction} to {Cluster Analysis}}. First ed. (Wiley series in
probability and mathematical statistics). {Hoboken, N.J}: {Wiley}.

\leavevmode\hypertarget{ref-khannaEconomyStatisticalRecurrent2020}{}%
Khanna, S. \& Tan, V.Y.F. 2020. \emph{Economy {Statistical Recurrent
Units For Inferring Nonlinear Granger Causality}}. (Paper 1911.09879).
{arXiv.org}.

\leavevmode\hypertarget{ref-kingmaAdamMethodStochastic2017}{}%
Kingma, D.P. \& Ba, J. 2017. \emph{Adam: {A Method} for {Stochastic
Optimization}}. (Paper 1412.6980). {arXiv.org}.

\leavevmode\hypertarget{ref-kingmaAutoEncodingVariationalBayes2014}{}%
Kingma, D.P. \& Welling, M. 2014. \emph{Auto-{Encoding Variational
Bayes}}. (Paper 1312.6114). {arXiv.org}.

\leavevmode\hypertarget{ref-koutmosEstimatingSystematicRisk2002}{}%
Koutmos, G. \& Knif, J. 2002. Estimating {Systematic Risk Using Time
Varying Distributions}. \emph{European Financial Management}.
8(1):59--73.

\leavevmode\hypertarget{ref-lewellenConditionalCAPMDoes2006}{}%
Lewellen, J. \& Nagel, S. 2006. The {Conditional CAPM Does Not Explain
Asset}-{Pricing Anomalies}. \emph{Journal of Financial Economics}.
82:289--314.

\leavevmode\hypertarget{ref-lintnerValuationRiskAssets1965}{}%
Lintner, J. 1965. The {Valuation} of {Risk Assets} and the {Selection}
of {Risky Investments} in {Stock Portfolios} and {Capital Budgets}.
\emph{The Review of Economics and Statistics}. 47(1):13--37.

\leavevmode\hypertarget{ref-lisboaEfficientEstimationGeneral2020}{}%
Lisboa, P.J.G., Ortega-Martorell, S., Jayabalan, M. \& Olier, I. 2020.
Efficient {Estimation} of {General Additive Neural Networks}: {A Case
Study} for {CTG Data}. in \emph{{ECML PKDD} 2020 {Workshops}} I.
Koprinska, M. Kamp, A. Appice, C. Loglisci, L. Antonie, A. Zimmermann,
R. Guidotti, Ö. Özgöbek, et al. (eds.). {Cham}: {Springer International
Publishing} I. Koprinska, M. Kamp, A. Appice, C. Loglisci, L. Antonie,
A. Zimmermann, R. Guidotti, Ö. Özgöbek, et al. (eds.). 432--446.

\leavevmode\hypertarget{ref-loweAmortizedCausalDiscovery2020}{}%
Löwe, S., Madras, D., Zemel, R. \& Welling, M. 2020. \emph{Amortized
{Causal Discovery}: {Learning} to {Infer Causal Graphs} from
{Time}-{Series Data}}. (Paper 2006.10833). {arXiv.org}.

\leavevmode\hypertarget{ref-lundbergUnifiedApproachInterpreting2017}{}%
Lundberg, S.M. \& Lee, S.-I. 2017. A {Unified Approach} to {Interpreting
Model Predictions}. in \emph{31st {Conference} on {Neural Information
Processing Systems} ({NIPS} 2017)} {Long Beach, CA, USA}.

\leavevmode\hypertarget{ref-lustigHousingCollateralConsumption2005}{}%
Lustig, H.N. \& Van Nieuwerburgh, S.G. 2005. Housing {Collateral},
{Consumption Insurance}, and {Risk Premia}: {An Empirical Perspective}.
\emph{The Journal of Finance}. 60(3):1167--1219.

\leavevmode\hypertarget{ref-marcinkevicsInterpretableModelsGranger2021}{}%
Marcinkevičs, R. \& Vogt, J.E. 2021. \emph{Interpretable {Models} for
{Granger Causality Using Self}-explaining {Neural Networks}}. (Paper
2101.07600). {arXiv.org}.

\leavevmode\hypertarget{ref-markowitzPortfolioSelection1952a}{}%
Markowitz, H. 1952. Portfolio {Selection}. \emph{The Journal of
Finance}. 7(1):77--91.

\leavevmode\hypertarget{ref-melisRobustInterpretabilitySelfExplaining2018}{}%
Melis, D.A. \& Jaakkola, T. 2018. Towards {Robust Interpretability} with
{Self}-{Explaining Neural Networks}. in \emph{32nd {Conference} on
{Neural Information Processing Systems} ({NeurIPS} 2018)} {Montreal,
Canada}.

\leavevmode\hypertarget{ref-molnarImlPackageInterpretable2018}{}%
Molnar, C. 2018. Iml: {An R} package for {Interpretable Machine
Learning}. \emph{Journal of Open Source Software}. 3(26):786.

\leavevmode\hypertarget{ref-molnarInterpretableMachineLearning2020}{}%
Molnar, C. 2020. \emph{Interpretable {Machine Learning}}. Online ed.
{lulu.com}.

\leavevmode\hypertarget{ref-montavonMethodsInterpretingUnderstanding2018}{}%
Montavon, G., Samek, W. \& Müller, K.-R. 2018. Methods for
{Interpreting} and {Understanding Deep Neural Networks}. \emph{Digital
Signal Processing}. 73:1--15.

\leavevmode\hypertarget{ref-mullainathanMachineLearningApplied2017}{}%
Mullainathan, S. \& Spiess, J. 2017. Machine {Learning}: {An Applied
Econometric Approach}. \emph{Journal of Economic Perspectives}.
31(2):87--106.

\leavevmode\hypertarget{ref-nautaCausalDiscoveryAttentionBased2019}{}%
Nauta, M., Bucur, D. \& Seifert, C. 2019. Causal {Discovery} with
{Attention}-{Based Convolutional Neural Networks}. \emph{Machine
Learning and Knowledge Extraction}. 1(1):312--340.

\leavevmode\hypertarget{ref-petkovaValueRiskierGrowth2005}{}%
Petkova, R. \& Zhang, L. 2005. Is {Value Riskier Than Growth}?
\emph{Journal of Financial Economics}. 78:187--202.

\leavevmode\hypertarget{ref-pottsGeneralizedAdditiveNeural1999}{}%
Potts, W.J.E. 1999. Generalized {Additive Neural Networks}. in
\emph{Proceedings of the fifth {ACM SIGKDD} international conference on
{Knowledge} discovery and data mining - {KDD} '99} {San Diego,
California, United States}: {ACM Press}. 194--200.

\leavevmode\hypertarget{ref-puAdversarialSymmetricVariational2017}{}%
Pu, Y., Wang, W., Henao, R., Chen, L., Gan, Z., Li, C. \& Carin, L.
2017. \emph{Adversarial {Symmetric Variational Autoencoder}}. (Paper
1711.04915). {arXiv.org}.

\leavevmode\hypertarget{ref-racineConsistentCrossValidatoryModelSelection2000}{}%
Racine, J. 2000. Consistent {Cross}-{Validatory Model}-{Selection} for
{Dependent Data}: Hv-{Block Cross}-{Validation}. \emph{Journal of
Econometrics}. 99(1):39--61.

\leavevmode\hypertarget{ref-rezendeStochasticBackpropagationApproximate2014}{}%
Rezende, D.J., Mohamed, S. \& Wierstra, D. 2014. \emph{Stochastic
{Backpropagation} and {Approximate Inference} in {Deep Generative
Models}}. (Paper 1401.4082). {arXiv.org}.

\leavevmode\hypertarget{ref-ribeiroWhyShouldTrust2016}{}%
Ribeiro, M.T., Singh, S. \& Guestrin, C. 2016. "{Why Should I Trust
You}?": {Explaining} the {Predictions} of {Any Classifier}. in
\emph{Proceedings of the 22nd {ACM SIGKDD International Conference} on
{Knowledge Discovery} and {Data Mining} - {KDD} '16} {San Francisco,
California, USA}: {ACM Press}. 1135--1144.

\leavevmode\hypertarget{ref-rossiCapitalAssetPricing2016}{}%
Rossi, M. 2016. The {Capital Asset Pricing Model}: {A Critical
Literature Review}. \emph{Global Business and Economics Review}.
18(5):604--617.

\leavevmode\hypertarget{ref-rudinStopExplainingBlack2019}{}%
Rudin, C. 2019. Stop {Explaining Black Box Machine Learning Models} for
{High Stakes Decisions} and {Use Interpretable Models Instead}.
\emph{Nature Machine Intelligence}. 1:206--215.

\leavevmode\hypertarget{ref-santosLaborIncomePredictable2006}{}%
Santos, T. \& Veronesi, P. 2006. Labor {Income} and {Predictable Stock
Returns}. \emph{The Review of Financial Studies}. 19(1):1--44.

\leavevmode\hypertarget{ref-shapleyValueNPersonGames1953}{}%
Shapley, L. 1953. A {Value} for {N}-{Person Games}. \emph{Contributions
to the Theory of Games}. 2:307--317.

\leavevmode\hypertarget{ref-sharpeCapitalAssetPrices1964}{}%
Sharpe, W.F. 1964. Capital {Asset Prices}: {A Theory} of {Market
Equilibrium Under Conditions} of {Risk}. \emph{The Journal of Finance}.
19(3):425--442.

\leavevmode\hypertarget{ref-shrikumarLearningImportantFeatures2019}{}%
Shrikumar, A., Greenside, P. \& Kundaje, A. 2019. \emph{Learning
{Important Features Through Propagating Activation Differences}}. (Paper
1704.02685). {arXiv.org}.

\leavevmode\hypertarget{ref-strumbeljExplainingPredictionModels2014}{}%
Štrumbelj, E. \& Kononenko, I. 2014. Explaining {Prediction Models} and
{Individual Predictions} with {Feature Contributions}. \emph{Knowledge
and Information Systems}. 41(3):647--665.

\leavevmode\hypertarget{ref-tankInterpretableSparseNeural2018}{}%
Tank, A., Cover, I., Foti, N.J., Shojaie, A. \& Fox, E.B. 2018. \emph{An
{Interpretable} and {Sparse Neural Network Model} for {Nonlinear Granger
Causality Discovery}}. (Paper 1711.08160). {arXiv.org}.

\leavevmode\hypertarget{ref-tiffinMachineLearningCausality2019}{}%
Tiffin, A.J. 2019. \emph{Machine {Learning} and {Causality}: {The
Impact} of {Financial Crises} on {Growth}}. (IMF Working Paper No.
19/228). {International Monetary Fund}.

\leavevmode\hypertarget{ref-varianBigDataNew2014}{}%
Varian, H.R. 2014. Big {Data}: {New Tricks} for {Econometrics}.
\emph{Journal of Economic Perspectives}. 28(2):3--28.

\leavevmode\hypertarget{ref-wagerEstimationInferenceHeterogeneous2018}{}%
Wager, S. \& Athey, S. 2018. Estimation and {Inference} of
{Heterogeneous Treatment Effects} using {Random Forests}. \emph{Journal
of the American Statistical Association}. 113(523):1228--1242.

\leavevmode\hypertarget{ref-woodFastStableRestricted2011}{}%
Wood, S.N. 2011. Fast {Stable Restricted Maximum Likelihood} and
{Marginal Likelihood Estimation} of {Semiparametric Generalized Linear
Models}: {Estimation} of {Semiparametric Generalized Linear Models}.
\emph{Journal of the Royal Statistical Society: Series B (Statistical
Methodology)}. 73(1):3--36.

\leavevmode\hypertarget{ref-wuDiscoveringNonlinearRelations2020}{}%
Wu, T., Breuel, T., Skuhersky, M. \& Kautz, J. 2020. \emph{Discovering
{Nonlinear Relations} with {Minimum Predictive Information
Regularization}}. (Paper 2001.01885). {arXiv.org}.

\leavevmode\hypertarget{ref-zhangAdvancesVariationalInference2018}{}%
Zhang, C., Butepage, J., Kjellstrom, H. \& Mandt, S. 2018.
\emph{Advances in {Variational Inference}}. (Paper 1711.05597).
{arXiv.org}.

\end{CSLReferences}

\newpage

\appendix

\hypertarget{proof-of-proposition}{%
\section{\texorpdfstring{Proof of Proposition \ref{prop:penn}
\label{appa}}{Proof of Proposition  }}\label{proof-of-proposition}}

Let \(q_{\boldsymbol{\theta}}(\boldsymbol{\beta}|\boldsymbol{x})\) be an
inference network that approximates a posterior density of
\(\boldsymbol{\beta}\), \(p(\boldsymbol{\beta}|y, \boldsymbol{x})\). The
density \(q_{\boldsymbol{\theta}}(\boldsymbol{\beta}|\boldsymbol{x})\)
follows a mean field variational distribution with

\begin{equation}
q_{\boldsymbol{\theta}}(\boldsymbol{\beta}|\boldsymbol{x}) = \prod_{i=1}^N q_{\boldsymbol{\theta}}(\boldsymbol{\beta}_i | \boldsymbol{x}_i),
\end{equation}

and

\begin{equation}
q_{\boldsymbol{\theta}}(\boldsymbol{\beta}_i | \boldsymbol{x}_i) = \mathcal{N}(\mu_{\boldsymbol{\theta}}(\boldsymbol{x}_i), \boldsymbol{\Sigma}_{\boldsymbol{x}_i}), \;\;\; \boldsymbol{\Sigma}_{\boldsymbol{x}_i} = \text{Diag}\left(\sigma_{\boldsymbol{\theta}}^2(\boldsymbol{x}_i)\right),
\end{equation}

where \(i \in 1,...,N\). The functions
\(\mu_{\boldsymbol{\theta}}(\cdot)\) and
\(\sigma_{\boldsymbol{\theta}}^2(\cdot)\) are components of the
inference network that return parameters for
\(q_{\boldsymbol{\theta}}(\boldsymbol{\beta}_i | \boldsymbol{x}_i)\)
given data.

The information difference between the approximate and true posteriors
can be measured using a Kullback-Leibler divergence, \(D_{KL}\), with:

\begin{equation}
D_{KL}(q_{\boldsymbol{\theta}}(\boldsymbol{\beta}_i| \boldsymbol{x}_i) || p(\boldsymbol{\beta}_i|y_i, \boldsymbol{x}_i)) = - \int q_{\boldsymbol{\theta}}(\boldsymbol{\beta}_i| \boldsymbol{x}_i) \log \frac{p(\boldsymbol{\beta}_i|y_i, \boldsymbol{x}_i)}{q_{\boldsymbol{\theta}}(\boldsymbol{\beta}_i| \boldsymbol{x}_i)} d\boldsymbol{\beta}_i \geq 0.
\label{eq:app_kl}
\end{equation}

The LHS of Eq. \ref{eq:app_kl} is simply referred to \(D_{KL}\) below.
Applying Bayes' rule to Eq. \ref{eq:app_kl} results in

\begin{equation}
D_{KL} = - \int q_{\boldsymbol{\theta}}(\boldsymbol{\beta}_i| \boldsymbol{x}_i) \log \frac{p(y_i | \boldsymbol{\beta}_i, \boldsymbol{x}_i) p(\boldsymbol{\beta}_i | \boldsymbol{x}_i)}{q_{\boldsymbol{\theta}}(\boldsymbol{\beta}_i| \boldsymbol{x}_i) p(y_i | \boldsymbol{x}_i)} d\boldsymbol{\beta}_i.
\end{equation}

Next, use the law of logarithms and distribute the integrand:

\begin{align}
D_{KL} &= - \int q_{\boldsymbol{\theta}}(\boldsymbol{\beta}_i| \boldsymbol{x}_i) \Big[ \log \frac{p(y_i | \boldsymbol{\beta}_i, \boldsymbol{x}_i) p(\boldsymbol{\beta}_i | \boldsymbol{x}_i)}{q_{\boldsymbol{\theta}}(\boldsymbol{\beta}_i| \boldsymbol{x}_i)} - \log p(y_i | \boldsymbol{x}_i) \Big] d\boldsymbol{\beta}_i\\
&= - \int q_{\boldsymbol{\theta}}(\boldsymbol{\beta}_i| \boldsymbol{x}_i) \log \frac{p(y_i | \boldsymbol{\beta}_i, \boldsymbol{x}_i) p(\boldsymbol{\beta}_i | \boldsymbol{x}_i)}{q_{\boldsymbol{\theta}}(\boldsymbol{\beta}_i| \boldsymbol{x}_i)} d\boldsymbol{\beta}_i + \int q_{\boldsymbol{\theta}}(\boldsymbol{\beta}_i| \boldsymbol{x}_i) \log p(y_i | \boldsymbol{x}_i) d\boldsymbol{\beta}_i.
\label{eq:app_distributed_integrand}
\end{align}

The term \(\log p(y_i|\boldsymbol{x}_i)\) can be removed from the second
integral in Eq. \ref{eq:app_distributed_integrand}, since it does not
depend on \(\boldsymbol{\beta}_i\), such that:

\begin{equation}
D_{KL} = - \int q_{\boldsymbol{\theta}}(\boldsymbol{\beta}_i| \boldsymbol{x}_i) \log \frac{p(y_i | \boldsymbol{\beta}_i, \boldsymbol{x}_i) p(\boldsymbol{\beta}_i | \boldsymbol{x}_i)}{q_{\boldsymbol{\theta}}(\boldsymbol{\beta}_i| \boldsymbol{x}_i)} d\boldsymbol{\beta}_i + \log p(y_i | \boldsymbol{x}_i) \int q_{\boldsymbol{\theta}}(\boldsymbol{\beta}_i| \boldsymbol{x}_i) d\boldsymbol{\beta}_i.
\label{eq:app_logp}
\end{equation}

Given that
\(q_{\boldsymbol{\theta}}(\boldsymbol{\beta}_i| \boldsymbol{x}_i)\) is a
probability distribution and integrates to one, Eq. \ref{eq:app_logp}
can be simplified further:

\begin{equation}
D_{KL} = - \int q_{\boldsymbol{\theta}}(\boldsymbol{\beta}_i| \boldsymbol{x}_i) \log \frac{p(y_i | \boldsymbol{\beta}_i, \boldsymbol{x}_i) p(\boldsymbol{\beta}_i | \boldsymbol{x}_i)}{q_{\boldsymbol{\theta}}(\boldsymbol{\beta}_i| \boldsymbol{x}_i)} d\boldsymbol{\beta}_i + \log p(y_i | \boldsymbol{x}_i).
\label{eq:app_logp2}
\end{equation}

Now, \(\log p(y_i | \boldsymbol{x}_i)\) is moved to the LHS, and the law
of logarithms is applied, once again, on the RHS, followed by a
distribution of the integrand:

\begin{align}
D_{KL} - \log p&(y_i | \boldsymbol{x}_i) \nonumber \\
&= - \int q_{\boldsymbol{\theta}}(\boldsymbol{\beta}_i| \boldsymbol{x}_i) \Big[ \log p(y_i | \boldsymbol{\beta}_i, \boldsymbol{x}_i)  + \log \frac{ p(\boldsymbol{\beta}_i | \boldsymbol{x}_i)}{q_{\boldsymbol{\theta}}(\boldsymbol{\beta}_i| \boldsymbol{x}_i)} \Big] d\boldsymbol{\beta}_i \\
&= - \int q_{\boldsymbol{\theta}}(\boldsymbol{\beta}_i| \boldsymbol{x}_i) \log p(y_i | \boldsymbol{\beta}_i, \boldsymbol{x}_i)d\boldsymbol{\beta}_i  - \int q_{\boldsymbol{\theta}}(\boldsymbol{\beta}_i| \boldsymbol{x}_i) \log \frac{ p(\boldsymbol{\beta}_i | \boldsymbol{x}_i)}{q_{\boldsymbol{\theta}}(\boldsymbol{\beta}_i| \boldsymbol{x}_i)} d\boldsymbol{\beta}_i.
\label{eq:app_distributed_integrand2}
\end{align}

The first term on the RHS of Eq. \ref{eq:app_distributed_integrand2} is
an expectation, while the second represents another Kullback-Leibler
divergence. Adjusting Eq. \ref{eq:app_distributed_integrand2} to reflect
this, results in:

\begin{equation}
D_{KL} - \log p(y_i | \boldsymbol{x}_i) = 
- \mathbb{E}_{\boldsymbol{\beta}_i \sim q_{\boldsymbol{\theta}}(\boldsymbol{\beta}_i| \boldsymbol{x}_i)} \log p(y_i | \boldsymbol{\beta}_i, \boldsymbol{x}_i)  + D_{KL}(q_{\boldsymbol{\theta}}(\boldsymbol{\beta}_i| \boldsymbol{x}_i) || p(\boldsymbol{\beta}_i|\boldsymbol{x}_i)).
\label{eq:app_final_ik}
\end{equation}

Eq. \ref{eq:app_final_ik} can be rearranged to yield the disaggregated
form of the equation in Proposition \ref{prop:penn}:

\begin{align}
\log p(y_i | \boldsymbol{x}_i) - D_{KL}(&q_{\boldsymbol{\theta}}(\boldsymbol{\beta}_i| \boldsymbol{x}_i) || p(\boldsymbol{\beta}_i|y_i, \boldsymbol{x}_i)) = \nonumber \\
&\mathbb{E}_{\boldsymbol{\beta}_i \sim q_{\boldsymbol{\theta}}(\boldsymbol{\beta}_i| \boldsymbol{x}_i)} \log p(y_i | \boldsymbol{\beta}_i, \boldsymbol{x}_i)  - D_{KL}(q_{\boldsymbol{\theta}}(\boldsymbol{\beta}_i| \boldsymbol{x}_i) || p(\boldsymbol{\beta}_i|\boldsymbol{x}_i)).
\label{eq:app_final_ik2}
\end{align}

Finally, aggregating Eq. \ref{eq:app_final_ik2} over \(N\) results in
the equation in Proposition \ref{prop:penn}:

\begin{align}
\log p(y | \boldsymbol{x}) - &\sum_{i=1}^N D_{KL}(q_{\boldsymbol{\theta}}(\boldsymbol{\beta}_i| \boldsymbol{x}_i) || p(\boldsymbol{\beta}_i|y_i, \boldsymbol{x}_i)) = \nonumber \\
&\mathbb{E}_{\boldsymbol{\beta} \sim q_{\boldsymbol{\theta}}(\boldsymbol{\beta}| \boldsymbol{x})} \log p(y | \boldsymbol{\beta}, \boldsymbol{x}) - \sum_{i=1}^N  D_{KL}(q_{\boldsymbol{\theta}}(\boldsymbol{\beta}_i| \boldsymbol{x}_i) || p(\boldsymbol{\beta}_i|\boldsymbol{x}_i)).
\end{align}

\newpage

\hypertarget{derivation-of-a-closed-form-kullback-leibler-loss}{%
\section{\texorpdfstring{Derivation of a closed form Kullback-Leibler
loss
\label{appb}}{Derivation of a closed form Kullback-Leibler loss }}\label{derivation-of-a-closed-form-kullback-leibler-loss}}

Let
\(q_{\boldsymbol{\theta}}(\beta_{ik}| \boldsymbol{x}_i) = \mathcal{N}(\mu^q_{ik}, \left[\sigma^q_{ik}\right]^2)\)
and
\(p(\beta_{ik}| \boldsymbol{x}_i) = \mathcal{N}(\mu^p_{ik}, \left[\sigma^p_{ik}\right]^2)\)
be the normally distributed approximate posterior and prior densities of
a parameter \(\beta_{ik} \in \boldsymbol{\beta}\), such that

\begin{equation}
q_{\boldsymbol{\theta}}(\beta_{ik}| \boldsymbol{x}_i) = \frac{1}{\sqrt{2\pi (\sigma^q_{ik})^2}} \exp \left( -\frac{(\beta_{ik} - \mu_{ik}^q)^2}{2(\sigma^q_{ik})^2} \right),
\end{equation}

and

\begin{equation}
p(\beta_{ik}| \boldsymbol{x}_i) = \frac{1}{\sqrt{2\pi (\sigma^p_{ik})^2}} \exp \left( -\frac{(\beta_{ik} - \mu_{ik}^p)^2}{2(\sigma^p_{ik})^2} \right).
\end{equation}

Given the definition of the Kullback-Leibler divergence, the
regularization term becomes:

\begin{align}
-D_{KL}(q_{\boldsymbol{\theta}}&(\beta_{ik}| \boldsymbol{x}_i) || p(\beta_{ik}|\boldsymbol{x}_i)) = \nonumber \\
&\int \frac{1}{\sqrt{2\pi (\sigma^q_{ik})^2}} \exp \left( -\frac{(\beta_{ik} - \mu_{ik}^q)^2}{2(\sigma^q_{ik})^2} \right) \log \left( \frac{\frac{1}{\sqrt{2\pi (\sigma^p_{ik})^2}} \exp \left( -\frac{(\beta_{ik} - \mu_{ik}^p)^2}{2(\sigma^p_{ik})^2} \right)}{\frac{1}{\sqrt{2\pi (\sigma^q_{ik})^2}} \exp \left( -\frac{(\beta_{ik} - \mu_{ik}^q)^2}{2(\sigma^q_{ik})^2} \right)} \right)d \beta_{ik}.
\end{align}

The term in the logarithm can be simplified, resulting in

\begin{align}
-&D_{KL}(q_{\boldsymbol{\theta}}(\beta_{ik}| \boldsymbol{x}_i) || p(\beta_{ik}|\boldsymbol{x}_i)) = \nonumber \\
&\frac{1}{\sqrt{2\pi (\sigma^q_{ik})^2}} \int \exp \left( -\frac{(\beta_{ik} - \mu_{ik}^q)^2}{2(\sigma^q_{ik})^2} \right) \left[ \log\left(\frac{\sigma_{ik}^q}{\sigma_{ik}^p}\right) -\frac{(\beta_{ik} - \mu_{ik}^p)^2}{2(\sigma^p_{ik})^2} +\frac{(\beta_{ik} - \mu_{ik}^q)^2}{2(\sigma^q_{ik})^2} \right]d \beta_{ik}.
\label{eq:appb_simplified}
\end{align}

Eq. \ref{eq:appb_simplified} can be expressed as an expectation and
transformed using the rules of the expectations operator, such that

\begin{align}
-D_{KL}(q_{\boldsymbol{\theta}}(\beta_{ik}| \boldsymbol{x}_i) &|| p(\beta_{ik}|\boldsymbol{x}_i)) \nonumber \\
&= \mathbb{E}_q \left[ \log\left(\frac{\sigma_{ik}^q}{\sigma_{ik}^p}\right) -\frac{(\beta_{ik} - \mu_{ik}^p)^2}{2(\sigma^p_{ik})^2} +\frac{(\beta_{ik} - \mu_{ik}^q)^2}{2(\sigma^q_{ik})^2} \right] \\
&= \log\left(\frac{\sigma_{ik}^q}{\sigma_{ik}^p}\right) + \mathbb{E}_q \left[ -\frac{(\beta_{ik} - \mu_{ik}^p)^2}{2(\sigma^p_{ik})^2} +\frac{(\beta_{ik} - \mu_{ik}^q)^2}{2(\sigma^q_{ik})^2} \right] \\
&= \log\left(\frac{\sigma_{ik}^q}{\sigma_{ik}^p}\right) - \frac{1}{2(\sigma^p_{ik})^2} \mathbb{E}_q \left[ (\beta_{ik} - \mu_{ik}^p)^2\right] + \frac{1}{2(\sigma^q_{ik})^2} \mathbb{E}_q \left[ (\beta_{ik} - \mu_{ik}^q)^2 \right].
\end{align}

The expectation of the squared difference from the mean is simply the
variance, i.e.,

\begin{equation}
(\sigma_{ik}^q)^2 = \mathbb{E}_q \left[(\beta_{ik} - \mu_{ik}^q)^2\right].
\end{equation}

Substituting the variance and simplifying further results in

\begin{align}
-D_{KL}(q_{\boldsymbol{\theta}}(\beta_{ik}| \boldsymbol{x}_i) || p(\beta_{ik}|\boldsymbol{x}_i)) &=
\log\left(\frac{\sigma_{ik}^q}{\sigma_{ik}^p}\right) - \frac{1}{2(\sigma^p_{ik})^2} \mathbb{E}_q \left[ (\beta_{ik} - \mu_{ik}^p)^2\right] + \frac{(\sigma^q_{ik})^2}{2(\sigma^q_{ik})^2} \\
&= \log\left(\frac{\sigma_{ik}^q}{\sigma_{ik}^p}\right) - \frac{1}{2(\sigma^p_{ik})^2} \mathbb{E}_q \left[ (\beta_{ik} - \mu_{ik}^p)^2\right] + \frac{1}{2} \\
&= \log\left(\frac{\sigma_{ik}^q}{\sigma_{ik}^p}\right) - \frac{1}{2(\sigma^p_{ik})^2} \mathbb{E}_q \left[ (\beta_{ik} - \mu_{ik}^q + \mu_{ik}^q - \mu_{ik}^p)^2\right] + \frac{1}{2}. 
\end{align}

Note that the final step expands the bracket by adding and subtracting
\(\mu_{ik}^q\). Grouping the terms in the bracket and multiplying out,
now yields

\begin{align}
-&D_{KL}(q_{\boldsymbol{\theta}}(\beta_{ik}| \boldsymbol{x}_i) || p(\beta_{ik}|\boldsymbol{x}_i))  \nonumber \\
=&\log\left(\frac{\sigma_{ik}^q}{\sigma_{ik}^p}\right)
- \frac{1}{2(\sigma^p_{ik})^2} \mathbb{E}_q \left[
(\beta_{ik} - \mu_{ik}^q)^2 + 2(\beta_{ik} - \mu_{ik}^q)(\mu_{ik}^q - \mu_{ik}^p) + (\mu_{ik}^q - \mu_{ik}^p)^2
\right]
+ \frac{1}{2} \\
= &\log\left(\frac{\sigma_{ik}^q}{\sigma_{ik}^p}\right)
- \frac{1}{2(\sigma^p_{ik})^2} \Bigg\{ \mathbb{E}_q \left[
(\beta_{ik} - \mu_{ik}^q)^2 \right] + 2\mathbb{E}_q \left[(\beta_{ik} - \mu_{ik}^q)(\mu_{ik}^q - \mu_{ik}^p) \right] + \mathbb{E}_q \left[(\mu_{ik}^q - \mu_{ik}^p)^2\right] \Bigg\} + \frac{1}{2}.
\end{align}

Given that \(\mathbb{E}_q[\beta_{ik} - \mu^q_{ik}] = 0\), the above
simplifies to

\begin{align}
-D_{KL}(q_{\boldsymbol{\theta}}(\beta_{ik}| \boldsymbol{x}_i) &|| p(\beta_{ik}|\boldsymbol{x}_i))  \nonumber \\
= &\log\left(\frac{\sigma_{ik}^q}{\sigma_{ik}^p}\right)
- \frac{1}{2(\sigma^p_{ik})^2} 
\left[
(\sigma_{ik}^q)^2 + 
2*0*(\mu_{ik}^q - \mu_{ik}^p)  + 
(\mu_{ik}^q - \mu_{ik}^p)^2
\right]
+ \frac{1}{2}  \\
= &\log\left(\frac{\sigma_{ik}^q}{\sigma_{ik}^p}\right)
- \frac{(\sigma_{ik}^q)^2+(\mu_{ik}^q - \mu_{ik}^p)^2}{2(\sigma^p_{ik})^2} + \frac{1}{2}. 
\end{align}

With the assumption of independently distributed prior and posterior
parameters, the stacked Kullback-Leibler penalty is obtained by summing
over \(K\), such that

\begin{equation}
-D_{KL}(q_{\boldsymbol{\theta}}(\boldsymbol{\beta}_{i}| \boldsymbol{x}_i) || p(\boldsymbol{\beta}_{i}|\boldsymbol{x}_i)) = \sum_{k = 1}^K \left[ \log\left(\frac{\sigma_{ik}^q}{\sigma_{ik}^p}\right)
- \frac{(\sigma_{ik}^q)^2+(\mu_{ik}^q - \mu_{ik}^p)^2}{2(\sigma^p_{ik})^2} + \frac{1}{2} \right].
\end{equation}

\newpage

\hypertarget{proof-of-proposition-1}{%
\section{\texorpdfstring{Proof of Proposition \ref{prop:prior}
\label{app:prop_prior}}{Proof of Proposition  }}\label{proof-of-proposition-1}}

Proposition \ref{prop:prior} states that when
\(\triangledown_{\boldsymbol{x}} \boldsymbol{\beta}_i = \boldsymbol{0}\)
and features are independent, the parameter vector
\(\boldsymbol{\beta}_i^m\) resembles the gradient of the PENN network
\(f(\boldsymbol{x})\) with respect to the inputs:

\begin{equation}
\boldsymbol{\beta}_i^m = \triangledown_{\boldsymbol{x}}f_m(\boldsymbol{x}_i).
\end{equation}

Using the product rule to compute the partial derivative of
\(f_m(\boldsymbol{x})\) with respect to \(x_{k}\), where
\(f_m(\boldsymbol{x}) = \sum_{k = 1}^K x_k\odot\beta_k^m\) (and
\(\odot\) represents the element-wise multiplication operator), yields:

\begin{equation}
\frac{\partial f_m(\boldsymbol{x})}{\partial x_k} = \beta_k^m + \frac{\partial q_{\boldsymbol{\theta}}(\boldsymbol{\beta}|\boldsymbol{x})}{\partial x_k} \boldsymbol{x}.
\label{app_prior_deriv_indep}
\end{equation}

Eq. \ref{app_prior_deriv_indep} implies that

\[
\left(\beta_k^m = \frac{\partial f_m(\boldsymbol{x})}{\partial x_k}\right) \Longleftrightarrow \left(\frac{\partial q_{\boldsymbol{\theta}}(\boldsymbol{\beta}|\boldsymbol{x})}{\partial x_k} = \boldsymbol{0}\right),
\]

where \(\boldsymbol{0}\) is a matrix of zeroes. Thus, when the gradient
of the inference network with respect to the inputs is zero, the
parameters are equal to the overall network gradient.

The derivative in the presence of dependent features changes slightly,
as shown in Eq. \ref{app_prior_deriv_dep}:

\begin{equation}
\frac{\partial f_m(\boldsymbol{x})}{\partial x_k} = \boldsymbol{\beta}^m\frac{\partial \boldsymbol{x}}{\partial x_k} + \frac{\partial q_{\boldsymbol{\theta}}(\boldsymbol{\beta}|\boldsymbol{x})}{\partial x_k} \boldsymbol{x}.
\label{app_prior_deriv_dep}
\end{equation}

Enforcing local stability such that
\(\partial q_{\boldsymbol{\theta}}(\boldsymbol{\beta}|\boldsymbol{x})/\partial x_k \approx \boldsymbol{0}\)
results in a neural network that encodes \(\boldsymbol{\beta}_i^m\)
subject to the constraint
\(\triangledown_{\boldsymbol{x}} f_m(\boldsymbol{x}_i) \approx \boldsymbol{\beta}_i^m \triangledown_{\boldsymbol{x}} \boldsymbol{x}_i\).

It is clear that the constraint is not equivalent to the parameter
values,
\(\boldsymbol{\beta}_i^m \triangledown_{\boldsymbol{x}} \boldsymbol{x}_i \neq \boldsymbol{\beta}_i^m\),
whenever
\(\triangledown_{\boldsymbol{x}} \boldsymbol{x}_i \neq \boldsymbol{I}_K\).
Training a neural network using a gradient penalty, such that
\(||\triangledown_{\boldsymbol{x}} f_m(\boldsymbol{x}_i)-\boldsymbol{\beta}_i^m||\approx 0\),
as in the case of the SENN, therefore results in a parameter bias when
features are dependent.

\newpage

\hypertarget{comparison-of-different-weighting-kernels}{%
\section{\texorpdfstring{Comparison of different weighting kernels
\label{app:kernel}}{Comparison of different weighting kernels }}\label{comparison-of-different-weighting-kernels}}

The prior density \(p(\boldsymbol{\beta}|\boldsymbol{x})\) is
parameterized using a kernel weighting function that identifies disjoint
neighborhoods (\(\mathcal{I}\)) of points in \(\boldsymbol{x}\)
satisfying the condition

\begin{equation}
||\boldsymbol{x}_i - \boldsymbol{x}_j|| < D \;\forall\; i,j \in \mathcal{I}.
\end{equation}

The number of neighborhoods is determined by the hyperparameter
\(\delta\).

Instead of defining disjoint neighborhoods, traditional kernels can be
used to calculate \(\boldsymbol{\pi}_{\boldsymbol{x}}\). This section
examines the effect of two compact support kernels (Epanechnikov and
tri-cube), with support regions determined using a bandwidth parameter
(\(b\)), such that

\begin{equation}
a = \frac{||\boldsymbol{x}_i - \boldsymbol{x}_j||}{b}.
\end{equation}

The Epanechnikov kernel is defined as

\begin{equation}
K_b(\boldsymbol{x}_i, \boldsymbol{x}_j) := H(a), \text{ where } H(a) := \begin{cases} \frac{3}{4}(1-a^2), \; &a<1 \\ 0, \; &\text{otherwise}\end{cases}
\end{equation}

and the tri-cube kernel is defined as

\begin{equation}
K_b(\boldsymbol{x}_i, \boldsymbol{x}_j) := H(a), \text{ where } H(a) := \begin{cases} (1-a^3)^3, \; &a<1 \\ 0, \; &\text{otherwise}\end{cases}.
\end{equation}

Now, a weighting matrix \(\boldsymbol{\pi}_{\boldsymbol{x};b}\) is
defined, whose \(i\)th row elements are given by
\(H(a) / \sum_{j}H(a)\), and which subsequently substitutes
\(\boldsymbol{\pi}_{\boldsymbol{x};\delta}\) during PENN training.

\begin{figure}
\centering
\includegraphics{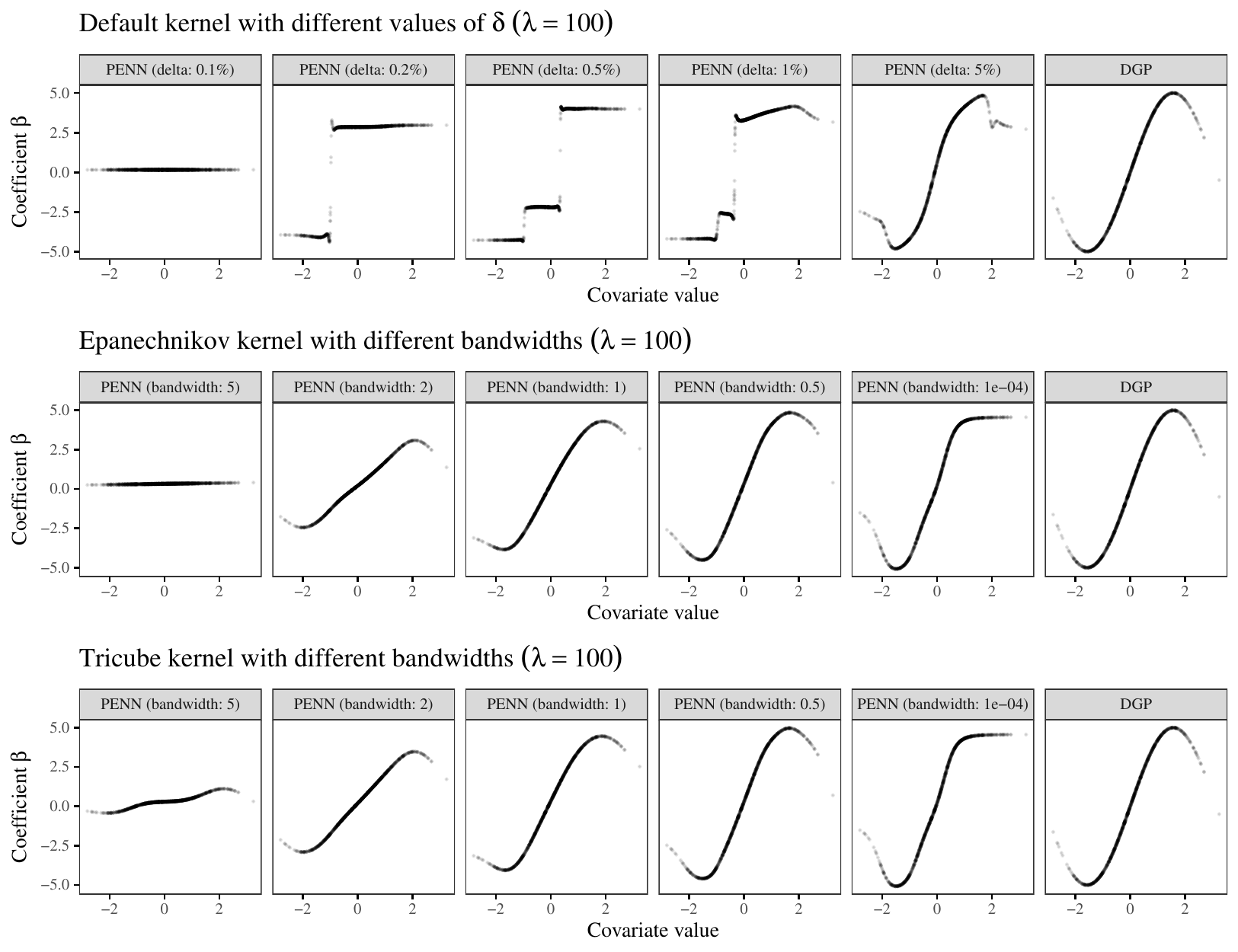}
\caption{Effect of kernel hyperparameter (\(\delta\) or bandwidth) on
parameter estimates when \(\lambda\) is high.\label{fig:kernels}}
\end{figure}

Fig. \ref{fig:kernels} plots the effect of the different kernels on the
coefficient used in Section \ref{role_of_hyperparameters}, when the
overall weight of the prior in the loss function is high
(\(\lambda = 100\)). The default kernel results in disjoint parameter
regimes, while both the Epanechnikov and tri-cube kernels smoothly
approximate the data generating process as the bandwidth is increased
(i.e.~the area of support is reduced). No clear preference is stated
here, with all kernels presenting valid approaches to learning
relational prior behavior. The smooth (as opposed to disjoint)
neighborhoods tend to result in much stronger regularization to static
parameters when data is noisy. Since smoothness in the parameter
estimates is also attained by reducing \(\lambda\) from the
illustratively high level used in the example, the disjoint kernel does
not limit flexibility. In fact, \(\delta\) has the intuitive appeal that
it defines clear parameter regimes with the smoothness of regime
transition depending on the level of \(\lambda\).

\newpage

\hypertarget{neural-network-hyperparameters}{%
\section{\texorpdfstring{Neural network hyperparameters
\label{appc}}{Neural network hyperparameters }}\label{neural-network-hyperparameters}}

Table \ref{tbl:nn_architecture} lists hyperparameters associated with
the neural network architecture used in the simulated and empirical
applications in Sections \ref{simulation} and \ref{application}:

\begin{table}
\centering
\begin{tabularx}{\textwidth}{ |l|c|X| } 
\hline
Hyperparameter & Value & Comments \\
\hline
Number of hidden layers & 2 & \\
Nodes per layer & 10-30 & Results relatively insensitive to the number of hidden nodes. Hyperparameter set to ensure a high degree of flexibility. \\
Activation function & Sigmoid & Similar results achieved with hyperbolic tangent or rectified linear unit functions. \\
Epochs & 200-500 & 500 for simulation, 200 for empirical application. Selected to ensure convergence. Longer training can lead to overfitting. \\
$M$ & 100 & The number of Monte Carlo draws from the posterior of the local parameters. \\
Optimizer & Adam & \\
Learning rate & 0.05 & Learning rate of the Adam optimizer. \\
$\beta_1$ & 0.9 & Exponential decay rate for first moment estimates for Adam optimizer. \\
$\beta_2$ & 0.999 & Exponential decay rate for second moment estimates for Adam optimizer. \\
$\ell_2$-norm clipping & 1 & Threshold for $\ell_2$-norm of gradients. \\ 
Gradient clipping & 0.5 & Threshold for the absolute size of gradients. \\
$\lambda$ & Varies & Selected using cross-validation procedures. \\
$\delta$ & Varies & Selected using cross-validation procedures. \\
\hline
\end{tabularx}
\caption{Neural network architecture and hyperparameter values used in Sections \ref{simulation} and \ref{application}.}
\label{tbl:nn_architecture}
\end{table}

\bibliography{Tex/ref}

\end{document}